\documentclass{article}

\PassOptionsToPackage{numbers, compress}{natbib}
\usepackage[final]{neurips_2023}

\usepackage[utf8]{inputenc} % allow utf-8 input
\usepackage[T1]{fontenc}    % use 8-bit T1 fonts
\usepackage{hyperref}       % hyperlinks
\usepackage{url}            % simple URL typesetting
\usepackage{booktabs}       % professional-quality tables
\usepackage{amsfonts}       % blackboard math symbols
\usepackage{nicefrac}       % compact symbols for 1/2, etc.
\usepackage{microtype}      % microtypography
\usepackage{xcolor}         % colors
\usepackage{wrapfig}
\usepackage{textcomp}
\usepackage[toc,page,header]{appendix}

\usepackage{silence}
\usepackage{math_commands}

\usepackage{titletoc}

\usepackage[textwidth=2.5cm]{todonotes}

\title{Complementary Benefits of Contrastive Learning and Self-Training 
Under Distribution Shift} % AR suggestion
\author{%
  Saurabh Garg\thanks{Equal contribution.} \\
  Carnegie Mellon University \\
  \texttt{\href{sgarg2@andrew.cmu.edu}{sgarg2@andrew.cmu.edu}}  
  \And
  Amrith Setlur$^{*}$\\
  Carnegie Mellon University \\
  \texttt{\href{asetlur@andrew.cmu.edu}{asetlur@andrew.cmu.edu}}
   \And
  Zachary C. Lipton\\
  Carnegie Mellon University \\
  \texttt{\href{zlipton@andrew.cmu.edu}{zlipton@andrew.cmu.edu}}
  \AND
  Sivaraman Balakrishnan\\
  Carnegie Mellon University \\
  \texttt{\href{sbalakri@andrew.cmu.edu}{sbalakri@andrew.cmu.edu}}
   \And
   Virginia Smith \\
  Carnegie Mellon University \\
  \texttt{\href{smithv@andrew.cmu.edu}{smithv@andrew.cmu.edu}}  
  \And
    Aditi Raghunathan\\
  Carnegie Mellon University \\
  \texttt{\href{aditirag@andrew.cmu.edu}{aditirag@andrew.cmu.edu}}  
}

\newcommand\si{\ensuremath{\mathcal{i}}}

\newcommand\sN{\ensuremath{\mathcal{N}}}

\newcommand\sY{\ensuremath{\mathcal{Y}}}

\ifthenelse{\isundefined{\definition}}{\newtheorem{definition}{Definition}}{}
\ifthenelse{\isundefined{\assumption}}{\newtheorem{assumption}{Assumption}}{}
\ifthenelse{\isundefined{\hypothesis}}{\newtheorem{hypothesis}{Hypothesis}}{}
\ifthenelse{\isundefined{\proposition}}{\newtheorem{proposition}{Proposition}}{}
\ifthenelse{\isundefined{\theorem}}{\newtheorem{theorem}{Theorem}}{}
\ifthenelse{\isundefined{\lemma}}{\newtheorem{lemma}{Lemma}}{}
\ifthenelse{\isundefined{\corollary}}{\newtheorem{corollary}{Corollary}}{}
\ifthenelse{\isundefined{\alg}}{\newtheorem{alg}{Algorithm}}{}
\ifthenelse{\isundefined{\example}}{\newtheorem{example}{Example}}{}
\newcommand{\E}{\ensuremath{\mathbb{E}}} %

\begin{document}

\maketitle

\begin{abstract}

Self-training and contrastive learning have emerged
as leading techniques for incorporating unlabeled data,
both under distribution shift (unsupervised domain adaptation)
and when it is absent (semi-supervised learning).
However, despite the popularity and compatibility of these techniques,
their efficacy in combination remains unexplored.
In this paper, we undertake a systematic
empirical investigation of this combination,
finding that (i) in domain adaptation settings,
self-training and contrastive learning 
offer significant complementary gains;
and (ii) in semi-supervised learning settings, surprisingly,
the benefits are not synergistic.
Across eight distribution shift datasets 
(\eg, BREEDs, WILDS), 
we demonstrate that the combined method 
obtains 3--8\% higher accuracy than either approach independently.
We then theoretically analyze these techniques
in a simplified model of distribution shift, 
demonstrating scenarios under which the features
produced by contrastive learning
can yield a good initialization for self-training to further amplify 
gains and achieve optimal performance,
even when either method alone would fail.

\end{abstract}

% \section{Introduction}
\section{Introduction}
\label{sec:intro}

Even when faced with natural, non-adversarial distribution shifts,
the performance of machine learning models may degrade~\citep{quinonero2008dataset, torralba2011unbiased, wilds2021, garg2022ATC}.
While we might hope to retrain such models
on labeled samples from the new distribution,
this option is often unavailable
due to the expense or impracticality
of collecting new labels.
Consequently, researchers have investigated 
solutions to Unsupervised Domain Adaptation (UDA). 
Here, given labeled \emph{source} data 
and unlabeled out-of-distribution (OOD) \emph{target} data,
the goal is to produce a classifier 
that performs well on the target. To address UDA in practice, two popular methods have emerged: self-training and contrastive pretraining.

Self-training~\citep{scudder1965probability, lee2013pseudo, sohn2020fixmatch, xie2020self, wang2021tent} and contrastive pretraining~\citep{caron2020unsupervised, chen2020simple, zbontar2021barlow} 
were both proposed, initially, for traditional Semi-Supervised Learning (SSL) problems,
where the labeled and unlabeled data are drawn from the same distribution.
Here, the central challenge is statistical: 
to exploit the unlabeled data to learn a better predictor 
than one would get by training on the (small) labeled data alone. 
More recently, these methods have emerged 
as favored empirical approaches for UDA,
demonstrating efficacy on many popular benchmarks~\citep{sagawa2021extending, 
garg2022RLSbench, cai2021theory, shen2022connect}. 
In self-training, one first learns a predictor using source labeled data.
The predictor then produces pseudolabels for the unlabeled target data,
and a new predictor is trained on the pseudolabeled data. 
Contrastive pretraining learns representations from unlabeled data 
by enforcing invariance to specified augmentations.
These representations are subsequently used to learn a classifier.
In UDA, the representations are trained 
on the union of the source and target data. Despite the strong performance of self-training and constrastive pretraining independently, there has been  surprisingly little work explaining when either might be expected to perform best and whether the benefits might be complementary.

\begin{figure*}[!t]
    \centering
    \begin{subfigure}[b]{0.48\linewidth}
    \includegraphics[width=\linewidth]{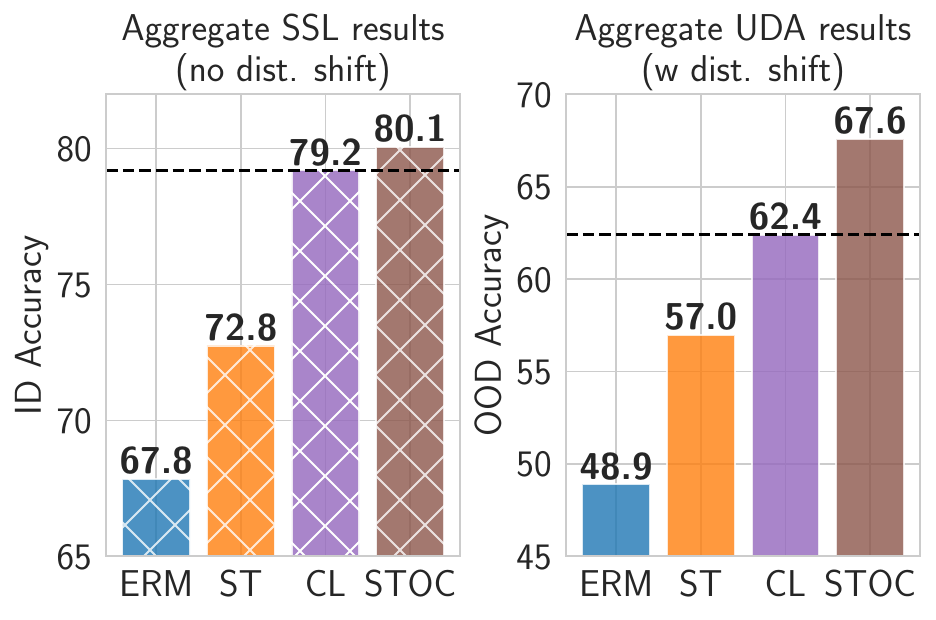}
    \caption{}
    \end{subfigure}
    \hfill 
    \color{gray}\vrule width 0.04cm{} \hfill
    \begin{subfigure}[b]{0.46\linewidth}
    \includegraphics[width=\linewidth]{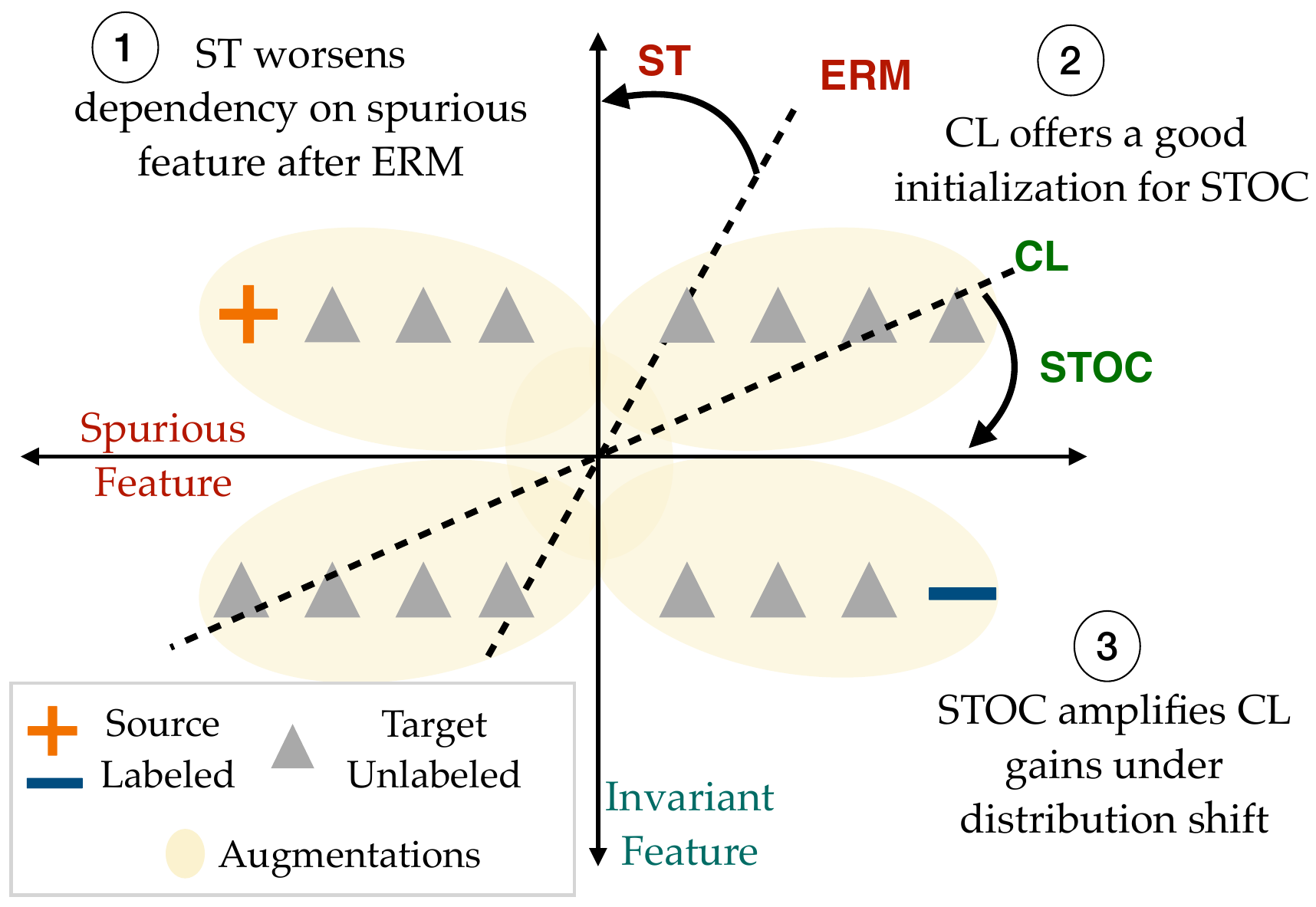} 
     \caption{}   
    \end{subfigure}
    \caption{\emph{Self-training over Contrastive learning (STOC) improves over Contrastive Learning (CL) under distribution shift.} \textbf{(a)} We observe that in SSL settings, where labeled and unlabeled data are drawn from the same distribution, STOC offers negligible improvements over CL. In contrast, in UDA settings where there is distribution shift between labeled and unlabeled data, STOC offers gains over CL. Results aggregated across 8 benchmarks. Results on individual data in Table \ref{table:UDA_results} and \ref{table:SSL_results}. \textbf{(b)} 2-D illustration of our simplified distribution setup, depicting decision boundaries learned by ERM and CL and how Self-Training (ST) updates those.  \textcircled{\fontsize{7pt}{0}\fontfamily{phv}\selectfont 1}, \textcircled{\fontsize{7pt}{0}\fontfamily{phv}\selectfont 2}, and \textcircled{\fontsize{7pt}{0}\fontfamily{phv}\selectfont 3} summarize our theoretical results in \secref{sec:theory}.
    }
    \label{fig:intro-figure}
\end{figure*}

In this paper, we investigate the complementary benefits
of self-training and contrastive pretraining. 
Interestingly, we find that the combination
yields significant gains in UDA 
despite producing negligible gains in SSL.
In experiments across eight distribution shift benchmarks
(\eg BREEDs~\citep{santurkar2020breeds}, 
FMoW~\citep{wilds2021}, Visda~\citep{visda2017}),
we observe that re-using unlabeled data for self-training 
(with FixMatch~\citep{sohn2020fixmatch}) 
after learning contrastive representations 
(with SwAV~\citep{caron2020unsupervised}),
yields $>5$\%  average improvement on OOD accuracy in UDA 
as compared to $<0.8\%$ average improvement in SSL 
(\figref{fig:intro-figure}).

{Next, we aim to understand
\textit{why} the combination of self-training and contrastive learning is synergistic under distribution shift. To do so, we analyze both methods in a simplified distribution shift setting that models domain-independent or invariant, and domain-specific or spurious features. Our theoretical analysis highlights that: (i) under suitable augmentations contrastive pretraining on unlabeled data can learn a feature extractor that amplifies the invariant feature over the spurious (\emph{feature amplification}); and (ii) self-training (ST) can learn the optimal target linear predictor, when initialized with a ``good'' classifier (learnt over contrastive features), thus improving \emph{linear transferability}. 
We also show that contrastive pretrained features continue to be correlated with spurious features, and as a result the linear predictor (CL) learnt using source labeled data over these features is suboptimal on target.
Still, Cl outperforms source-only ERM in providing ``good'' initial pseudolabels on the target unlabeled data.  
Thus, self-training over the CL predictor (STOC) pretrained features unlearns any reliance on domain-dependent features 
and improves OOD performance relative to either method independently.}

Finally, we connect our theoretical understanding of feature amplification done by contrastive learning, and improved linear transferability from self-training back to observed empirical gains. %
We linearly probe representations (fix representations and train only the linear head) learned by contrastive pretraining vs. no pretraining and find: 
(i) contrastive pretraining substantially improves the ceiling on
the target accuracy (performance of optimal linear probe) compared to ERM; (ii) self-training mainly improves linear transfer, 
\ie OOD performance for the linear probe trained with source labeled data.

{The remainder of the paper is organized as follows. We introduce the problem setup and algorithmic preliminaries in ~\secref{sec:problem-setup}, followed by our main empirical results in~\secref{sec:exp}, for both UDA and SSL settings. In ~\secref{sec:theory} we present our theoretical analysis explaining the complimentary gains of ST and CL under distribution shift, and finally reconnect this theoretical understanding to empirical trends from real-world settings in~\secref{sec:discussion}.}

% \input{sections/introduction}

% \vspace{-7pt}
\section{Setup and Preliminaries}
% \vspace{-5pt}
\label{sec:problem-setup}

\textbf{Task.~~} Our goal is to learn a predictor 
that maps inputs $x \in \calX \subseteq \R^d$ 
to outputs $y \in \sY$. 
We parameterize predictors $f = h \circ \Phi : \R^d \mapsto \sY$, where $\Phi:\R^d \mapsto \R^k$ is a feature map 
and $h \in \R^{k}$ is a classifier 
that maps the representation to the final scores or logits.
Let $\ProbS, \ProbT$ be the source and target joint probability measures over $\inpt\times\out$ with $\ps$ and $\pt$ 
as the corresponding probability density (or mass) functions. 
The distribution over unlabeled samples
from both the union of source and target
is denoted as $\ProbU = (1/2)\cdot \ProbS(x) + (1/2) \cdot \ProbT(x)$.

We study two particular scenarios:
(i) Unsupervised Domain Adaptation (UDA);
and (ii) Semi-Supervised Learning (SSL). 
In UDA, we assume that the source and target distributions 
have the same label marginals $\ProbS(y) = \ProbT(y)$ 
(\ie, no label proportion shift) 
and the same Bayes optimal predictor, 
\ie, $\argmax_y \ps(y \mid x) = \argmax_y \pt(y \mid x)$. 
We are given labeled samples from the source, 
and unlabeled pool from the target. 
In contrast in SSL, there is no distribution shift, 
\ie,  $\ProbS=\ProbT=\ProbU$. 
Here, we are given a small number of labeled examples 
and a comparatively large amount of unlabeled examples, 
both drawn from the same distribution, 
which we denote as $\ProbT$. 

Unlabeled data is typically much cheaper to obtain, 
and our goal in both these settings is to leverage this along with labeled data to achieve good performance on the target distribution. 
In the UDA scenario, the challenge lies in generalizing out-of-distribution, while in SSL, the challenge is to generalize in-distribution despite the paucity of labeled examples.
A predictor $f$ is evaluated on distribution $\Prob$ via its accuracy, \ie, $A(f, \Prob) = \mathbb{E}_{\Prob} (\argmax f(x) = y)$.

\textbf{Methods.} 
We now introduce the algorithms used for learning
from labeled and unlabeled data.

\begin{enumerate}[itemsep=0pt,topsep=0pt,  leftmargin=*]
    \item \emph{Source-only ERM (ERM)}: 
    A standard approach is to simply perform supervised learning
    on the labeled data by minimizing the empirical risk 
    $\sum_{i=1}^n \ell(h \circ \Phi(x), y)$, 
    for some classification loss 
    $\ell:\Real \times \calY \mapsto \Real$
    (\eg, softmax cross-entropy) and labeled points 
    $\{(x_i, y_i)\}_{i=1}^n$.
    \item \emph{Contrastive Learning (CL)}: 
    We first use the unlabeled data to learn a feature extractor.
    In particular, the objective is to learn a feature extractor $\Phi_\cl$ that maps augmentations (for e.g. crops or rotations) 
    of the same input close to each other 
    and far from augmentations of random other inputs~\citep{caron2020unsupervised, chen2020simple, zbontar2021barlow}. Once we have $\Phi_\cl$, we learn a linear classifier $h$ on top to minimize a classification loss on the labeled source data. We could either keep $\Phi_\cl$ fixed or propagate gradients through. 
    
    When clear from context, we also use CL to refer to just the contrastively pretrained backbone 
    without training for downstream classification. 
    \item \emph{Self-training (ST)}: This is a two-stage procedure, where the first stage performs source-only ERM by just looking at source-labeled data. In the second stage, we iteratively apply the current classifier on the unlabeled data to generate ``pseudo-labels'' and then update the classifier by minimizing a classification loss on the pseudolabeled data~\citep{lee2013pseudo}. 
\end{enumerate}

\section{Self-Training Improves Contrastive Pretraining Under Distribution Shift} \label{sec:exp}
\label{sec:main-claim}

\textbf{Self-Training Over Contrastive learning (STOC).~~}      
    Finally, rather than starting with a source-only ERM classifier,
    we propose to initialize self-training with a CL classifier,
    that was pretrained on unlabeled source and target data.
    ST uses that same unlabeled data again for pseudolabeling. 
    As we demonstrate experimentally and theoretically, 
    this combination of methods improves substantially
    over each independently.

\textbf{Datasets.~~} 
For both UDA and SSL, we conduct experiments 
across eight benchmark datasets: 
four BREEDs datasets~\citep{santurkar2020breeds}---{Entity13}, {Entity30}, {Nonliving26}, {Living17}; {FMoW}~\citep{wilds2021,christie2018functional} from {WILDS} benchmark;
Officehome~\citep{venkateswara2017deep};
{Visda}~\citep{peng2018syn2real, visda2017};
and CIFAR-10~\citep{krizhevsky2009learning}. 
Each of these datasets consists of domains,
enabling us to construct source-target pairs
(e.g., CIFAR10, we consider CIFAR10$\to$CINIC shift~\citep{darlow2018cinic}). 
In the UDA setup, we adopt the source and target domains
standard to previous studies
(details in \appref{app:dataset}). 
Because the SSL setting lacks distribution shift,
we do not need to worry about domain designations and default to using source alone. 
To simulate limited supervision in SSL, 
we sub-sample the original labeled training set to 10\%.  

\begin{table}[t]
  \centering
  \footnotesize
  \setlength{\tabcolsep}{6pt}
  \renewcommand{\arraystretch}{1.2}
  \caption{\emph{Results in the UDA setup}. We report accuracy on target (OOD) data from which we only observe unlabeled examples during training. 
  For benchmarks with multiple target distributions (\eg, OH, Visda), we report avg accuracy on those targets. Results with source performance, individual target performance, and standard deviation numbers are in \appref{app:additional_UDA_results}.}\label{table:UDA_results}
  \vspace{5pt}
  \resizebox{\linewidth}{!}{%
  \begin{tabular}{@{}*{10}{c}@{}|c}
  \toprule
  {Method} & \parbox{1.cm}{Living17} & \parbox{1.cm}{Nonliv26} & \parbox{1.cm}{Entity13} & \parbox{1.cm}{Entity30} & \parbox{1.cm}{FMoW (2 tgts)} & \parbox{1.cm}{Visda (2 tgts)}  & \parbox{1.cm}{~~OH \\(3 tgts)}   & \parbox{1.cm}{CIFAR$\to$ CINIC}  && Avg  \\
  \midrule
  ERM & $60.31$  & $45.54$  &  $68.32$ & $55.75$  & $56.50$  & $20.91$  & $9.51$  & $74.33$ && $48.90$ \\
  ST  & $71.29$ & $56.79$ & $77.93$ & $66.37$ &  $56.79$ & $38.03$ & $10.47$  & $78.19$ && $56.98$ \\
  CL &  $74.14$ & $57.02$ &  $76.58$ &  $66.01$ &   $61.78$ & 
 $63.49$ &  $22.63$ & $77.51$ && $62.39$ \\ 
  STOC (ours) & $\mathbf{82.22}$& $\mathbf{62.23}$ &$\mathbf{81.84}$ & $\mathbf{72.00}$  & $\mathbf{65.25}$ & $\mathbf{70.08}$  & $\mathbf{27.12}$ & $\mathbf{79.94}$ && $\mathbf{67.59}$  \\
  \bottomrule 
  \end{tabular}}  
\end{table}

\begin{table}[t]
  \centering
  \footnotesize
  \setlength{\tabcolsep}{6pt}
  \renewcommand{\arraystretch}{1.2}
  \caption{\emph{Results in the SSL setup}. We report accuracy on hold-out ID data. Recall that SSL uses labeled and unlabeled data from the same distribution during training. Refer to \appref{app:additional_SSL_results} for ERM and ST. }\label{table:SSL_results}
  \vspace{5pt}
  \resizebox{\linewidth}{!}{%
  \begin{tabular}{@{}*{10}{c}@{}|c}
  \toprule
  {Method} & \parbox{1.cm}{Living17} & \parbox{1.cm}{Nonliv26} & \parbox{1.cm}{Entity13} & \parbox{1.cm}{Entity30} & \parbox{1.cm}{FMoW} & \parbox{1.cm}{Visda}  & \parbox{1.cm}{~~OH}   & \parbox{1.cm}{CIFAR}  && Avg  \\
  \midrule
  CL  & $91.15$ & $84.58$ & $90.73$ & $85.47$ & $43.05$ & $97.67$ & $49.73$ & $91.78$ && $79.27$ \\
  STOC (ours)  & $92.00$ & $85.95$ & $91.27$ & $86.14$ & $44.43$ & $97.70$ & $49.95$ & $93.06$ && $80.06$ \\
  \bottomrule 
  \end{tabular}}  
\end{table}

\textbf{Experimental Setup and Protocols.~~} 
SwAV~\citep{caron2020unsupervised} is the specific algorithm
that we use for contrastive pretraining.
In all UDA settings, unless otherwise specified,
we pool all the (unlabeled) data 
from the source and target to perform SwAV.
For self-training, we apply FixMatch~\citep{sohn2020fixmatch},
where the loss on source labeled data 
and on pseudolabeled target data
are minimized simultaneously.
For both methods, we fix the algorithm-specific hyperparameters
to the original recommendations. 
For SSL settings, we perform SwAV and FixMatch 
on in-distribution unlabeled data.
We experiment with Resnet18, Resnet50~\citep{he2016deep} 
trained from scratch (\ie random initialization).
We do not consider off-the-shelf pretrained models
(\eg, on Imagenet~\citep{russakovsky2015imagenet}) 
to avoid confounding our conclusions about contrastive pretraining.
However, we note that our results on most datasets 
tend to be comparable to and sometimes exceed
those obtained with ImageNet-pretrained models.
For source-only ERM, as with other methods (FixMatch, SwAV), 
we default to using strong augmentation techniques: 
random horizontal flips, random crops,
augmentation with Cutout~\citep{devries2017improved}, 
and RandAugment~\citep{cubuk2020randaugment}.   
Moreover, unless otherwise specified, 
we default to full finetuning with source-only ERM, 
both from scratch and after contrastive pretraining, 
and for ST with FixMatch. 
For UDA, given that the setup precludes access 
to labeled data from the target distribution,
we use source hold-out performance 
to pick the best hyperparameters.
During pretraining, early stopping is done according to
lower values of pretraining loss. 
For more details on datasets, model architectures, and 
experimental protocols, see \appref{app:experimental details}\footnote{For SwAV we use the code from \url{https://github.com/facebookresearch/swav}, and for self-training we use \url{https://github.com/acmi-lab/RLSbench}.}.

\textbf{Results on UDA setup.~~} Both ST and CL individually improve over ERM across all datasets, with CL significantly performing better than ST on 5 out of 8 benchmarks (see \tabref{table:UDA_results}). Even on datasets where ST is better than CL, their performance remains close. Combining ST and CL with STOC shows 
an  $3$--$8\%$ improvement over the best alternative, 
yielding an absolute improvement in average accuracy of $5.2\%$. 

Note that by default, we train with CL on
the combined unlabeled data from source and target. 
However, to better understand the significance 
of unlabeled target data in contrastive pretraining, 
we perform an ablation where the CL model was trained 
solely on unlabeled source data (refer to this as CL (source only); see \appref{app:additional_UDA_results}). 
We observe that ST on top of CL (source only) 
improves over ST (from scratch). 
However, the average performance of ST over CL (source only) 
is similar to that of standalone CL, 
maintaining an approximate 6\% performance gap 
observed between CL and ST.
This brings two key insights to the fore: 
(i) the observed benefit is not merely a result 
of the contrastive pretraining objective alone, 
but specifically CL with unlabeled target data helps;
and (ii) both CL and ST leverage using target unlabeled data
in a complementary nature.

\textbf{Results on SSL setup.~~} 
While CL improves over ST (as in UDA), 
unlike UDA, STOC doesn't offer any significant improvements 
over CL (see \tabref{table:SSL_results}; 
ERM and ST results (refer to \appref{app:additional_SSL_results}). 
We conduct ablation studies with varying proportions
of labeled data used for SSL, 
illustrating that there's considerable potential 
for improvement (see \appref{app:additional_SSL_results}). 
These findings highlight that the complementary nature
of STOC over CL and ST individually 
is an artifact of distribution shift.

\section{Theoretical Analysis and Intuitions}
\label{sec:theory}

Our results on real-world datasets 
suggest that although self-training may offer 
little to no improvement over contrastive pretraining 
for in-distribution (\ie, SSL) settings, 
it leads to substantial improvements 
when facing distribution shifts in UDA (\secref{sec:exp}). 
Why do these methods offer complementary gains, 
but only under distribution shifts? 
In this section, we seek to answer this question 
by first replicating all the empirical trends of interest 
in a simple data distribution with an intuitive story (\secref{subsec:intuitive-story}). 
In this toy model, we formally characterize the gains afforded 
by contrastive pretraining and self-training both individually 
(Secs. \ref{subsec:self-training-analysis}, \ref{subsec:contrastive-pretraining-analysis}) 
and when used together 
(\secref{subsec:combining-self-training-and-contrastive-pretraining}).

\newcommand{\xone}{\rvx_1}
\newcommand{\xtwo}{\rvx_2}
\newcommand{\h}{\rvh}
\newcommand{\B}{\mathbf{\Phi}}
\newcommand{\Bone}{\rmB_1}
\newcommand{\Btwo}{\rmB_2}
\newcommand{\zone}{\rvz_1}
\newcommand{\ztwo}{\rvz_2}

\textbf{Data distribution~~} 
We consider binary classification and model 
the inputs as consisting of two kinds of features:
$x = [\xin, \xsp]$,
where $\xin \in \R^{\din}$ is the invariant feature 
that is predictive of the label
across both source $\ProbS$ and target $\ProbT$
and $\xsp \in \R^{\dsp}$ is the spurious feature 
that is correlated with the label $y$ 
only on the source domain $\ProbS$ 
but uncorrelated with label $y$ in $\ProbT$.
Formally, we sample $\ry \sim \unif \{-1, 1\}$ 
and generate inputs $x$ conditioned on $\ry$ as follows: 
\begin{align}
\footnotesize
\ProbS&:~ \xin \sim  \sN(\gamma \cdot \ry \wstar, \Sigmain)~~\xsp = \ry \mathbf{1}_{\dsp}\, \nonumber  \\  
\ProbT&:~ \xin \sim \sN(\gamma \cdot \ry \wstar, \Sigmain)~~\xsp \sim \sN(\mathbf{0}, \Sigmasp), \label{eq:toy-setup}
\end{align}
where $\gamma$ is the margin afforded by the invariant feature\footnote{See \appref{app:toy_description} for similarities 
and differences of our setup with prior works.}. We set the covariance of the invariant features $\Sigmain = \sigmain^2\cdot(\mathbf{I}_{\din} - {\wstar\wstar}^\top)$. This makes the variance along the unknown predictive direction $\wstar$ to be zero. Note that the spurious feature is also completely predictive
of the label in the source data. In fact, when $\dsp$ is sufficiently large, $\xsp$ is more predictive (than $\xin$) of $\ry$ in the source. 
In the target, $\xsp$ is distributed as a Gaussian 
with $\Sigmasp = \sigmasp^2 \mathbf{I}_{\dsp}$.
We use 
$\winv$$=$$[\wstar, 0, ..., 0]^\top$ to refer to the invariant predictor (or direction), and $\wspu = [0, ..., 0, \nicefrac{\mathbf{1}_{d_\mathrm{sp}}}{\sqrt{d_\mathrm{sp}}}]^\top$ for the spurious direction. 

\paragraph{Data for UDA vs. SSL} For convenience, whenever we have unlabeled data, we assume access to infinite unlabeled data and replace their empirical quantities with population counterparts. 
For SSL, we sample both finite labeled and infinite unlabeled data from the same distribution $\ProbT$, where spurious features are absent (to exclude easy-to-generalize features). For UDA, we further assume infinite labeled data from $\ProbS$ (in addition to infinite unlabeled from $\ProbT$). Importantly, note that due to distribution shift, population access of $\ProbS$ still captures the interesting aspects of distribution shifts---ERM on infinite labeled source data \emph{does not} achieve optimal performance on target.

\textbf{Methods and objectives~~} Recall from Section~\ref{sec:problem-setup}
that we learn linear classifiers $h$ over feature extractor $\Phi$.
For our toy setup, we consider linear feature extractors i.e. $\Phi$ is a matrix in $\R^{d \times k}$ and
the prediction $f(x) = \sgn(h^\top \Phi x)$. We use the exponential loss $\ell (f(x), y) = \exp\left(-y f(x)\right)$.

\textit{Self-training.} ST performs ERM
in the first stage using labeled data from the source,
and then subsequently updates the head $h$ 
by iteratively generating pseudolabels on the unlabeled target:
\begin{align}
\footnotesize
    \calL_\st(h; \Phi) \;\; & \coloneqq \;\; \Exp_{\ProbT(x)} \ell(h^\top \Phi x, \sgn(h^\top \Phi(x)))\qquad \nonumber \\
   \textrm{Update:}
   \;\; & h^{t+1}  \;\; = \;\; \frac{h^{t} - \eta \nabla_h \calL_\st (h^t; \Phi)}{\norm{h^{t} - \eta \nabla_h \calL_\st (h^t; \Phi)}{2}}
    \label{eq:self-training}
\end{align}
For ERM and ST, we train both $h$ and $\Phi$ (equivalent to $\Phi$ being identity and training a linear head). 

\textit{Contrastive pretraining.} We obtain $\Phi_\cl \coloneqq \argmin_\Phi \calL_\cl(\Phi)$ by minimizing the Barlow Twins objective~\citep{zbontar2021barlow}, 
which prior works have shown is also equivalent to spectral contrastive and non-contrastive objectives~\citep{garrido2022duality, cabannes2023ssl}. 
Given probability distribution $\ProbA(a \mid x)$ for input $x$,
and marginal $\ProbA$, 
we consider a constrained form of Barlow Twins
in \eqref{eq:cont-loss} which enforces 
features of ``positive pairs''  $a_1, a_2$
to be close while ensuring feature diversity.
We assume a strict regularization $(\rho=0)$ 
for the theory arguments in the rest of the paper, 
and in \appref{app:objectives} we prove that
all our claims hold for small $\rho$ as well.
For augmentations, we scale the magnitude 
of each co-ordinate uniformly by an independent amount, 
i.e., $a \sim \ProbA(\cdot \mid x) = \rvc \odot x $,
where $\rvc \sim \unif[0, 1]^d$. 
We try to mirror practical settings
where the augmentations are fairly ``generic'', 
not encoding information about which features 
are invariant or spurious, 
and hence perturb all features symmetrically. 
\begin{align}
    \footnotesize
    \calL_\cl(\Phi) \; \coloneqq & \; \Exp_{x \sim \ProbU}\Exp_{a_1, a_2 \sim \ProbA(\cdot \mid x)} \; \|\Phi(a_1) - \Phi(a_2)\|_2^2 \;\;\nonumber \\  \textrm{s.t.} & \;\; 
    \norm{\Exp_{a \sim \ProbA}\brck{\Phi(a)\Phi(a)^\top} - \mathbf{I}_k}{F}^2 \leq \rho
     \label{eq:cont-loss}
\end{align}
Keeping the $\Phi_\cl$ fixed, we then learn a linear classifier $h_\cl$ over $\Phi_\cl$ to minimize the exponential loss on labeled source data (refer to as \emph{linear probing}). 
For STOC, keeping the $\Phi_\cl$ fixed and initializing the linear head with the CL linear probe (instead of source only ERM), we perform ST with \eqref{eq:self-training}.

\begin{example}\label{ex:toy_setup}
For the setup in \eqref{eq:toy-setup}, we choose $\gamma = 0.5$, $\sigmasp^2 = 1.$, and $\sigmain^2 = 0.05$ with $\din = 5$ and $\dsp=20$ for our running example. 
$\nicefrac{\gamma}{\sqrt{\dsp}}$ controls signal to noise ratio in the source such that spurious feature is easy-to-learn and the invariant feature is harder-to-learn. 
$\sigma_2$ controls the noise in target which we show later is critical in unlearning the spurious feature with CL.  
\end{example}

\subsection{Simulations and Intuitive Story: A Comparative Study Between SSL and DA}
\label{subsec:intuitive-story}
\begin{figure*}[!t]
    \centering
    \begin{subfigure}[b]{0.26\linewidth}
    \includegraphics[width=\linewidth]{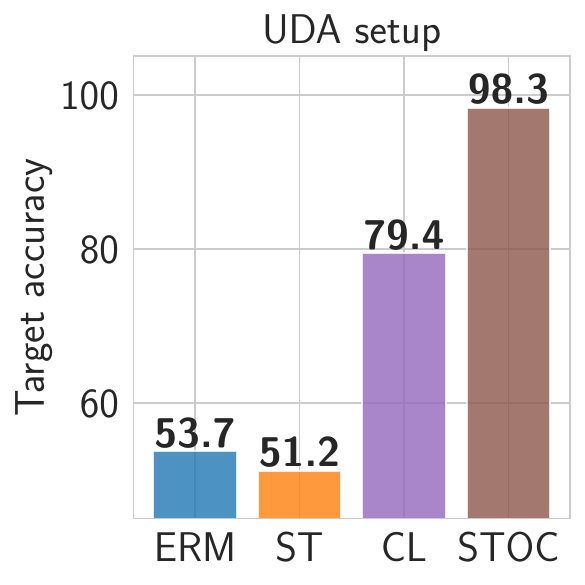}\hfill
    \caption{}
    \end{subfigure}
    \hfill
    \color{gray}\vrule width 0.04cm{}
    \hfill
    \begin{subfigure}[b]{0.44\linewidth}
    \includegraphics[width=\linewidth]{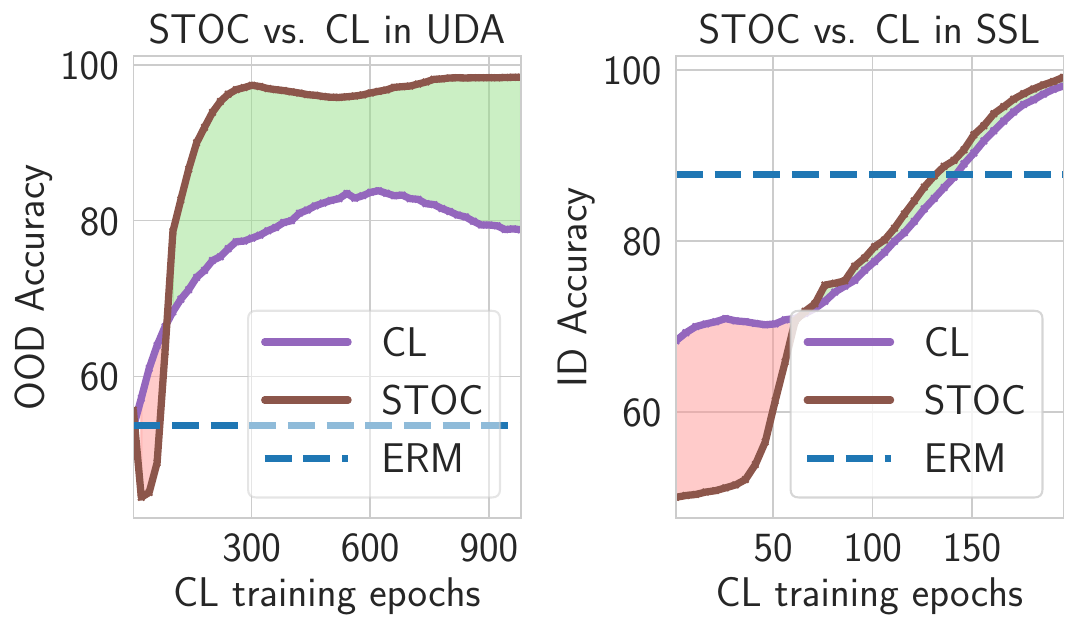}\hfill
    \caption{}
    \end{subfigure}
    \hfill
    \color{gray}\vrule width 0.04cm{}
    \hfill
    \begin{subfigure}[b]{0.26\linewidth}
    \includegraphics[width=\linewidth]{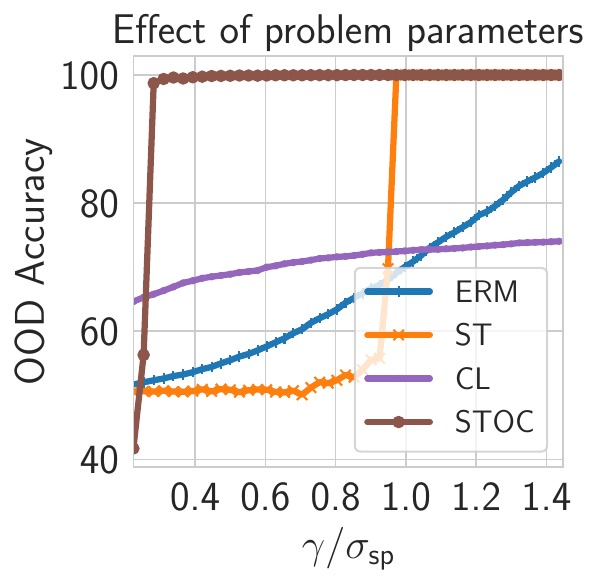}
    \caption{}
    \end{subfigure}
    \vspace{-5pt}
    \caption{
    \emph{Our simplified model of shift captures real-world trends and theoretical behaviors:}
    \textbf{(a)} Target (OOD) accuracy separation in the UDA setup (for problem parameters in Example~\ref{ex:toy_setup}).
    \textbf{(b)} Comparison of the benefits of STOC (ST over CL) over just CL in UDA and SSL settings, done across training iterations for contrastive pretraining.
    \textbf{(c)} Comparison between different methods in UDA setting, as we vary problem parameters $\gamma$ and $\sigmasp$, connecting our theory results in \secref{sec:theory}.
    }
    \vspace{-12pt}
    \label{fig:analysis-empirical}
\end{figure*}

\textbf{Our setup captures real-world trends in UDA setting.~~} Our toy setup (in Example~\ref{ex:toy_setup}) accentuates the 
behaviors observed on real-world datasets (\figref{fig:analysis-empirical}{(a)}):
(i) both ERM and ST yield close to random performance (though ST performs slightly worse than ERM);  %
(ii) CL improves over ERM but still 
yields sub-optimal target performance; %
(iii) STOC then further 
improves over CL, achieving near-optimal target performance. 
Note that, a linear predictor can improve target performance only by reducing its dependence on spurious feature $\xsp$, and increasing it on invariant feature $\xin$ (along $\wstar$). 
Given this, we can explain our trends if we understand the following: 
(i) how ST reduces dependence on spurious feature
when done after CL; 
(ii) why CL helps reduce but not completely eliminate the reliance of linear head 
on spurious features. 
Before we present intuitions, we ablate over a key problem parameter that affects both the target performance and conditions for ST to work.

\textbf{An intuitive story.~} 
We return to the question of why self-training improves over 
contrastive learning under distribution shift in our Example~\ref{ex:toy_setup}. 
When the classifier at initialization of ST
relies more on spurious features, ST 
aggravates this dependency. 
However, as the problem becomes easier (with increasing $\gamma/\sigmasp$), 
the source-only ERM classifier will start relying more on invariant rather than spurious feature. 
Once this ERM classifier is sufficiently accurate on the target, ST 
unlearns any dependency on spurious features achieving optimal target performance. This is because the initial pseudolabels on target unlabeled data are sufficiently accurate for self-training to improve \emph{linear transferability}. 
In contrast, we observe that CL performs better than ERM since contrastive pretraining learns a feature map that is correlated more with the invariant than the spurious feature. 
This implies that CL does \emph{feature amplification}: decreasing reliance on spurious features (as compared to ERM), 
but doesn't completely eliminate them, thereby remaining sub-optimal on target.
Combining ST and CL, a natural hypothesis explaining our trends is 
that CL provides a favorable initialization (through feature amplification) for ST to now improve linear transferability.

\textbf{Effect of $\nicefrac{\gamma}{\sigmasp}$ on success of ST.~~} 
Our intuitive understanding is reinforced by our experiment that increases the ratio of margin $\gamma$ and variance of spurious feature on target $\sigmasp$ (keeping others constant). Doing this makes the problem becomes easier because $\gamma$ directly affects the signal on $\xin$ and reducing $\sigmasp$ helps ST to unlearn $\xsp$ (see \appref{app:additional_toy_exp}).
In \figref{fig:analysis-empirical}(c), we see that a phase transition occurs for ST, \ie,  after a certain threshold of $\nicefrac{\gamma}{\sigmasp}$, ST successfully recovers the optimal target predictor.
This hints that ST has a binary effect, where beyond a certain magnitude of $\nicefrac{\gamma}{\sigmasp}$, ST can amplify the signal on domain invariant feature to obtain optimal target predictor. This explains the ability of ST to improve linear transferability when the initial classifier has sufficiently low target error. 
On the other hand, the performance of CL and ERM improve gradually where CL achieves high performance due to feature amplification, which occurs at even small ratios of $\nicefrac{\gamma}{\sigmasp}$. 
One way of viewing this trend with CL is that it magnifies the effective $\nicefrac{\gamma}{\sigmasp}$ in its representation space, because of which a linear head trained over these representations has a good performance at low values of the ratio. Consequently, the \emph{phase transition} of STOC occurs much sooner then that of ST. 
Finally, we note that for CL the rate of performance increase diminishes at high values of $\nicefrac{\gamma}{\sigmasp}$ because CL fails to reduce dependency along $\xsp$ beyond a certain point.

\textbf{Why disparate behaviors for out-of-distribution vs. in distribution?~~}
In the SSL setup, recall, there is no distribution shift. 
In Example~\ref{ex:toy_setup}, 
we sample $50k$ unlabeled data 
and $100$ labeled data from the 
same (target) distribution to simulate SSL setup. 
Substantiating our findings on real-world data, we observe 
that STOC provides a small to negligible improvement 
over CL (refer to \appref{appsec:additional-toy-results}). 
To understand why such disparate behaviors 
emerge, recall that in the UDA setting, 
the main benefit of STOC lies in picking up reliance on 
``good'' features for OOD data, facilitated by CL initialization. 
While contrastive pretraining uncovers features that are 
``good'' for OOD data, it also learns more predictive source-only features (which are not predictive at all on target).
As a result, linear probing with source-labeled data
picks up these source-only features, 
leaving considerable room for 
improvement on OOD data with further self-training.
On the other hand, in the SSL setting, 
the limited ID labeled data might provide enough signal 
to pick up features predictive on ID data, 
leaving little to no room for improvement for further self-training. 
Corroborating our intuitions, throughout the CL training in the toy setup, when CL doesn't achieve near-perfect generalization, 
the improvements provided by STOC for each checkpoint remain minimal.
On the other hand, for UDA setup, after reaching a certain training checkpoint 
in CL, STOC yields significant improvement (\figref{fig:analysis-empirical}(b)). 

In the next sections, we formalize our intuitions and analyze 
why ST and CL offer complementary benefits when dealing with distribution shifts.
Formal statements and proofs are in \appref{appsec:proofs}.

\subsection{Conditions for Success and Failure of Self-training over ERM from Scratch}

\label{subsec:self-training-analysis}

In our results on Example~\ref{ex:toy_setup}, 
we observe that 
performing ST after ERM 
yields a classifier with near-random target accuracy. 
In \thmref{thm:scratch_training}, we 
characterize conditions under which 
ST fails and succeeds. %

\begin{theorem}[Informal; Conditions for success and failure of ST over ERM]\label{thm:scratch_training}
    The target accuracy of ERM classifier,
    is given by $0.5\cdot \erfc\left( -\nicefrac{\gamma^2}{\paren{\sqrt{2 \dsp}\cdot \sigmasp}} \right)$. 
    Then ST performed in the second stage yields:
    (i) a classifier with $\approx 0.5$ target accuracy when $\gamma < \nicefrac{1}{2\sigmasp}$ and $\sigmasp \ge 1$; and (ii) a classifier with near-perfect target accuracy when $\gamma \ge \sigmasp$.
\end{theorem}

The informal theorem above abstracts the exact dependency of $\gamma, \sigmasp$, and $\dsp$ for the success and failure of ST over ERM. 
Our analysis highlights that while ERM learns a perfect predictor along  
$\winv$ (with norm $\gamma$), it also learns to depend on $\wspu$ (with norm $\sqrt{\dsp}$) 
because of the perfect correlation of $\xsp$ with labels on the source. Our conditions depict that when the $\gamma/\sigmasp$ is sufficiently small, then ST continues to erroneously enhance its reliance on the $\xsp$ feature for target prediction, resulting in near-random target performance. 
Conversely, when $\gamma/\sigmasp$ is larger than 1, the signal in $\xin$ is correctly used for predictor on the majority of target points, and ST eliminates the $\xsp$ dependency, converging to an optimal target classifier. 

Our proof analysis shows that if the ratio of the norm of the classifier along in the direction of $\wstar$ is smaller than $\wspu$ by a certain ratio then the generated pseudolabels (incorrectly) use $\xsp$ for its prediction further increasing the component along $\wspu$.  Moreover, normalization further diminishes the reliance along $\wstar$, culminating in a near-random performance. The opposite occurs when the ERM classifier achieves a signal along $\wstar$ that is sufficiently stronger than along $\wspu$.
Upon substituting the parameters used in Example~\ref{ex:toy_setup}, the ERM and ST performances as determined by \thmref{thm:scratch_training} align with our empirical results, notably, ST performance on target being near-random.

\subsection{CL Captures Both Features But Amplifies Invariant Over Spurious Features}
\label{subsec:contrastive-pretraining-analysis}

Here we show that minimizing the contrastive loss~\eqref{eq:cont-loss} on unlabeled data from both $\ProbS$ and $\ProbT$ gives us a feature extractor $\Phi_\cl$ that has a higher inner product with the invariant feature over the spurious feature.   
First, we derive a closed form expression for $\Phi_\cl$ that holds for any linear backbone and  augmentation distribution. Then, we introduce assumptions on the augmentation distribution (or equivalently on $\wstar$) and other problem parameters, that are sufficient to prove amplification. 

\begin{proposition}[Barlow Twins solution]
\label{prp:bt-closedform}
The solution for \eqref{eq:cont-loss} is $U_k^\top\Sigma_\mathsf{A}^{-1/2}$ where $U_k$ are the top $k$ eigenvectors of $\Sigma_\mathsf{A}^{-1/2}\, \tilde{\Sigma} \, \Sigma_\mathsf{A}^{-1/2}$. Here,  $\Sigma_\mathsf{A} \coloneqq \Exp_{a\sim\ProbA}[aa^\top]$ is the covariance over augmentations, and $\tilde{\Sigma} \coloneqq \Exp_{x\sim\ProbU} [\tilde{a}(x) \tilde{a}(x)^\top]$ is the covariance matrix of mean augmentations $\tilde{a}(x) \coloneqq \Exp_{\ProbA(a \mid x)}[a]$.
\end{proposition}
The above result captures the effect of augmentations through the matrix $U_k$. If there were no augmentations, then $ \Sigma_\mathsf{A} = \tilde{\Sigma}$, implying that $U_k$ could then be any random orthonormal matrix.
On the other hand if augmentation distributions change prevalent covariances in the data, \ie, $\Sigma_\mathsf{A}$ is very different from $\tilde{\Sigma}$, the matrix $U_k$ would bias the CL solution towards directions that capture significant variance in marginal distribution on augmented data, but have low conditional variance, when conditioned on original point $x$---precisely the directions with low invariance loss. 
Hence, we can expect that CL would learn components along both invariant $\winv$ and spurious $\wspu$ because: (i)  these directions explain a large fraction of variance in the raw data; (ii) augmentations that randomly scale down dimensions would add little variance along $\wspu$ and $\winv$ compared to noise  directions in their null space. On the other hand it is unclear which of these directions is amplified more in $\Phi_\cl$. The following assumption and amplification result conveys that when the noise in target $(\sigmasp)$ is suficiently large, the CL solution amplifies the invariant feature over the spurious feature.

\begin{assumption}[Informal; Alignment of $\wstar$ with augmentations] 
\label{assm:augs}
We assume that $\wstar$ aligns with  $\ProbA(\cdot \mid x)$, \ie,  $\forall x$, $\Exp_{a \mid x}[a^\top \wstar] = \nicefrac{1}{2} \cdot x^\top\mathrm{diag}(\mathbbm{1}_d) \wstar$ is high. Hence, we assume $w^\star = \nicefrac{\mathbbm{1}_{\din}}{\sqrt{\din}}$.
\end{assumption}

One implication of Assumption~\ref{assm:augs} is that when $w^\star = \nicefrac{\mathbbm{1}_{\din}}{\sqrt{\din}}$, only the top two eigenvectors lie in the space spanned by  $\winv$ and $\wspu$.
To analyze our amplification with fewer eigenvectors from Proposition~\ref{prp:bt-closedform} while retaining all relevant phenomena, we assume $\wstar = \nicefrac{\mathbbm{1}_{\din}}{\sqrt{\din}}$ for mathematical convenience.
While Assumption~\ref{assm:augs} permits a tighter theoretical analysis, our empirical results in \secref{subsec:intuitive-story} hold more generally for  $\wstar \sim \mathcal{N}(0, \mathbf{I}_{\din})$.

\begin{theorem}[Informal; CL recovers both $\winv$ and $\wspu$ but amplifies $\winv$]
\label{thm:bt-blockform}
Under Assumption~\ref{assm:augs}, the CL solution $\Phi_\cl$$=$$\brck{\phi_1, \phi_2, ..., \phi_k}$ satisfies $\phi_j^\top \winv = \phi_j^\top 
\wspu=0$ $\forall j\geq 3$, $\phi_1 = c_1 \winv + c_3 \wspu$ and $\phi_2 = c_2 \winv + c_4 \wspu$. 
For constants $K_1, K_2 >0$, $\gamma = \nicefrac{K_1K_2}{\sigmasp}$, $\dsp = \nicefrac{\sigmasp^2}{K_2^2}$, $\forall \epsilon > 0$, $\exists {\sigmasp}_{0}$, such that for $\sigmasp \geq {\sigmasp}_{0}$, $\abs{\nicefrac{c_1}{c_3}-\nicefrac{K_1K_2^2\din}{2L\sigmain^2 ({\din}-1)}} \leq \epsilon$, and $\abs{\abs{\nicefrac{c_2}{c_4}}- \nicefrac{L\sqrt{\dsp}}{\gamma}} \leq \epsilon$, where $L = {1+K_2^2}$.
\end{theorem}

We analyze the amplification of $\winv/\wspu$ with contrastive learning in the regime where $\sigmasp$ is large enough. In other words, if the target distribution has sufficient noise along the spurious feature, the augmentations prevent the CL solution from extracting components along $\wspu$.
Thus, in our analysis, we first analyze the amplification factors asymptotically $(\sigmasp \rightarrow \infty)$, 
and then use the asymptotic behavior to draw conclusions for the regime where $\sigmasp$ is large but finite.

\thmref{thm:bt-blockform} conveys two results: (i) CL recovers components along both $\winv$ and $\wspu$ through $\phi_1, \phi_2$; and (ii) it increases the norm along $\winv$ more than $\wspu$. The latter is evident because the margin separating labeled points along $\winv$ is now amplified by a factor of $|\nicefrac{c_2}{c_4}| = \Omega(\nicefrac{L\sqrt{d_\mathrm{sp}}}{\gamma})$ in $\phi_2$.
Naturally, this will improve the target performance of a linear predictor trained over CL representations. At the same time, we also see that in $\phi_1$, the component along $\wspu$ is still significant ($\nicefrac{c_1}{c_3} = \mathcal{O}(\nicefrac{1}{L \sigmain^2})$). 
Intuitively, CL prefers the invariant feature since augmentations amplify the noise along $\wspu$ in the target domain.  
At the same time, the variance induced by augmentations along $\wspu$ in source is still very small due to which the dependence on $\wspu$ is not completely alleviated.
Due to the remaining components along $\wspu$, the target performance for CL can remain less than ideal. Both the above arguments on target performance are captured in Corollary~\ref{corollary:BT_ERM}.

\begin{corollary}[Informal; CL improves OOD error over ERM but is still imperfect] \label{corollary:BT_ERM}
For $\gamma, \sigmasp, \dsp$ defined as in Theorem~\ref{thm:bt-blockform}, $\exists {\sigmasp}_{1}$ such that for all $\sigmasp \geq {\sigmasp}_{1},$ the target accuracy of CL (linear predictor on $\Phi_\cl$) is $\geq 0.5 \erfc\paren{-{L'} \cdot \nicefrac{\gamma}{\sqrt{2}\sigmasp}}$ and $\leq 0.5 \erfc\paren{-4L' \cdot \nicefrac{\gamma}{\sqrt{2}\sigmasp}}$, where $L' = \nicefrac{K_2^2 K_1}{\sigmain^2 (1-\nicefrac{1}{\din})}$. 
When ${\sigmasp}_1$ $>$ $\sigmain \sqrt{1-\nicefrac{1}{\din}}$, the lower bound on accuracy is strictly better than ERM from scratch.
\end{corollary}
While $\Phi_\cl$ is still not ideal for linear probing, in the next part we will see how $\Phi_\cl$ can instead be sufficient for subsequent self-training to unlearn the remaining components along spurious features.

\subsection{Improvements with Self-training Over Contrastive Learning}
\label{subsec:combining-self-training-and-contrastive-pretraining}

The result in the previous section highlights that 
while CL may improve over ERM, 
the linear probe continues to depend on the spurious feature. 
Next, we 
characterize the behavior STOC. 
Recall, in the ST stage, we iteratively 
update the linear head with \eqref{eq:self-training}
starting with the CL backbone and head.  

\begin{theorem}[Informal; ST improves over CL] \label{thm:SToverCL}
Under the conditions of \thmref{thm:bt-blockform} and $\dsp \le K_1^2 \cdot K_2^{2/3}$, 
the target accuracy of ST over CL is lower bounded by $0.5\cdot \erfc\left({-\abs{\nicefrac{c2}{c4}}\cdot \nicefrac{\gamma}{(\sqrt{2} \sigma_2)}}\right) \approx 0.5\cdot \erfc\left(-L\nicefrac{\sqrt{\dsp}}{(\sqrt{2} \sigmasp)}\right)$ where $c_2$ and $c_4$ are the coefficients of feature $\phi_2$ along $\winv$ and $\wspu$ learned by BT.  
\end{theorem}
The above theorem states that 
when $\nicefrac{\sqrt{\dsp}}{\sigmasp} \gg 1$ 
the target accuracy of ST over CL is close to 1. 
In Example~\ref{ex:toy_setup}, the lower bound of the 
accuracy of ST over CL is $\erfc\left({-\sqrt{10}}\right) \approx 2$ 
showing near-perfect target generalization. 
Recall that \thmref{corollary:BT_ERM} shows that CL yields a 
linear head that mainly depends on both the invariant direction $\winv$ 
and the spurious direction $\wspu$. 
At initialization, the linear head trained on the CL backbone has negligible dependence on $\phi_2$ (under conditions in \thmref{corollary:BT_ERM}). 
Building on that, the analysis in \thmref{thm:SToverCL}  
captures that ST gradually reduces the dependence on $\wspu$ by learning 
a linear head that has a larger reliance on $\phi_2$, which has a higher ``effective'' margin on 
the target, thus increasing overall dependency on $\winv $.

\textbf{Theoretical comparison with SSL.~~} Our analysis until now shows that
linear probing with source labeled data during CL  picks up features that are more predictive of source label under distribution shift, leaving a significant room for improvement on OOD data when self-trained further. In UDA, the primary benefit of ST lies in picking up the features with a high ``effective'' margin on target data that are not picked up by linear head trained during CL.
In contrast, in the SSL setting, the limited ID labeled data may provide enough signal in picking up high-margin features that are predictive on ID data, leaving little to no room for improvement for further ST. We formalize this intuition in \appref{appsec:proofs} when the CL/ERM predictors are trained with margin based surrogate losses for learning the classifier.

\subsection{Reconciling Practice: Implications for Deep Non-Linear Networks }
\label{subsec:deep-networks}

In this section, we experiment with deep non-linear backbone (\ie, $\phicl$). When we continue to fix $\phicl$ during CL and STOC, the trends we observed with linear networks in \secref{subsec:intuitive-story} continue to hold. We then perform full fine-tuning with CL and STOC, i.e.,  propagate gradients even to $\phicl$, as commonly done in practice. 
We present key takeaways here but detailed experiments are   
in \appref{app:deep_toy}.

\textbf{Benefits of augmentation for self-training.~~}
ST while updating $\phicl$ can hurt due to overfitting issues when training with the finite sample of labeled and unlabeled data (drop by >10\% over CL).
This is due to the ability of deep networks to overfit on 
confident but incorrect pseudolabels on target data~\citep{zhang2016understanding}. 
This exacerbates components along $\wspu$ and we find that augmentations (and other heuristics) typically used in practice (\eg in FixMatch~\cite{sohn2020fixmatch}) help avoid overfitting on incorrect pseudolabels.

\textbf{Can ERM and ST over contrastive pretraining improve features?~~} 
We find that self-training can also slightly improve features when we update the backbone with the second stage of STOC and when the CL backbone is early stopped sub-optimally (\ie at an earlier checkpoint in \figref{fig:analysis-empirical}(b)).
This feature finetuning can now widen the gap between STOC and CL in SSL settings, as compared to the linear probing gap (as in  \ref{fig:analysis-empirical}). 
This is because STOC can now improve performance beyond just recovering the generalization gap for the linear head (which is typically small). However, STOC benefits are negligible when CL is not early stopped sub-optimally, \ie, trained till convergence. 
Thus, it remains unclear if STOC and CL have complementary benefits for feature learning in UDA or SSL settings. Investigating this is an interesting avenue for future work.

\vspace{-5pt}
\section{Connecting Experimental Gains with Theoretical Insights} \label{sec:discussion}
\vspace{-5pt}
\begin{figure}
  \centering
  \includegraphics[width=0.5\linewidth]{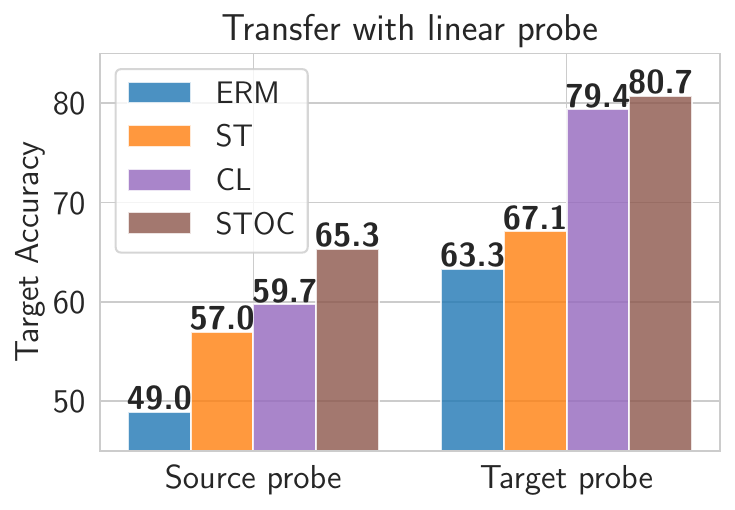}
  \caption{\emph{Target accuracy with source and target linear probes}, which freezes backbones trained with various objectives and trains only the head in UDA setup. Avg. accuracy across all datasets. We observe that: (i) ST improves the linear transferability of source probes, and (ii)  CL improves representations.\vspace{-16pt}}
  \label{fig:LP-transfer}
\end{figure}

Our theory emphasizes that under distribution shift %
contrastive pretraining does feature amplification which effectively improves the representations for target data, while self-training primarily improves linear transferability for the classifier learned on top of CL features. 
To investigate different methods in our UDA setup, we study the representations learned by each of them. We fix the representations and train linear heads over them to answer two questions: 
(i) How good are the representations in terms of their \emph{ceiling} of target accuracy (performance of the optimal linear probe)?---we evaluate this by training the classifier head on target labeled data (\ie, target linear probe); and
(ii) How well do heads trained on source generalize to target?---we assess this by training a head on source labeled data (source linear probe) and evaluate its difference with target linear probe. For both, we plot target accuracy. We make two \emph{intriguing} observations
\figref{fig:LP-transfer}):

\textbf{Does CL improve representations over ERM features?} Yes.
We observe a substantial difference in accuracy ($\approx 14\%$ gap)
of target linear probes on backbones trained with contrastive pretraining 
(\ie CL, STOC) 
and without it (\ie, ERM, ST) highlighting that CL significantly pushes 
the performance ceiling over non-contrastive features.
As a side, our findings also stand in contrast to recent studies
suggesting that ERM features might 
be ``good enough'' for OOD generalization~\citep{rosenfeld2022domain, kirichenko2022last}. 
Instead, the observed gains with contrastively pretrained backbones (\ie CL, STOC)
demonstrate that target unlabeled data can be leveraged to further improve over ERM features.

\textbf{Do CL features yield \emph{perfect} linear transferability from source to target?~~}
Recent works~\citep{haochen2022beyond, shen2022connect} conjecture that under certain conditions CL representations, linear probes learned with source labeled data may transfer perfectly from source to target. 
However, we observe that this doesn't hold strictly in practice, and in fact, the linear transferability can be further improved with ST. 
We first note a significant gap between the performance of source linear probes
and target linear probes illustrating that linear transferability is not perfect in practice. 
Moreover, while the accuracy of target linear probes doesn't change substantially 
between CL and STOC, the accuracy of the source linear probe improves significantly. 
Similar observations hold for ERM and ST, methods trained without contrastive pretraining. 
This highlights that ST performs ``feature refinement'' to improve source to target linear transfer
(with relatively small improvements in their respective target probe performance).
\emph{The findings highlight the complementary nature of benefits on real-world data: ST improves linear transferability while CL improves representations.}

\vspace{-5pt}
\section{Connections to Prior Work} %
\vspace{-5pt}

Our empirical results and our analyses offer 
a perspective that contrasts with the prior literature that argues for the individual optimality of contrastive pretraining and self-training.  We outline the key differences from existing studies here, and delve into other related works in \appref{appsec:other-relwork}.

\textbf{Limitations of prior work analyzing contrastive learning {} {}} 
Prior works~\citep{haochen2022beyond,shen2022connect} analyzing CL first make assumptions on the consistency of augmentations with labels~\citep{haochen2021provable,cabannes2023ssl,saunshi2022understanding,johnson2022contrastive}, and specifically for UDA make stronger ones on the augmentation graph connecting 
examples from same domain or class more than cross-class/cross-domain ones.
While this is sufficient to prove linear transferability, it is unclear if this holds in practice when augmentations are imperfect, \ie if they fail to mask the spurious features completely---as corroborated by our findings in \secref{sec:discussion}. We show why this also fails in our simplified setup in \appref{app:CL_prior_work_assumption}.

\textbf{Limitations of prior work analyzing self-training {} {}} 
Prior research views self-training as consistency regularization, ensuring pseudolabels for original samples align with their augmentations~\citep{cai2021theory, wei2020theoretical, sohn2020fixmatch}. This approach abstracts the role played by the optimization algorithm and instead evaluates the global minimizer of a population objective promoting pseudolabel consistency. It also relies on specific assumptions about class-conditional distributions to guarantee pseudolabel accuracy across domains. However, this framework doesn't address issues in iterative label propagation. 
For example, when augmentation distribution has long tails, the consistency of pseudolabels depends on the sampling frequency of ``favorable'' augmentations (for more discussion see \appref{app:iterative-propagation}).
Our analysis thus follows the iterative examination of self-training~\cite{chen2020self}.

\vspace{-5pt}
\section{Conclusion}
\vspace{-5pt}

In this study, we highlight the synergistic behavior of 
self-training improving linear transferability and contrastive pretraining learning more ``invariant'' features under distribution shift. Shifts in distribution are commonplace in real-world applications of machine learning, and even under natural, non-adversarial distribution shifts, the performance of machine learning models often drops. By simply combining existing techniques in self-training and constrastive learning, we find that we can improve accuracy by 3--8\%  rather than using either approach independently. Despite these significant improvements, we note that one limitation of this combined approach is that performing self-training sequentially after contrastive 
pretraining increases the computation cost for UDA. 
The potential for integrating these benefits into one 
unified training paradigm is yet unclear, 
presenting an interesting direction for future exploration.

Beyond this, we note that our theoretical framework primarily confines the analysis to 
training the backbone and linear network 
independently during the pretraining and 
fine-tuning/self-training phases. Although our empirical observations 
apply to deep networks with full fine-tuning, 
we leave a more rigorous theoretical study of full fine-tuning for future work. 
Our theory also relies on a covariate shift 
assumption (where we assume that label distribution also doesn't shift). 
Investigating the complementary nature of self-training 
and contrastive pretraining beyond the covariate shift assumption would be another interesting direction for future work.

\subsection*{Acknowledgements}

SG acknowledges the JP Morgan AI Ph.D. Fellowship and  Bloomberg Ph.D. Fellowship for their support.  AR acknowledges support from Open
Philanthropy, Google, Apple and Schmidt AI2050 Early Career Fellowship.

\bibliographystyle{apalike}
\bibliography{domain_adaptation}

\begin{thebibliography}{}

\bibitem[Alexandari et~al., 2021]{alexandari2019adapting}
Alexandari, A., Kundaje, A., and Shrikumar, A. (2021).
\newblock Adapting to label shift with bias-corrected calibration.
\newblock In {\em International Conference on Machine Learning (ICML)}.

\bibitem[Arora et~al., 2019]{arora2019theoretical}
Arora, S., Khandeparkar, H., Khodak, M., Plevrakis, O., and Saunshi, N. (2019).
\newblock A theoretical analysis of contrastive unsupervised representation learning.
\newblock {\em arXiv preprint arXiv:1902.09229}.

\bibitem[Azizzadenesheli et~al., 2019]{azizzadenesheli2019regularized}
Azizzadenesheli, K., Liu, A., Yang, F., and Anandkumar, A. (2019).
\newblock Regularized learning for domain adaptation under label shifts.
\newblock In {\em International Conference on Learning Representations (ICLR)}.

\bibitem[Bardes et~al., 2021]{bardes2021vicreg}
Bardes, A., Ponce, J., and LeCun, Y. (2021).
\newblock Vicreg: Variance-invariance-covariance regularization for self-supervised learning.
\newblock {\em arXiv preprint arXiv:2105.04906}.

\bibitem[Baricz, 2008]{baricz2008mills}
Baricz, {\'A}. (2008).
\newblock Mills' ratio: Monotonicity patterns and functional inequalities.
\newblock {\em Journal of Mathematical Analysis and Applications}, 340(2):1362--1370.

\bibitem[Bekker and Davis, 2020]{pusurvey}
Bekker, J. and Davis, J. (2020).
\newblock Learning from positive and unlabeled data: a survey.
\newblock {\em Machine Learning}.

\bibitem[Ben-David et~al., 2010]{ben2010impossibility}
Ben-David, S., Lu, T., Luu, T., and P{\'a}l, D. (2010).
\newblock {Impossibility Theorems for Domain Adaptation}.
\newblock In {\em International Conference on Artificial Intelligence and Statistics (AISTATS)}.

\bibitem[Berthelot et~al., 2019]{berthelot2019remixmatch}
Berthelot, D., Carlini, N., Cubuk, E.~D., Kurakin, A., Sohn, K., Zhang, H., and Raffel, C. (2019).
\newblock Remixmatch: Semi-supervised learning with distribution alignment and augmentation anchoring.
\newblock {\em arXiv preprint arXiv:1911.09785}.

\bibitem[Bishop, 2006]{bishop2006pattern}
Bishop, C.~M. (2006).
\newblock {\em {Pattern Recognition and Machine Learning}}.
\newblock Springer.

\bibitem[Blanchard et~al., 2011]{blanchard2011generalizing}
Blanchard, G., Lee, G., and Scott, C. (2011).
\newblock Generalizing from several related classification tasks to a new unlabeled sample.
\newblock {\em Advances in neural information processing systems}, 24.

\bibitem[Cabannes et~al., 2023]{cabannes2023ssl}
Cabannes, V., Kiani, B.~T., Balestriero, R., LeCun, Y., and Bietti, A. (2023).
\newblock The ssl interplay: Augmentations, inductive bias, and generalization.
\newblock {\em arXiv preprint arXiv:2302.02774}.

\bibitem[Cai et~al., 2021]{cai2021theory}
Cai, T., Gao, R., Lee, J., and Lei, Q. (2021).
\newblock A theory of label propagation for subpopulation shift.
\newblock In {\em International Conference on Machine Learning}, pages 1170--1182. PMLR.

\bibitem[Caron et~al., 2020]{caron2020unsupervised}
Caron, M., Misra, I., Mairal, J., Goyal, P., Bojanowski, P., and Joulin, A. (2020).
\newblock Unsupervised learning of visual features by contrasting cluster assignments.
\newblock {\em Advances in Neural Information Processing Systems}, 33:9912--9924.

\bibitem[Caron et~al., 2021]{caron2021emerging}
Caron, M., Touvron, H., Misra, I., J{\'e}gou, H., Mairal, J., Bojanowski, P., and Joulin, A. (2021).
\newblock Emerging properties in self-supervised vision transformers.
\newblock In {\em Proceedings of the IEEE/CVF international conference on computer vision}, pages 9650--9660.

\bibitem[Chapelle et~al., 2006]{chapelle2006semi}
Chapelle, O., Scholkopf, B., and Zien, A. (2006).
\newblock Semi-supervised learning. 2006.
\newblock {\em Cambridge, Massachusettes: The MIT Press View Article}, 2.

\bibitem[Chen et~al., 2020a]{chen2020simple}
Chen, T., Kornblith, S., Norouzi, M., and Hinton, G. (2020a).
\newblock A simple framework for contrastive learning of visual representations.
\newblock In {\em International conference on machine learning}, pages 1597--1607. PMLR.

\bibitem[Chen et~al., 2020b]{chen2020self}
Chen, X., Chen, W., Chen, T., Yuan, Y., Gong, C., Chen, K., and Wang, Z. (2020b).
\newblock Self-pu: Self boosted and calibrated positive-unlabeled training.
\newblock In {\em International Conference on Machine Learning}, pages 1510--1519. PMLR.

\bibitem[Christie et~al., 2018]{christie2018functional}
Christie, G., Fendley, N., Wilson, J., and Mukherjee, R. (2018).
\newblock Functional map of the world.
\newblock In {\em Proceedings of the IEEE Conference on Computer Vision and Pattern Recognition}.

\bibitem[Cortes et~al., 2010]{cortes2010learning}
Cortes, C., Mansour, Y., and Mohri, M. (2010).
\newblock {Learning Bounds for Importance Weighting}.
\newblock In {\em Advances in Neural Information Processing Systems (NIPS)}.

\bibitem[Cortes and Mohri, 2014]{cortes2014domain}
Cortes, C. and Mohri, M. (2014).
\newblock Domain adaptation and sample bias correction theory and algorithm for regression.
\newblock {\em Theoretical Computer Science}, 519.

\bibitem[Cubuk et~al., 2020]{cubuk2020randaugment}
Cubuk, E.~D., Zoph, B., Shlens, J., and Le, Q.~V. (2020).
\newblock Randaugment: Practical automated data augmentation with a reduced search space.
\newblock In {\em Proceedings of the IEEE/CVF conference on computer vision and pattern recognition workshops}, pages 702--703.

\bibitem[Darlow et~al., 2018]{darlow2018cinic}
Darlow, L.~N., Crowley, E.~J., Antoniou, A., and Storkey, A.~J. (2018).
\newblock Cinic-10 is not imagenet or cifar-10.
\newblock {\em arXiv preprint arXiv:1810.03505}.

\bibitem[Deledalle et~al., 2017]{deledalle2017closed}
Deledalle, C.-A., Denis, L., Tabti, S., and Tupin, F. (2017).
\newblock {\em Closed-form expressions of the eigen decomposition of 2 x 2 and 3 x 3 Hermitian matrices}.
\newblock PhD thesis, Universit{\'e} de Lyon.

\bibitem[DeVries and Taylor, 2017]{devries2017improved}
DeVries, T. and Taylor, G.~W. (2017).
\newblock Improved regularization of convolutional neural networks with cutout.
\newblock {\em arXiv preprint arXiv:1708.04552}.

\bibitem[Elkan and Noto, 2008]{elkan2008learning}
Elkan, C. and Noto, K. (2008).
\newblock Learning classifiers from only positive and unlabeled data.
\newblock In {\em International Conference Knowledge Discovery and Data Mining (KDD)}, pages 213--220.

\bibitem[Ganin et~al., 2016]{ganin2016domain}
Ganin, Y., Ustinova, E., Ajakan, H., Germain, P., Larochelle, H., Laviolette, F., Marchand, M., and Lempitsky, V. (2016).
\newblock Domain-adversarial training of neural networks.
\newblock {\em The journal of machine learning research}.

\bibitem[Gardner et~al., 2018]{gardner2018gpytorch}
Gardner, J., Pleiss, G., Weinberger, K.~Q., Bindel, D., and Wilson, A.~G. (2018).
\newblock Gpytorch: Blackbox matrix-matrix gaussian process inference with gpu acceleration.
\newblock In {\em Advances in Neural Information Processing Systems (NeurIPS)}.

\bibitem[Garg et~al., 2022a]{garg2022OSLS}
Garg, S., Balakrishnan, S., and Lipton, Z. (2022a).
\newblock Domain adaptation under open set label shift.
\newblock In {\em Advances in Neural Information Processing Systems (NeurIPS)}.

\bibitem[Garg et~al., 2022b]{garg2022ATC}
Garg, S., Balakrishnan, S., Lipton, Z., Neyshabur, B., and Sedghi, H. (2022b).
\newblock Leveraging unlabeled data to predict out-of-distribution performance.
\newblock In {\em International Conference on Learning Representations (ICLR)}.

\bibitem[Garg et~al., 2023]{garg2022RLSbench}
Garg, S., Erickson, N., Sharpnack, J., Smola, A., Balakrishnan, S., and Lipton, Z. (2023).
\newblock Rlsbench: A large-scale empirical study of domain adaptation under relaxed label shift.
\newblock In {\em International Conference on Machine Learning (ICML)}.

\bibitem[Garg et~al., 2020]{garg2020labelshift}
Garg, S., Wu, Y., Balakrishnan, S., and Lipton, Z. (2020).
\newblock A unified view of label shift estimation.
\newblock In {\em Advances in Neural Information Processing Systems (NeurIPS)}.

\bibitem[Garg et~al., 2021]{garg2021PUlearning}
Garg, S., Wu, Y., Smola, A., Balakrishnan, S., and Lipton, Z. (2021).
\newblock Mixture proportion estimation and {PU} learning: A modern approach.
\newblock In {\em Advances in Neural Information Processing Systems (NeurIPS)}.

\bibitem[Garrido et~al., 2022]{garrido2022duality}
Garrido, Q., Chen, Y., Bardes, A., Najman, L., and Lecun, Y. (2022).
\newblock On the duality between contrastive and non-contrastive self-supervised learning.
\newblock {\em arXiv preprint arXiv:2206.02574}.

\bibitem[Grandvalet and Bengio, 2006]{grandvalet2006entropy}
Grandvalet, Y. and Bengio, Y. (2006).
\newblock Entropy regularization.

\bibitem[Gretton et~al., 2009]{gretton2009covariate}
Gretton, A., Smola, A.~J., Huang, J., Schmittfull, M., Borgwardt, K.~M., and Sch{\"o}lkopf, B. (2009).
\newblock {Covariate Shift by Kernel Mean Matching}.
\newblock {\em Journal of Machine Learning Research (JMLR)}.

\bibitem[Grill et~al., 2020]{grill2020bootstrap}
Grill, J.-B., Strub, F., Altch{\'e}, F., Tallec, C., Richemond, P., Buchatskaya, E., Doersch, C., Avila~Pires, B., Guo, Z., Gheshlaghi~Azar, M., et~al. (2020).
\newblock Bootstrap your own latent-a new approach to self-supervised learning.
\newblock {\em Advances in neural information processing systems}, 33:21271--21284.

\bibitem[Gulrajani and Lopez-Paz, 2020]{gulrajani2020search}
Gulrajani, I. and Lopez-Paz, D. (2020).
\newblock In search of lost domain generalization.
\newblock {\em arXiv preprint arXiv:2007.01434}.

\bibitem[HaoChen and Ma, 2022]{haochen2022theoretical}
HaoChen, J.~Z. and Ma, T. (2022).
\newblock A theoretical study of inductive biases in contrastive learning.
\newblock {\em arXiv preprint arXiv:2211.14699}.

\bibitem[HaoChen et~al., 2021]{haochen2021provable}
HaoChen, J.~Z., Wei, C., Gaidon, A., and Ma, T. (2021).
\newblock Provable guarantees for self-supervised deep learning with spectral contrastive loss.
\newblock {\em Advances in Neural Information Processing Systems}, 34:5000--5011.

\bibitem[HaoChen et~al., 2022]{haochen2022beyond}
HaoChen, J.~Z., Wei, C., Kumar, A., and Ma, T. (2022).
\newblock Beyond separability: Analyzing the linear transferability of contrastive representations to related subpopulations.
\newblock {\em arXiv preprint arXiv:2204.02683}.

\bibitem[He et~al., 2020]{he2020momentum}
He, K., Fan, H., Wu, Y., Xie, S., and Girshick, R. (2020).
\newblock Momentum contrast for unsupervised visual representation learning.
\newblock In {\em Proceedings of the IEEE/CVF conference on computer vision and pattern recognition}, pages 9729--9738.

\bibitem[He et~al., 2016]{he2016deep}
He, K., Zhang, X., Ren, S., and Sun, J. (2016).
\newblock {Deep Residual Learning for Image Recognition}.
\newblock In {\em Computer Vision and Pattern Recognition (CVPR)}.

\bibitem[Joachims et~al., 1999]{joachims1999transductive}
Joachims, T. et~al. (1999).
\newblock Transductive inference for text classification using support vector machines.
\newblock In {\em Icml}, volume~99, pages 200--209.

\bibitem[Johnson et~al., 2022]{johnson2022contrastive}
Johnson, D.~D., Hanchi, A.~E., and Maddison, C.~J. (2022).
\newblock Contrastive learning can find an optimal basis for approximately view-invariant functions.
\newblock {\em arXiv preprint arXiv:2210.01883}.

\bibitem[Kakade et~al., 2008]{kakade2008complexity}
Kakade, S.~M., Sridharan, K., and Tewari, A. (2008).
\newblock On the complexity of linear prediction: Risk bounds, margin bounds, and regularization.
\newblock {\em Advances in neural information processing systems}, 21.

\bibitem[Kirichenko et~al., 2022]{kirichenko2022last}
Kirichenko, P., Izmailov, P., and Wilson, A.~G. (2022).
\newblock Last layer re-training is sufficient for robustness to spurious correlations.
\newblock {\em arXiv preprint arXiv:2204.02937}.

\bibitem[Koh et~al., 2021]{wilds2021}
Koh, P.~W., Sagawa, S., Marklund, H., Xie, S.~M., Zhang, M., Balsubramani, A., Hu, W., Yasunaga, M., Phillips, R.~L., Gao, I., Lee, T., David, E., Stavness, I., Guo, W., Earnshaw, B.~A., Haque, I.~S., Beery, S., Leskovec, J., Kundaje, A., Pierson, E., Levine, S., Finn, C., and Liang, P. (2021).
\newblock {WILDS}: A benchmark of in-the-wild distribution shifts.
\newblock In {\em International Conference on Machine Learning (ICML)}.

\bibitem[Krizhevsky and Hinton, 2009]{krizhevsky2009learning}
Krizhevsky, A. and Hinton, G. (2009).
\newblock {Learning Multiple Layers of Features from Tiny Images}.
\newblock Technical report, Citeseer.

\bibitem[Kschischang, 2017]{kschischang2017complementary}
Kschischang, F.~R. (2017).
\newblock The complementary error function.
\newblock {\em Online, April}.

\bibitem[Kumar et~al., 2020]{kumar2020understanding}
Kumar, A., Ma, T., and Liang, P. (2020).
\newblock Understanding self-training for gradual domain adaptation.
\newblock In {\em International Conference on Machine Learning}, pages 5468--5479. PMLR.

\bibitem[Kumar et~al., 2022]{kumar2022finetuning}
Kumar, A., Raghunathan, A., Jones, R.~M., Ma, T., and Liang, P. (2022).
\newblock Fine-tuning can distort pretrained features and underperform out-of-distribution.
\newblock In {\em International Conference on Learning Representations}.

\bibitem[Lee et~al., 2013]{lee2013pseudo}
Lee, D.-H. et~al. (2013).
\newblock Pseudo-label: The simple and efficient semi-supervised learning method for deep neural networks.
\newblock In {\em Workshop on challenges in representation learning, ICML}, volume~3, page 896.

\bibitem[Lipton et~al., 2018]{lipton2018detecting}
Lipton, Z.~C., Wang, Y.-X., and Smola, A. (2018).
\newblock {Detecting and Correcting for Label Shift with Black Box Predictors}.
\newblock In {\em International Conference on Machine Learning (ICML)}.

\bibitem[Long et~al., 2015]{long2015learning}
Long, M., Cao, Y., Wang, J., and Jordan, M. (2015).
\newblock Learning transferable features with deep adaptation networks.
\newblock In {\em International conference on machine learning}, pages 97--105. PMLR.

\bibitem[Long et~al., 2017]{long2017deep}
Long, M., Zhu, H., Wang, J., and Jordan, M.~I. (2017).
\newblock Deep transfer learning with joint adaptation networks.
\newblock In {\em International conference on machine learning}. PMLR.

\bibitem[Loshchilov and Hutter, 2016]{loshchilov2016sgdr}
Loshchilov, I. and Hutter, F. (2016).
\newblock Sgdr: Stochastic gradient descent with warm restarts.
\newblock {\em arXiv preprint arXiv:1608.03983}.

\bibitem[Ma et~al., 2021]{ma2021conditional}
Ma, M.~Q., Tsai, Y.-H.~H., Liang, P.~P., Zhao, H., Zhang, K., Salakhutdinov, R., and Morency, L.-P. (2021).
\newblock Conditional contrastive learning for improving fairness in self-supervised learning.
\newblock {\em arXiv preprint arXiv:2106.02866}.

\bibitem[Mishra et~al., 2021]{mishra2021surprisingly}
Mishra, S., Saenko, K., and Saligrama, V. (2021).
\newblock Surprisingly simple semi-supervised domain adaptation with pretraining and consistency.
\newblock {\em arXiv preprint arXiv:2101.12727}.

\bibitem[Muandet et~al., 2013]{muandet2013domain}
Muandet, K., Balduzzi, D., and Sch{\"o}lkopf, B. (2013).
\newblock Domain generalization via invariant feature representation.
\newblock In {\em International Conference on Machine Learning}, pages 10--18. PMLR.

\bibitem[Nagarajan et~al., 2020]{nagarajan2020understanding}
Nagarajan, V., Andreassen, A., and Neyshabur, B. (2020).
\newblock Understanding the failure modes of out-of-distribution generalization.
\newblock {\em arXiv preprint arXiv:2010.15775}.

\bibitem[Oord et~al., 2018]{oord2018representation}
Oord, A. v.~d., Li, Y., and Vinyals, O. (2018).
\newblock Representation learning with contrastive predictive coding.
\newblock {\em arXiv preprint arXiv:1807.03748}.

\bibitem[Peng et~al., 2019]{peng2019moment}
Peng, X., Bai, Q., Xia, X., Huang, Z., Saenko, K., and Wang, B. (2019).
\newblock Moment matching for multi-source domain adaptation.
\newblock In {\em Proceedings of the IEEE/CVF international conference on computer vision}, pages 1406--1415.

\bibitem[Peng et~al., 2017]{visda2017}
Peng, X., Usman, B., Kaushik, N., Hoffman, J., Wang, D., and Saenko, K. (2017).
\newblock Visda: The visual domain adaptation challenge.

\bibitem[Peng et~al., 2018]{peng2018syn2real}
Peng, X., Usman, B., Saito, K., Kaushik, N., Hoffman, J., and Saenko, K. (2018).
\newblock Syn2real: A new benchmark forsynthetic-to-real visual domain adaptation.
\newblock {\em arXiv preprint arXiv:1806.09755}.

\bibitem[Quinonero-Candela et~al., 2008]{quinonero2008dataset}
Quinonero-Candela, J., Sugiyama, M., Schwaighofer, A., and Lawrence, N.~D. (2008).
\newblock {\em Dataset shift in machine learning}.
\newblock Mit Press.

\bibitem[Roberts et~al., 2022]{roberts2022LLS}
Roberts, M., Mani, P., Garg, S., and Lipton, Z. (2022).
\newblock Unsupervised learning under latent label shift.
\newblock In {\em Advances in Neural Information Processing Systems (NeurIPS)}.

\bibitem[Rosenfeld et~al., 2022]{rosenfeld2022domain}
Rosenfeld, E., Ravikumar, P., and Risteski, A. (2022).
\newblock Domain-adjusted regression or: Erm may already learn features sufficient for out-of-distribution generalization.
\newblock {\em arXiv preprint arXiv:2202.06856}.

\bibitem[Russakovsky et~al., 2015]{russakovsky2015imagenet}
Russakovsky, O., Deng, J., Su, H., Krause, J., Satheesh, S., Ma, S., Huang, Z., Karpathy, A., Khosla, A., Bernstein, M., et~al. (2015).
\newblock Imagenet large scale visual recognition challenge.
\newblock {\em International journal of computer vision}, 115(3):211--252.

\bibitem[Saerens et~al., 2002]{saerens2002adjusting}
Saerens, M., Latinne, P., and Decaestecker, C. (2002).
\newblock {Adjusting the Outputs of a Classifier to New a Priori Probabilities: A Simple Procedure}.
\newblock {\em Neural Computation}.

\bibitem[Sagawa et~al., 2021]{sagawa2021extending}
Sagawa, S., Koh, P.~W., Lee, T., Gao, I., Xie, S.~M., Shen, K., Kumar, A., Hu, W., Yasunaga, M., Marklund, H., Beery, S., David, E., Stavness, I., Guo, W., Leskovec, J., Saenko, K., Hashimoto, T., Levine, S., Finn, C., and Liang, P. (2021).
\newblock Extending the wilds benchmark for unsupervised adaptation.
\newblock In {\em NeurIPS Workshop on Distribution Shifts}.

\bibitem[Sagawa et~al., 2020]{sagawa2020investigation}
Sagawa, S., Raghunathan, A., Koh, P.~W., and Liang, P. (2020).
\newblock An investigation of why overparameterization exacerbates spurious correlations.
\newblock In {\em International Conference on Machine Learning}, pages 8346--8356. PMLR.

\bibitem[Santurkar et~al., 2021]{santurkar2020breeds}
Santurkar, S., Tsipras, D., and Madry, A. (2021).
\newblock Breeds: Benchmarks for subpopulation shift.
\newblock In {\em International Conference on Learning Representations (ICLR)}.

\bibitem[Saunshi et~al., 2022]{saunshi2022understanding}
Saunshi, N., Ash, J., Goel, S., Misra, D., Zhang, C., Arora, S., Kakade, S., and Krishnamurthy, A. (2022).
\newblock Understanding contrastive learning requires incorporating inductive biases.
\newblock In {\em International Conference on Machine Learning}, pages 19250--19286. PMLR.

\bibitem[Sch{\"o}lkopf et~al., 2012]{scholkopf2012causal}
Sch{\"o}lkopf, B., Janzing, D., Peters, J., Sgouritsa, E., Zhang, K., and Mooij, J. (2012).
\newblock {On Causal and Anticausal Learning}.
\newblock In {\em International Conference on Machine Learning (ICML)}.

\bibitem[Scudder, 1965]{scudder1965probability}
Scudder, H. (1965).
\newblock Probability of error of some adaptive pattern-recognition machines.
\newblock {\em IEEE Transactions on Information Theory}, 11(3):363--371.

\bibitem[Shen et~al., 2022]{shen2022connect}
Shen, K., Jones, R.~M., Kumar, A., Xie, S.~M., HaoChen, J.~Z., Ma, T., and Liang, P. (2022).
\newblock Connect, not collapse: Explaining contrastive learning for unsupervised domain adaptation.
\newblock In {\em International Conference on Machine Learning}, pages 19847--19878. PMLR.

\bibitem[Shimodaira, 2000]{shimodaira2000improving}
Shimodaira, H. (2000).
\newblock {Improving Predictive Inference Under Covariate Shift by Weighting the Log-Likelihood Function}.
\newblock {\em Journal of Statistical Planning and Inference}.

\bibitem[Shu et~al., 2018]{shu2018dirt}
Shu, R., Bui, H.~H., Narui, H., and Ermon, S. (2018).
\newblock A dirt-t approach to unsupervised domain adaptation.
\newblock {\em arXiv preprint arXiv:1802.08735}.

\bibitem[Sohn et~al., 2020]{sohn2020fixmatch}
Sohn, K., Berthelot, D., Carlini, N., Zhang, Z., Zhang, H., Raffel, C.~A., Cubuk, E.~D., Kurakin, A., and Li, C.-L. (2020).
\newblock Fixmatch: Simplifying semi-supervised learning with consistency and confidence.
\newblock {\em Advances in Neural Information Processing Systems}, 33.

\bibitem[Stewart, 1993]{stewart1993early}
Stewart, G.~W. (1993).
\newblock On the early history of the singular value decomposition.
\newblock {\em SIAM review}, 35(4):551--566.

\bibitem[Sun et~al., 2017]{sun2017correlation}
Sun, B., Feng, J., and Saenko, K. (2017).
\newblock Correlation alignment for unsupervised domain adaptation.
\newblock In {\em Domain Adaptation in Computer Vision Applications}. Springer.

\bibitem[Sun and Saenko, 2016]{sun2016deep}
Sun, B. and Saenko, K. (2016).
\newblock Deep coral: Correlation alignment for deep domain adaptation.
\newblock In {\em European conference on computer vision}. Springer.

\bibitem[Torralba and Efros, 2011]{torralba2011unbiased}
Torralba, A. and Efros, A.~A. (2011).
\newblock Unbiased look at dataset bias.
\newblock In {\em CVPR 2011}, pages 1521--1528. IEEE.

\bibitem[Van~Engelen and Hoos, 2020]{van2020survey}
Van~Engelen, J.~E. and Hoos, H.~H. (2020).
\newblock A survey on semi-supervised learning.
\newblock {\em Machine learning}, 109(2):373--440.

\bibitem[Venkateswara et~al., 2017]{venkateswara2017deep}
Venkateswara, H., Eusebio, J., Chakraborty, S., and Panchanathan, S. (2017).
\newblock Deep hashing network for unsupervised domain adaptation.
\newblock In {\em Proceedings of the IEEE Conference on Computer Vision and Pattern Recognition}, pages 5018--5027.

\bibitem[Wainwright, 2019]{wainwright2019high}
Wainwright, M.~J. (2019).
\newblock {\em High-dimensional statistics: A non-asymptotic viewpoint}, volume~48.
\newblock Cambridge university press.

\bibitem[Wang et~al., 2021]{wang2021tent}
Wang, D., Shelhamer, E., Liu, S., Olshausen, B., and Darrell, T. (2021).
\newblock Tent: Fully test-time adaptation by entropy minimization.
\newblock In {\em International Conference on Learning Representations}.

\bibitem[Wei et~al., 2020]{wei2020theoretical}
Wei, C., Shen, K., Chen, Y., and Ma, T. (2020).
\newblock Theoretical analysis of self-training with deep networks on unlabeled data.
\newblock {\em arXiv preprint arXiv:2010.03622}.

\bibitem[Wu et~al., 2018]{wu2018unsupervised}
Wu, Z., Xiong, Y., Yu, S.~X., and Lin, D. (2018).
\newblock Unsupervised feature learning via non-parametric instance discrimination.
\newblock In {\em Proceedings of the IEEE conference on computer vision and pattern recognition}, pages 3733--3742.

\bibitem[Xie et~al., 2020a]{xie2020selfb}
Xie, Q., Luong, M.-T., Hovy, E., and Le, Q.~V. (2020a).
\newblock Self-training with noisy student improves imagenet classification.
\newblock In {\em Proceedings of the IEEE/CVF conference on computer vision and pattern recognition}, pages 10687--10698.

\bibitem[Xie et~al., 2020b]{xie2020self}
Xie, X., Chen, J., Li, Y., Shen, L., Ma, K., and Zheng, Y. (2020b).
\newblock Self-supervised cyclegan for object-preserving image-to-image domain adaptation.
\newblock In {\em Computer Vision--ECCV 2020: 16th European Conference, Glasgow, UK, August 23--28, 2020, Proceedings, Part XX 16}, pages 498--513. Springer.

\bibitem[Yang et~al., 2022]{yang2022survey}
Yang, X., Song, Z., King, I., and Xu, Z. (2022).
\newblock A survey on deep semi-supervised learning.
\newblock {\em IEEE Transactions on Knowledge and Data Engineering}.

\bibitem[Zadrozny, 2004]{zadrozny2004learning}
Zadrozny, B. (2004).
\newblock {Learning and Evaluating Classifiers Under Sample Selection Bias}.
\newblock In {\em International Conference on Machine Learning (ICML)}.

\bibitem[Zbontar et~al., 2021]{zbontar2021barlow}
Zbontar, J., Jing, L., Misra, I., LeCun, Y., and Deny, S. (2021).
\newblock Barlow twins: Self-supervised learning via redundancy reduction.
\newblock In {\em International Conference on Machine Learning}, pages 12310--12320. PMLR.

\bibitem[Zhang et~al., 2017]{zhang2016understanding}
Zhang, C., Bengio, S., Hardt, M., Recht, B., and Vinyals, O. (2017).
\newblock Understanding deep learning requires rethinking generalization.
\newblock In {\em International Conference on Learning Representations (ICLR)}.

\bibitem[Zhang et~al., 2021]{zhang2020coping}
Zhang, J., Menon, A., Veit, A., Bhojanapalli, S., Kumar, S., and Sra, S. (2021).
\newblock Coping with label shift via distributionally robust optimisation.
\newblock In {\em International Conference on Learning Representations (ICLR)}.

\bibitem[Zhang et~al., 2013]{zhang2013domain}
Zhang, K., Sch{\"o}lkopf, B., Muandet, K., and Wang, Z. (2013).
\newblock {Domain Adaptation Under Target and Conditional Shift}.
\newblock In {\em International Conference on Machine Learning (ICML)}.

\bibitem[Zhang, 2019]{zhang2019shiftinvar}
Zhang, R. (2019).
\newblock Making convolutional networks shift-invariant again.
\newblock In {\em ICML}.

\bibitem[Zhang et~al., 2018]{zhang2018collaborative}
Zhang, W., Ouyang, W., Li, W., and Xu, D. (2018).
\newblock Collaborative and adversarial network for unsupervised domain adaptation.
\newblock In {\em Proceedings of the IEEE conference on computer vision and pattern recognition}.

\bibitem[Zhang et~al., 2019]{zhang2019bridging}
Zhang, Y., Liu, T., Long, M., and Jordan, M. (2019).
\newblock Bridging theory and algorithm for domain adaptation.
\newblock In {\em International Conference on Machine Learning}. PMLR.

\bibitem[Zhu and Ghahramani, 2003]{zhu2002learning}
Zhu, X. and Ghahramani, Z. (2003).
\newblock Learning from labeled and unlabeled data with label propagation.
\newblock {\em CMU CALD tech report CMU-CALD-02-107, 2002}.

\end{thebibliography}

\clearpage
\appendix
\section*{Appendix}
\startcontents[appendices]
\printcontents[appendices]{l}{1}{\section*{Appendix Table of Contents}}

\section{Other Related Works}
\label{appsec:other-relwork}

\textbf{Unsupervised domain adaption.~~} 
Without assumption on the nature of shift, 
UDA is  underspecified~\citep{ben2010impossibility}.
This challenge has been addressed in various ways by researchers. 
One approach is to investigate additional
structural assumptions
under which UDA problems are well posed~\citep{shimodaira2000improving, scholkopf2012causal}.
Popular settings for which DA is well-posed include 
(i) \emph{covariate shift}~\citep{zhang2013domain,zadrozny2004learning,cortes2010learning,cortes2014domain,gretton2009covariate}
where $p(x)$ can change from source to target but 
$p(y|x)$ remains invariant; 
and (ii) \emph{label shift}~\citep{saerens2002adjusting, lipton2018detecting, azizzadenesheli2019regularized, alexandari2019adapting, garg2020labelshift, zhang2020coping, roberts2022LLS, garg2022RLSbench} 
where the label marginal $p(y)$ can change 
but $p(x|y)$ is shared across source and target. 
Principled methods with strong theoretical guarantees exists
for adaptation under these settings 
when target distribution's support 
is a subset of the source support.
Other works~\citep{elkan2008learning, pusurvey, garg2021PUlearning, garg2022OSLS} extend the label shift setting to scenarios where previously unseen classes may appear in the target and $p(x|y)$ remains invariant among seen classes.
A complementary line of research focuses on constructing benchmarks to develop  heuristics
for incorporating the unlabeled target data, 
relying on benchmark datasets 
ostensibly representative of ``real-world shifts''
to adjudicate progress~\citep{santurkar2020breeds,venkateswara2017deep,sagawa2021extending, peng2019moment, visda2017}. As a result, 
various benchmark-driven heuristics have been proposed~\citep{long2015learning,long2017deep, sun2016deep, sun2017correlation, zhang2019bridging, zhang2018collaborative, ganin2016domain, sohn2020fixmatch}.  
Our work engages with the latter,
focusing on two popular methods: 
self-training and contrastive pretraining.

\textbf{Domain generalization.~~} 
In domain generalization, the model is given access to data from
multiple different domains and the goal is to generalize to a previously 
unseen domain at test time~\citep{blanchard2011generalizing, muandet2013domain}. For a survey of different algorithms for 
domain generalization, we refer the reader to \citet{gulrajani2020search}. A crucial distinction here  
is that unlike the domain generalization setting, 
in DA problems, we have access
to unlabeled examples from the test domain. 

\textbf{Semi-supervised learning.~~} 
To learn from a small amount of labeled supervision, semi-supervised learning methods leverage unlabeled data alongside to improve learning models. One of the seminal works in SSL is the pseudolabeling method \cite{scudder1965probability}, where a classifier is trained on the labeled data and then used to classify the unlabeled data, which are then added to the training set. The work of \citet{zhu2002learning} built on this by introducing graph-based methods, and the transductive SVMs \citep{joachims1999transductive} presented an SVM-based approach. 
More recent works have focused on deep learning techniques, and similar to UDA, self-training and contrastive pretraining have emerged as two prominent choices. We delve into these methods in greater detail in the following paragraphs. 
For a discussion on other SSL methods, we refer interested readers to  \citep{chapelle2006semi,van2020survey,yang2022survey}. 
 
\textbf{Self-training.~~} 
Two popular forms of self-training
are pseudolabeling~\cite{lee2013pseudo} and conditional entropy minimization~\citep{grandvalet2006entropy}, which have been observed to be closely
connected~\citep{berthelot2019remixmatch, lee2013pseudo, sohn2020fixmatch, shu2018dirt}. Motivated by its strong performance in SSL and UDA settings~\citep{sohn2020fixmatch,xie2020selfb, garg2022RLSbench, shu2018dirt}, several theoretical works have made attempts to understand its behavior~\citep{kumar2020understanding, wei2020theoretical, chen2020self}. \citep{wei2020theoretical, cai2021theory} aims to understand the behavior of the global minimizer of self-training objective by studying input
consistency regularization, which enforces stability of the prediction for different augmentations of the unlabeled data.
Our analysis of self-training is motivated by the work of \citet{chen2020self} which explores the iterative behavior of self-training to unlearn spurious features. The setting of spurious features is of particular interest, since prior works  have specifically analyzed the failures of out-of-distribution generalization in the presence of spurious features~\cite{nagarajan2020understanding,sagawa2020investigation}.

\textbf{Contrastive learning.~~}
An alternate line of work that uses unlabeled data for learning representations in the pretraining stage is contrastive learning~\citep{grill2020bootstrap,oord2018representation,caron2020unsupervised,chen2020simple,wu2018unsupervised}. Given an augmentation distribution, the main goal of contrastive objectives is to map augmentations drawn from the same input (positive pairs) to similar features, and force apart features corresponding to augmentations of different inputs (negative pairs)~\cite{caron2020unsupervised,caron2021emerging,he2020momentum}. Prior works~\cite{cabannes2023ssl,johnson2022contrastive,haochen2022theoretical} have also shown a close relationship between contrastive~\citep{chen2020simple,haochen2021provable} and non-contrastive objectives~\citep{bardes2021vicreg,zbontar2021barlow}. Consequently, in our analysis pertaining to the toy setup we focus on the mathematically non-contrastive objective Barlow Twins~\citep{zbontar2021barlow}. Using this pretrained backbone (either as an initialization or as a fixed feature extractor) a downstream predictor is learned using labeled examples. Several works~\citep{haochen2021provable,saunshi2022understanding,haochen2022theoretical,arora2019theoretical,johnson2022contrastive} have analyzed the in-distribution generalization of the downstream predictor via label consistency arguments on the graph of positive pairs (augmentation graph). In contrast, we study the impact of contrastive learning under distribution shifts in the UDA setup. Other works~\citep{shen2022connect,haochen2022beyond} that examine contrastive learning for UDA also conform to the augmentation graph view point, making additional assumptions that guarantee linear transferability. In our simplified setup involving spurious correlations, these abstract assumptions break easily when the augmentations are of a generic nature, akin to practice. Finally, some empirical works~\cite{mishra2021surprisingly,ma2021conditional} have found self-supervised objectives like contrastive pretraining to reduce dependence on spurious correlations. Corroborating their findings, we extensively evaluate the complementary benefits of contrastive learning and self-training on real-world datasets. Finding differing results in SSL  and UDA settings, we further examine their behavior theoretically in our toy setup.

\section{More Details on Problem Setup} \label{app:problem_setup}
In this section, we elaborate on our setup and methods studied in our work. 

\paragraph{Unsupervised Domain Adaptation (UDA).} We assume that we are given labeled data from the \emph{source} distribution and unlabeled data from a shifted, \emph{target} distribution, with the goal of performing well on target data.  
We assume that the source and target distributions 
have the same label marginals $\ProbS(y) = \ProbT(y)$ 
(\ie, no label proportion shift) 
and the same Bayes optimal predictor, 
\ie, $\argmax_y \ps(y \mid x) = \argmax_y \pt(y \mid x)$. 
Here, even with infinite labeled source data, 
the challenge lies in generalizing out-of-distribution. 
In experiments, we assume access to finite data but in theory, we assume population access to labeled source and unlabeled target. 

\paragraph{Semi-Supervised Learning (SSL).} Here, there is no distribution shift, 
\ie,  $\ProbS=\ProbT=\ProbU$. 
We are given a small number of labeled examples 
and a comparatively large amount of unlabeled examples, 
both drawn from the same distribution. Without loss of generality, we 
denote this distribution with $\ProbT$. The goal in SSL is to generalize in-distribution. The challenge is primarily due to limited access to labeled data. 
Here, in experiments, we assume limited access to labeled data but a comparatively larger amount of unlabeled in-distribution data. In theory, we assume population access to unlabeled data but limited labeled examples.

\paragraph{Methods.} As discussed in the main paper, we compare four methods for learning with labeled and  unlabeled data. \tabref{tab:methods-setup} summarizes the main methods and key differences between those methods in UDA and SSL setup. For exact implementation in our experiments, we refer reader to \appref{app:method}.

\section{Additional Experiments and Details} \label{app:experimental details}

\subsection{Additional setup and notation}
\label{app:additional-notation}

Recall, our goal is to learn a predictor 
that maps inputs $x \in \calX \subseteq \R^d$ 
to outputs $y \in \sY$. 
We parameterize predictors $f = h \circ \Phi : \R^d \mapsto \sY$, where $\Phi:\R^d \mapsto \R^k$ is a feature map 
and $h \in \R^{k}$ is a classifier 
that maps the representation to the final scores or logits.
With $A: \calX \to \calA$, we denote the augmentation function that takes in an input $x$ and outputs an augmented view of the input $A(x)$. 
Unless specified otherwise, we perform full-finetuning in all of our experiments on real-world data. That is, we backpropagate gradients in both the linear head $h$ and the backbone $\phi$. 
For UDA, we denote source labeled points as $\{(x_i, y_i)\}_{i=1}^n$ and target unlabeled points as  $\{(x^\prime_i)\}_{i=1}^m$.  
For SSL, we use the same notation for labeled and unlabeled in-distribution data. 

\subsection{Dataset details} \label{app:dataset}

For both UDA and SSL, we conduct experiments 
across eight benchmark datasets. 
Each of these datasets consists of domains,
enabling us to construct source-target pairs for UDA.  
The adopted source and target domains
are standard to previous studies~\citep{shen2022connect,garg2022RLSbench,sagawa2021extending}. 
Because the SSL setting lacks distribution shift,
we do not need to worry about domain designations and default to using source alone. 
To simulate limited supervision in SSL, 
we sub-sample the original labeled training set to 10\%.
Below provide exact details about the datasets used in our benchmark study.

\begin{itemize}[leftmargin=*]
    \item \textbf{CIFAR10 {} {}} We use the original CIFAR10 dataset~\citep{krizhevsky2009learning} as the source dataset. For target domains, we consider CINIC10~\citep{darlow2018cinic} which is a subset of Imagenet restricted to CIFAR10 classes and downsampled to 32$\times$32.

    \item \textbf{FMoW {} {}} In order to consider distribution shifts faced in the wild, we consider FMoW-WILDs~\citep{wilds2021,christie2018functional} from \textsc{Wilds} benchmark, which contains satellite images taken  in different geographical regions and at different times. We use the original train as source and OOD val and OOD test splits as target domains as they are collected over different time-period. Overall, we obtain 3 different domains (1 source and 2 targets).

    \item \textbf{BREEDs {} {}} We also consider {BREEDs} benchmark~\citep{santurkar2020breeds}  in our setup to assess robustness to subpopulation shifts. {BREEDs} leverage class hierarchy in ImageNet~\citep{russakovsky2015imagenet} to re-purpose original classes to be the subpopulations and defines a classification task on superclasses. 
    We consider distribution shift due to subpopulation shift which is induced by directly making the subpopulations present in the training and test distributions disjoint. 
    {BREEDs}  benchmark contains 4 datasets \textbf{Entity-13}, \textbf{Entity-30},
    \textbf{Living-17}, and \textbf{Non-living-26}, each focusing on 
    different subtrees and levels in the hierarchy. 
    Overall, for each of the 4 BREEDs datasets (i.e., {Entity-13}, {Entity-30},
    {Living-17}, and {Non-living-26}), we obtain one different domain which we consider as target. We refer to  source and target as follows: BREEDs sub-population 1, BREEDs sub-population 2.

    \item \textbf{OfficeHome {} {}} We use four domains (art, clipart, product and real) from OfficeHome dataset~\citep{venkateswara2017deep}. We use the product domain as source and the other domains as target.
        
    \item \textbf{Visda {} {}} We use three domains (train, val and test) from the Visda dataset~\citep{peng2018syn2real, visda2017}. While `train' domain contains synthetic renditions of the objects, `val' and `test' domains contain real world images. To avoid confusing, the domain names with their roles as splits, we rename them as `synthetic', `Real-1' and `Real-2'. We use the synthetic (original train set) as the source domain and use the other domains as target. 
\end{itemize}

\begin{figure}[t!]
    \centering
    \includegraphics[width=0.6\textwidth]{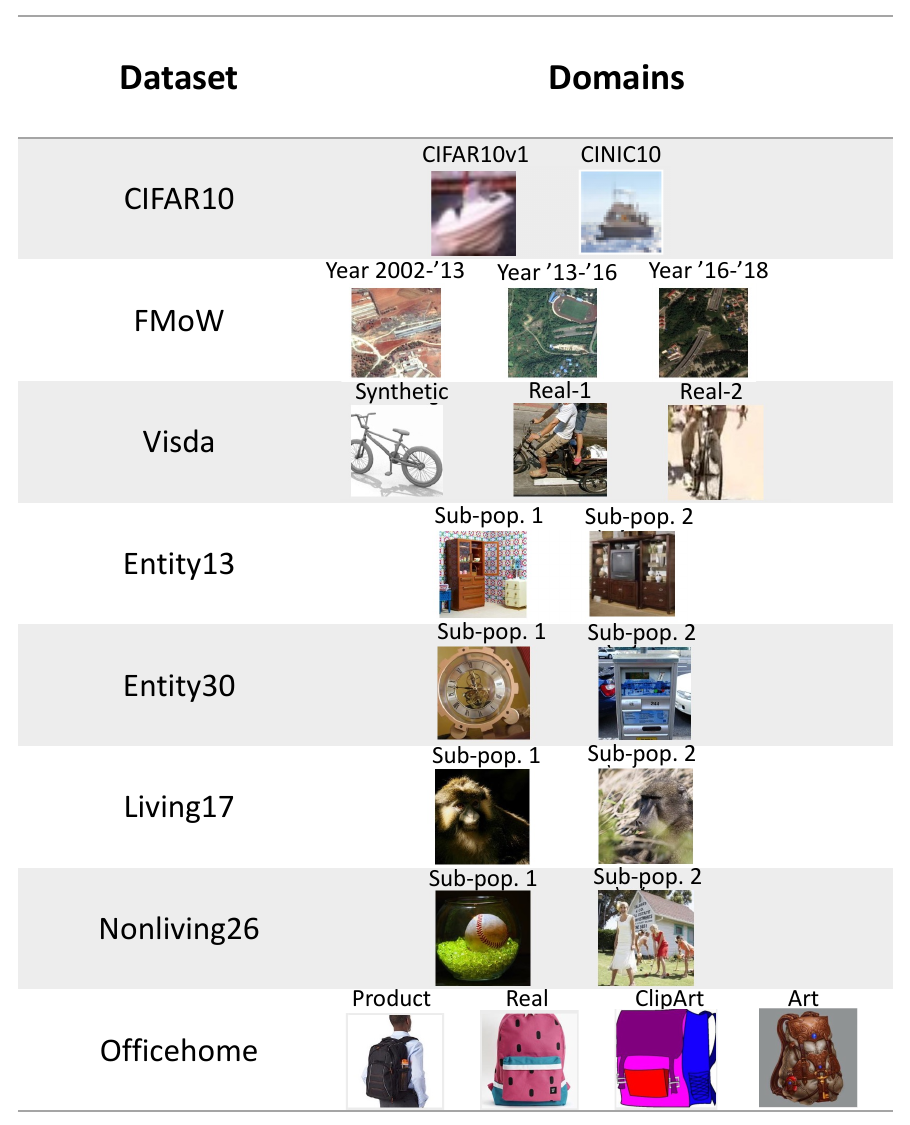}
    \caption{Examples from all the domains in each dataset. }
    \label{fig:testbed_imgs}
\end{figure}

We summarize the information about source and target domains in \tabref{table:dataset}.

\begin{table}[ht]
    \begin{adjustbox}{width=.9\columnwidth,center}
    \centering
    \small
    \tabcolsep=0.12cm
    \renewcommand{\arraystretch}{1.5}
    \begin{tabular}{lcccc}
    \toprule    
    Dataset && Source & & Target  \\
    \midrule
    CIFAR10 &&  CIFAR10v1  && CINIC10 \\ 
    FMoW && FMoW (2002--'13) && FMoW (2013--'16), FMoW (2016--'18)\\
    Entity13 && Entity13 (sub-population 1) && Entity13 (sub-population 2) \\
    Entity30 && Entity30 (sub-population 1) &&  Entity30 (sub-population 2), \\
    Living17 && Living17 (sub-population 1) &&  Living17 (sub-population 2), \\
    Nonliving26 && Nonliving26 (sub-population 1) &&  Nonliving26 (sub-population 2), \\
    Officehome && Product && Product, Art, ClipArt, Real\\
    Visda && \thead{Synthetic\\(originally referred \\to as train)} && \thead{Synthetic, Real-1 (originally referred to as val), \\ Real-2  (originally referred to as test)} \\
    \bottomrule 
    \end{tabular}  
    \end{adjustbox}  
    \vspace{5pt}
    \caption{Details of source and target sets in each dataset considered in our testbed.}\label{table:dataset}
 \end{table}

\textbf{Train-test splits~~} We partition each source and target dataset into $80\%$ and $20\%$ i.i.d. splits.  We use $80\%$ splits for training and $20\%$ splits for evaluation (or validation). We throw away labels for the $80\%$ target split and only use labels in the $20\%$ target split for final evaluation. 
The rationale behind splitting the target data is to use a completely unseen batch of data for evaluation. This avoids evaluating on examples where a model potentially could have overfit. 
over-fitting to unlabeled examples for evaluation. In practice, if the aim is to make predictions on all the target data (i.e., transduction), we can simply use the (full) target set for training and evaluation.

\textbf{Simulating SSL settings and limited supervision.}
For SSL settings, we choose the in-distribution domain as the source domain. 
To simulate limited supervision in SSL, 
we sub-sample the original labeled training set to 10\% and use all the original dataset as unlabeled data. For evaluation, we further split the original holdout set into two partitions (one for validation and the other to report final accuracy numbers).

\subsection{Method details} \label{app:method}

For implementation, we build on top of WILDs~\citep{sagawa2021extending} and RLSbench~\citep{garg2022RLSbench}  open source libraries.

\paragraph{ERM (Source only) training.}
We consider Empirical Risk Minimization (ERM) on the labeled source data 
as a baseline. Since this simply ignores the unlabeled target data, we call this as source only training. 
As mentioned in the main paper, we perform source only training with data augmentations. Formally, we minimize the following ERM loss: 
\begin{equation}
    L_{\text{source only}} (f) = \frac{1}{n}\sum_{i=1}^n \ell ( f(A(x_i), y_i) ) \,,
\end{equation}
where $A$ is the stochastic data augmentation operation and $\ell$ is a loss function. For SSL, the ERM baseline only uses the small of labeled data available. 

\paragraph{Contrastive Learning (CL).}
We perform contrastive pretraining on the unlabeled dataset to obtain the backbone $\phi_\cl$. And then we perform full fine-tuning with source labeled data by initializing the backbone with $\phi_\cl$. 
We use SwAV~\citep{caron2020unsupervised} for contrastive pretraining. The main idea behind SwAV is to train a model to identify different views of the same image as similar, while also ensuring that it finds different images to be distinct. This is accomplished through a \emph{swapped} prediction mechanism, 
where the goal is to compute a code from an augmented version of the image and predict this code from other augmented versions of the same image. 
In particular, given two image features $\phi(x^\prime_{a1})$ and $\phi(x^\prime_{a2})$ from two
different augmentations of the same image $x^\prime$, i.e., $x^\prime_{a1}, x^\prime_{a2} \sim A(x^\prime)$, SwAV computes their codes $z_{a1}$  and $z_{a2}$ by matching the
features to a set of $K$ prototypes $\{c_1, \cdots , c_K\}$. Then SwAV minimizes the following loss such that $\phi(x^\prime_{a1})$ can compute codes $z_{a2}$ and $\phi(x^\prime_{a2})$ can compute codes $z_{a1}$:  
\begin{align}
    L_\text{SwAV}(\phi) = \sum_{i=1}^{m} \sum_{x^\prime_{i, a1}, x^\prime_{i, a2} \sim A(x^\prime_{i})} \ell^\prime(\phi(x^\prime_{i, a1}), z_{i, a2}) + \ell^\prime(\phi(x^\prime_{i, a2}), z_{i, a1})  \,,
\end{align}
where $\ell^\prime$ computes KL-divergence between codes computed with features  (e.g. $\phi(x_{a1})$) and the code computed by another view (e.g. $z_{a2}$). 
For more details about the algorithm, we refer the reader to \citet{caron2020unsupervised}. 
In all UDA settings, unless otherwise specified,
we pool all the (unlabeled) data 
from the source and target to perform SwAV.
For SSL, we leverage in-distribution unlabeled data.

We employ SimCLR~\citep{chen2020simple} for the CIFAR10 dataset, aligning with previous studies that have utilized contrastive pretraining on the same dataset~\citep{kumar2022finetuning, shen2022connect}. The reason for this choice is that SwAV relies on augmentations that involve cropping images to a smaller resolution, making it more suitable for datasets with larger resolutions beyond $32\times32$.

\paragraph{Self-Training (ST).} For self-training, we apply FixMatch~\citep{sohn2020fixmatch},
where the loss on labeled data 
and on pseudolabeled unlabeled data
are minimized simultaneously.
\citet{sohn2020fixmatch} proposed FixMatch as a variant of the simpler Pseudo-label method~\citep{lee2013pseudo}. This algorithm dynamically generates psuedolabels and overfits on them in each batch. FixMatch employs consistency regularization on the  unlabeled data. 
In particular, while pseudolabels are generated on a
weakly augmented view of the unlabeled examples, the loss is computed with respect to predictions
on a strongly augmented view. The intuition behind such an update is to encourage a model to make predictions on weakly augmented data consistent with the strongly augmented example. Moreover, FixMatch only overfits to the assigned labeled with weak augmentation if the confidence of the prediction with strong augmentation is greater than some threshold $\tau$. 
Refer to $A_{\text{weak}}$ as the weak-augmentation and $A_{\text{strong}}$ as the strong-augmentation function. Then, FixMatch uses the following loss function: 
\begin{align*}
    L_{\text{FixMatch}} (f) &= \frac{1}{n}\sum_{i=1}^n \ell ( f(A_{\text{strong}}(x_i), y_i) ) \\ 
    &+ \frac{\lambda}{m} \sum_{i=1}^{m} \ell ( f(A_{\text{strong}}(x^\prime_i), \wt y_i) ) \cdot \indict{\max_y f_y(A_{\text{strong}} (x^\prime_i)) \ge \tau }\,,  
\end{align*}
where $\wt y_i = \argmax_y f_y(T_{\text{weak}} (x_i))$. For UDA, our unlabeled data is the union of source and target unlabeled data. For SSL, we only leverage in-distribution unlabeled data.

We adapted our implementation from \citet{sagawa2021extending} which matches the implementation of \citet{sohn2020fixmatch} except for one detail. 
While \citet{sohn2020fixmatch} augments labeled examples with weak augmentation, \citet{sagawa2021extending} proposed to strongly augment the labeled source examples. 

\textbf{Self-Training Over Contrastive learning (STOC).} Finally, rather than performing FixMatch from a randomly initialized backbone, 
we initialize FixMatch with a contrastive pretrained backbone.

\subsection{Additional UDA experimemts} \label{app:additional_UDA_results}

\begin{table}[H]
  \centering
  \footnotesize
  \setlength{\tabcolsep}{6pt}
  \renewcommand{\arraystretch}{1.2}
  \caption{\emph{Results in the UDA setup}. We report accuracy on target (OOD) data from which we only observe unlabeled examples during training. 
  For benchmarks with multiple target distributions (\eg, OH, Visda), we report average accuracy on those targets.}\label{table:UDA_results_STD}
  \vspace{5pt}
  \resizebox{\linewidth}{!}{%
  \begin{tabular}{@{}*{10}{c}@{}|c}
  \toprule
  {Method} & \parbox{1.cm}{Living17} & \parbox{1.cm}{Nonliv26} & \parbox{1.cm}{Entity13} & \parbox{1.cm}{Entity30} & \parbox{1.cm}{FMoW (2 tgts)} & \parbox{1.cm}{Visda (2 tgts)}  & \parbox{1.cm}{~~OH \\(3 tgts)}   & \parbox{1.cm}{CIFAR$\to$ CINIC}  \\
  \midrule
  ERM & \eentry{60.2}{0.1}  & \eentry{45.4}{0.2} &  \eentry{68.6}{0.1} & \eentry{55.7}{0.0}  & \eentry{56.5}{0.1}  & \eentry{20.8}{0.2}  & \eentry{9.5}{0.2}  & \eentry{74.3}{0.1} \\
  ST & \eentry{71.1}{0.2}  & \eentry{56.8}{0.1} &  \eentry{78.0}{0.3} & \eentry{66.7}{0.1}  & \eentry{56.9}{0.4}  & \eentry{39.1}{0.1}  & \eentry{11.1}{0.1}  & \eentry{78.3}{0.3} \\
  CL &  \eentry{74.1}{0.2}& \eentry{57.4}{0.3} &  \eentry{76.9}{0.2} &  \eentry{66.6}{0.3} &   \eentry{61.5}{0.5} & 
\eentry{63.2}{0.2} &  \eentry{22.8}{0.1} & \eentry{77.5}{0.1}\\ 
  STOC (ours) & \beentry{82.6}{0.1}& \beentry{62.1}{0.2} & \beentry{81.9}{0.2} & \beentry{72.0}{0.2}  & \beentry{65.3}{0.1} & \beentry{70.1}{0.2}  & \beentry{27.1}{0.3} & \beentry{79.9}{0.3}  \\
  \bottomrule 
  \end{tabular}}  
\end{table}

\begin{table}[H]
  \centering
  \footnotesize
  \setlength{\tabcolsep}{6pt}
  \renewcommand{\arraystretch}{1.2}
  \caption{\emph{Results in the UDA setup with source only contrastive pretraining}. We report accuracy on target (OOD) data from which we only observe unlabeled examples during training.  
  For benchmarks with multiple target distributions (\eg, OH, Visda), we report average accuracy on those targets.}\label{table:UDA_results_STD_source_only}
  \vspace{5pt}
  \resizebox{\linewidth}{!}{%
  \begin{tabular}{@{}*{10}{c}@{}|c}
  \toprule
  {Method} & \parbox{1.cm}{Living17} & \parbox{1.cm}{Nonliv26} & \parbox{1.cm}{Entity13} & \parbox{1.cm}{Entity30} & \parbox{1.cm}{FMoW (2 tgts)} & \parbox{1.cm}{Visda (2 tgts)}  & \parbox{1.cm}{~~OH \\(3 tgts)}   & \parbox{1.cm}{CIFAR$\to$ CINIC}  \\
  \midrule
  CL (source only) & \eentry{67.3}{0.1}  & \eentry{49.1}{0.2} &  \eentry{71.5}{0.1} & \eentry{58.5}{0.3}  & \eentry{53.9}{0.1}  & \eentry{33.3}{0.2}  & \eentry{21.7}{0.1}  & \eentry{77.7}{0.1} \\
  STOC (source only) & \eentry{75.0}{0.2}  & \eentry{58.4}{0.1} &  \eentry{79.8}{0.3} & \eentry{67.5}{0.1}  & \eentry{56.3}{0.4}  & \eentry{42.7}{0.1}  & \eentry{25.7}{0.1}  & \eentry{77.8}{0.1} \\
  CL &  \eentry{74.1}{0.2}& \eentry{57.4}{0.3} &  \eentry{76.9}{0.2} &  \eentry{66.6}{0.3} &   \eentry{61.5}{0.5} & 
\eentry{63.2}{0.2} &  \eentry{22.8}{0.1} & \eentry{77.5}{0.1}\\ 
  STOC & \beentry{82.6}{0.1}& \beentry{62.1}{0.2} & \beentry{81.9}{0.2} & \beentry{72.0}{0.2}  & \beentry{65.3}{0.1} & \beentry{70.1}{0.2}  & \beentry{27.1}{0.3} & \beentry{79.9}{0.3}  \\
  \bottomrule 
  \end{tabular}}  
\end{table}

\subsection{Additional SSL experimemts} \label{app:additional_SSL_results}

\begin{table}[H]
  \centering
  \footnotesize
  \setlength{\tabcolsep}{6pt}
  \renewcommand{\arraystretch}{1.2}
  \caption{\emph{Results in the SSL setup}. We report accuracy on hold-out ID data. Recall that SSL uses labeled and unlabeled data from the same distribution during training.}\label{table:SSL_results_STD}
  \vspace{5pt}
  \resizebox{\linewidth}{!}{%
  \begin{tabular}{@{}*{10}{c}@{}|c}
  \toprule
  {Method} & \parbox{1.cm}{Living17} & \parbox{1.cm}{Nonliv26} & \parbox{1.cm}{Entity13} & \parbox{1.cm}{Entity30} & \parbox{1.cm}{FMoW} & \parbox{1.cm}{Visda}  & \parbox{1.cm}{~~OH}   & \parbox{1.cm}{CIFAR}  \\
  \midrule
  ERM  & \eentry{76.8}{0.1} & \eentry{64.9}{0.2} & \eentry{80.1}{0.0} & \eentry{70.4}{0.3} & \eentry{33.6}{0.4} & \eentry{99.2}{0.0} & \eentry{32.0}{0.2} & \eentry{85.5}{0.1}  \\
  ST  & \eentry{85.4}{0.1} & \eentry{75.7}{0.2} & \eentry{85.4}{0.2} & \eentry{77.3}{0.1} & \eentry{33.6}{0.3} & \eentry{99.2}{0.1} & \eentry{32.0}{0.1} & \eentry{93.1}{0.1} \\
  CL  & \eentry{91.1}{0.5} & \eentry{84.6}{0.6} & \eentry{90.7}{0.4} & \eentry{85.5}{0.3} & \eentry{43.1}{0.2} & \eentry{97.6}{0.3} & \eentry{49.7}{0.2} & \eentry{91.7}{0.2}\\
  STOC (ours)  & \eentry{92.0}{0.1} & \eentry{85.8}{0.2} & \eentry{91.3}{0.3} & \eentry{86.1}{0.2} & \eentry{44.4}{0.1} & \eentry{97.7}{0.2} & \eentry{49.9}{0.2} & \eentry{93.06}{0.3}  \\
  \bottomrule 
  \end{tabular}}  
\end{table}

\subsection{Other experimental details} \label{app:other_exp}

\textbf{Augmentations.~~}
For weak augmentation, we leverage random horizontal flips and random crops of pre-defined size. For SwAV, we also perform multicrop augmentation as proposed in \citet{caron2020unsupervised}. 
For strong augmentation, we apply the following transformations sequentially: random horizontal flips, random crops of pre-defined size, augmentation with Cutout~\citep{devries2017improved}, and RandAugment~\citep{cubuk2020randaugment}.  For the exact
implementation of RandAugment, we directly use the implementation of \citet{sohn2020fixmatch}.  Unless specified otherwise, for all methods, 
we default to using strong augmentation techniques.

\textbf{Architectures.~~} In our work, we experiment with Resnet18, Resnet50~\citep{he2016deep} 
trained from scratch (\ie random initialization).
We do not consider off-the-shelf pretrained models
(\eg, on Imagenet~\citep{russakovsky2015imagenet}) 
to avoid confounding our conclusions about contrastive pretraining.
However, we note that our results on most datasets 
tend to be comparable to and sometimes exceed
those obtained with ImageNet pretrained models.
For BREEDs datasets, we employ Resnet18 architecture. For other datasets, we train a Resnet50 architecture. 

Except for Resnets on CIFAR dataset, we used the standard pytorch implementation~\citep{gardner2018gpytorch}. For Resnet on Cifar, we refer to the implementation here: \url{https://github.com/kuangliu/pytorch-cifar}. For all the architectures, whenever applicable, we add antialiasing~\citep{zhang2019shiftinvar}. 
We use the official library released with the paper.

\textbf{Hyperparameters.~~}
For all the methods, we fix the algorithm-specific hyperparameters
to the original recommendations. 
For UDA, given that the setup precludes access 
to labeled data from the target distribution,
we use source hold-out performance 
to pick the best hyperparameters.
During pretraining, early stopping is done according to
lower values of pretraining loss.

We tune the learning rate and $\ell_2$ regularization parameter by fixing the batch size for each dataset that corresponds to the maximum we can fit to 15GB GPU memory. 
We default to using cosine learning rate schedule~\citep{loshchilov2016sgdr}.  
We set the number of epochs for training as per the 
suggestions of the authors of respective benchmarks. 
For SSL, we run both ERM and FixMatch for approximately $2000$ epochs.  
Note that we define the number of epochs as a 
full pass over the labeled training source data.  
We summarize the learning rate, batch size, number of epochs, and 
$\ell_2$ regularization parameter used in our study in \tabref{table:hyperparameter_dataset}.

\begin{table}[ht!]
    \begin{adjustbox}{width=0.8\columnwidth,center}
    \centering
    \small
    \tabcolsep=0.12cm
    \renewcommand{\arraystretch}{1.5}
    \begin{tabular}{lccccccccc}
    \toprule    
    Dataset && Batch size && $\ell_2$ regularization set && Learning rate set \\
    \midrule
    CIFAR10 && 200 && $\{0.001, 0.0001, 10^{-5}, 0.0\}$ &&  $\{0.2, 0.1, 0.05, 0.01, 0.003, 0.001\}$ \\ 
    FMoW && 64 && $\{0.001,  0.0001, 10^{-5}, 0.0\}$ && $\{0.01, 0.003, 0.001,   0.0003, 0.0001\}$  \\
    Entity13 && 256 && $\{0.001, 0.0001,  10^{-5}, 0.0\}$ && $\{0.4, 0.2, 0.1, 0.05, 0.02, 0.01, 0.005\}$  \\
    Entity30 && 256 && $\{0.001, 0.0001,  10^{-5}, 0.0\}$ &&  $\{0.4, 0.2, 0.1, 0.05, 0.02, 0.01, 0.005\}$  \\
    Entity30 && 256 && $\{0.001, 0.0001,  10^{-5}, 0.0\}$ && $\{0.4, 0.2, 0.1, 0.05, 0.02, 0.01, 0.005\}$  \\
    Nonliving26 && 256 && $\{ 0.001, 0.0001,10^{-5}, 0.0\}$ && $\{0.4, 0.2, 0.1, 0.05, 0.02, 0.01, 0.005\}$  \\
    Officehome && 96 && $\{ 0.001, 0.0001,10^{-5}, 0.0\}$ && $\{ 0.01, 0.003, 0.001,  0.0003,0.0001\}$ \\
    Visda && 96  && $\{0.001, 0.0001, 10^{-5}, 0.0\}$ && $\{0.03, 0.01, 0.003, 0.001, 0.0003\}$ \\
    \bottomrule 
    \end{tabular}  
    \end{adjustbox}  
    \vspace{5pt}
    \caption{Details of the batch size, learning rate set and $\ell_2$ regularization set considered in our testbed.} \label{table:hyperparameter_dataset}
 \end{table}

\textbf{Compute infrastructure.~~}
Our experiments were performed across a combination of Nvidia T4, A6000, and V100 GPUs. 

\section{Additional Results in Toy Setup}
\label{appsec:additional-toy-results}
In this section we will first give more details on our simplified setup that captures both contrastive pretraining and self-training in the same framework. Then, we provide some additional empirical results that are not captured theoretically but mimic behaviors observed in real world settings, highlighting the richness of our setup.

\subsection{Detailed description of our simplified setup} \label{app:toy_description}

In this subsection, we will first re-iterate the problem setup in \secref{sec:theory} and provide some comparisons between our setup and those in closely related works. We will then describe the four methods: ERM, ST, CL, and STOC, providing details on the exact estimates returned by these algorithms in the SSL and UDA settings.

\textbf{Data distribution.~~} We consider binary classification and model the inputs as consisting of two kinds of features: $x = [\xin, \xsp]$
where 
$\xin \in \R^{\din}$ is the invariant feature that is predictive of the label across both source $\ProbS$ and target $\ProbT$ and $\xsp \in \R^{\dsp}$ is the spurious feature that is correlated with the label $y$ only on the source domain $\ProbS$ but uncorrelated with label $y$ in $\ProbT$.
Here, $\xin \in \Real^{\din}$ determines the label using the ground truth classifier $\wstar \sim \unif(\Sphere^{\din-1})$, and $\xsp \in \Real^{\dsp}$ is strongly correlated with the label on source but random noise on target. 
Formally, we sample $\ry \sim \unif \{-1, 1\}$ and generate inputs $x$ conditioned on $\ry$ as follows
\begin{align}
\ProbS:~~ &\xin \sim  \sN(\gamma \cdot \ry \wstar, \Sigmain)~~~\xsp = \ry \mathbf{1}_{\dsp} \nonumber \\
\ProbT:~~ &\xin \sim \sN(\gamma \cdot \ry \wstar, \Sigmain) ~~~\xsp \sim \sN(\mathbf{0}, \Sigmasp), \label{eq:toy-model-main-paper}
\end{align}
where $\gamma$ is the margin afforded by the invariant feature. We set covariance of the invariant features $\Sigmain = \sigmain^2\cdot(\mathbf{I}_{\din} - {\wstar\wstar}^\top)$ to capture structure in the invariant feature that the variance is less along the latent predictive direction $\wstar$. Note that the spurious feature is completely predictive of the label in the source data, and is distributed as spherical Gaussian in the target data with $\Sigmasp = \sigmasp^2 \mathbf{I}_{\dsp}$.

\textbf{Why is our simplified setup interesting?~~} 
In our setup, $\xin$ is the hard to learn feature that generalizes from source to target. The hardness of learning this feature is determined by the value of the margin $\gamma$ and how it compares with size of the spurious feature ($\sqrt{\dsp}$). Since, $\gamma/\sqrt{\dsp}$ is small in our setup, $\xin$ is much harder to learn on source data (even with population access) compared to the spurious feature $\xsp$ which generalizes poorly from source to target. 
These two types of features have been captured in similar analysis on spurious correlations~\cite{sagawa2020investigation,nagarajan2020understanding} since it imitates pitfalls emanating from the presence of spurious features in real world datasets (\eg, the easy to learn background feature in image classification problems). While this setup is simple, it is also expressive enough to elucidate both self-training and contrastive learning behaviors we observe in real world settings. Specifically, it captures the separation results we observe in \secref{sec:exp}.  

\textbf{Differences of our setup with prior works.~~}
While our distribution shift settings bears the above similarities it also has important differences with works analyzing self-training and contrastive pretraining individually.
\citet{chen2020self} analyze the iterative nature of self-training algorithm, where the premise is that we are given a classifier that not only has good performance on source data but in addition does not rely too much on the spurious feature. Under the strong condition of small norms along the spurious feature, they show that self-training can provably unlearn this small dependence when the target data along the spurious feature is random noise. This assumption is clearly violated in setups where the spurious correlation is strong (as in our toy setup), \ie, the dependence on the spurious feature is rather large (much larger than that on the invariant feature) for any classifier that is trained directly on source data. Consequently, we show the need for ``good'' pretrained representations from contrastive pretraining over which if we train a linear predictor (using source labeled data), it will provably have a reduced ``effective'' dependence on the spurious feature. 

Using an augmentation distribution similar to ours, \citet{saunshi2022understanding} carried out contrastive pretraining analysis with the backbone belonging to a capacity constrained function class (similar analysis also in ~\cite{haochen2022beyond}). Our setup differs from this in two key ways: (i) we specifically consider a distribution shift from source to target. Unlike their setting, it is not sufficient to make augmentations consistent with ground truth labels, since the predictor that uses just the spurious feature also assigns labels consistent with both ground truth predictions and augmentations on the source data; and (ii) our augmentation distribution assumes no knowledge of the invariant feature, which is why we augment all dimensions uniformly, as opposed to selectively augmenting a set of dimensions. In other words, we assume no knowledge of the structure of the optimal target predictor. For \eg, if we had knowledge of the spurious dimensions we could have just selectively augmented those. Assuming knowledge of these perfect augmentations is not ideal for two reasons: (a) it makes the problem so easy that just training an ERM model on source data with these augmentations would already yield a good target predictor (which rarely happens in practice); and (b) in real-world datasets perfect augmentations for the downstream task are not known. Hence, we stick to generic augmentations in our setup.

\subsection{Discussion on self-training and contrastive learning objectives} \label{app:objectives}

\begin{table}[!ht]
    \footnotesize
    \renewcommand{\arraystretch}{1.5} 
    \centering
    \begin{tabular}{p{0.6cm}cc}
        {Method} & \textbf{UDA Setup} & \textbf{SSL Setup} \\ \midrule
        \multirow{2}{*}{\textbf{ERM}:} & \multirow{2}{*}{$h_{\mathrm{erm}} = \argmin_h \E_\ProbS \ell(h(x), y)$} & $h_{\mathrm{erm}} = \argmin_h \frac{1}{n}\sum_{i=1}^n \ell(h(x_i), y_i)$   \\
        & & $\{(x_i, y_i)\}_{i=1}^n \sim \ProbT^n$ \\ \midrule
        \multirow{2}{*}{\textbf{ST}:} &  Starting from $h_{\mathrm{erm}}$ optimize over $h$ (to get $h_\mathrm{st}$):  & Starting from $h_{\mathrm{erm}}$ optimize over $h$ (to get $h_\mathrm{st}$): \\
            &  $ \E_{\ProbT(x)} \ell(h(x), \sgn(h(x)))$ & $ \E_{\ProbT(x)} \ell(h(x), \sgn(h(x)))$   \\    \midrule
        \multirow{3}{*}{\textbf{CL}:} &  $\Phi_\cl = \argmin_\phi \calL_\cl(\Phi)$ & $\Phi_\cl = \argmin_\phi \calL_\cl(\Phi)$  \\
                            &   Use $(\ProbS(x) + \ProbT(x))/2$ for $\calL_\cl(\Phi)$ & Use $\ProbT(x)$ for $\calL_\cl(\Phi)$ \\
                            &  $h_\cl = \argmin_h \E_\ProbS \ell(h\circ\Phi_\cl(x), y)$ &  $h_\cl = \argmin_h \frac{1}{n}\sum_{i=1}^n \ell(h \circ \Phi_\cl(x_i), y_i)$\\ \midrule
     \multirow{3}{*}{\textbf{STOC}:} & Starting from $h_{\mathrm{cl}}$ optimize over $h$ (to get $h_\mathrm{stoc}$): & Starting from $h_{\mathrm{cl}}$ optimize over $h$ (to get $h_\mathrm{stoc}$):\\  
                & $ \E_{\ProbT(x)} \ell(h\circ \Phi_\cl(x), \sgn(h\circ \Phi_\cl(x)))$ &  $  \E_{\ProbT(x)} \ell(h\circ \Phi_\cl(x), \sgn(h\circ \Phi_\cl(x)))$  \\
    \end{tabular}
    \vspace{2em} 
    \caption{\textbf{Description of methods for SSL vs. UDA}: For each method we provide exact objectives used for experiments and analysis in the SSL and UDA setups (pertaining to \secref{sec:theory}).}
    \label{tab:methods-setup}
\end{table}

In text we will describe our objectives and methods for the UDA setup. In Table~\ref{tab:methods-setup} we constrast the differences in the methods and objectives for SSL and UDA setups. Recall from Section~\ref{sec:problem-setup} that we learn linear classifiers $h$ over features extractors $\Phi$. We consider linear feature extractor i.e. $\Phi$ is a matrix in $\R^{k \times d}$. For mathematical convenience, we assume access to infinite unlabeled data and hence replace the empirical quantities over unlabeled data with their population counterpart. In the UDA setting, we further assume access to infinite labeled data from the source. Note that due to distribution shift between source and target, ``ERM'' on infinite labeled data from the source does not necessarily achieve optimal performance on the target. For binary classification, we assume that the linear layer $h$ maps features to a scalar in $\R$ such that the prediction is $\sgn(h^\top \Phi x)$. We use the exponential loss $\ell (f(x), y) = \exp\left(-y f(x)\right)$ as the classification loss. 

\textit{Contrastive pretraining.} We obtain $\Phi_\cl \coloneqq \argmin_\Phi \calL_\cl(\Phi)$ by minimizing the Barlow Twins objective~\citep{zbontar2021barlow}, which prior works have shown is also equivalent to spectral contrastive and non-contrastive objectives~\citep{garrido2022duality, cabannes2023ssl}. 
In \secref{sec:theory}, we consider a constrained form of Barlow Twins in \eqref{eq:cont-loss} which enforces representations of different augmentations $a_1, a_2$ of the same input $x$ to be close in representation space, while ensuring feature diversity by staying in the constraint set. We assume a strict constraint on regularization $(\rho=0)$ for the theoretical arguments in the rest of the main paper. In \appref{appsubsec:bt-optimization} we prove that all our claims hold for small $\rho$ as well. In \eqref{eq:cont-loss-redefined}, we redefine the pretraining objective with a regularization term (instead of a constraint set) where $\kappa$ controls the strength of the regularization term, with higher values of $\kappa$ corresponding to stronger constraints on feature diversity.  We then learn a linear classifier $h_\cl$ over $\Phi_\cl$ to minimize the exponential loss on labeled source data. 

\begin{align}
    \calL_\cl(\Phi) \; \coloneqq \; \Exp_{x \sim \ProbU}\Exp_{a_1, a_2 \sim \ProbA(\cdot \mid x)} \; \|\Phi(a_1) - \Phi(a_2)\|_2^2 \;\; + \;\; \kappa \cdot \norm{\Exp_{a \sim \ProbA}\brck{\Phi(a)\Phi(a)^\top} - \mathbf{I}_k}{F}^2
     \label{eq:cont-loss-redefined}
\end{align}

\textit{Augmentations.} Data augmentations play a key role in contrastive pre-training (and also as we see later, state-of-the-art self-training variants like FixMatch). Given input $x \in \calX$, let $\ProbA(a \mid x)$ denote the distribution over its augmentations, and $\ProbA$ denote the marginal distribution over all possible augmentations.
 We use the following simple augmentations where we scale the magnitude of each co-ordinate by a uniformly independent amount, \ie,
 \begin{align}
     a \sim \ProbA(\cdot \mid x)  \equiv c \odot x \;\;\; \textrm{where,}\;\;\; c \sim \unif[0, 1]^d.
 \end{align} The performance of different methods heavily depends on the assumptions we make on augmentations. We try to mirror practical settings where the augmentations are fairly ``generic'', not encoding any information about which features are invariant or spurious, and hence perturb all features symmetrically. 

\textit{Self-training.} ST performs ERM in the first stage using labeled data from the source,
and then subsequently updates the head $h$ by iteratively generating pseudolabels on the unlabeled target:
\begin{align}
    \calL_\st(h; \Phi) \; \coloneqq \Exp_{\ProbT(x)} \ell(h^\top \Phi x, \sgn(h^\top \Phi(x)))\qquad
   \textrm{Update:}
   \;\; h^{t+1} = \frac{h^{t} - \eta \nabla_h \calL_\st (h^t; \Phi)}{\norm{h^{t} - \eta \nabla_h \calL_\st (h^t; \Phi)}{2}}
    \label{eq:self-training-redefined}
\end{align}
For convenience, we keep the feature backbone $\Phi$ fixed across the self-training iterations and only update the linear head on the pseudolabels.

\textit{\method (Self-training after contrastive learning).} Finally, we can combine the two unsupervised objectives where we do the self-training updates(~\ref{eq:self-training}) with $h_0 = h_\cl$ and $\Phi_0 = \Phi_\cl$ starting with the contrastive learning model rather than just source-only ERM. Here, we only update $h$ and fix $\Phi_\cl$.

\subsection{Additional empirical results in our simplified setup} \label{app:additional_toy_exp}

\begin{figure}
  \centering
  \begin{subfigure}{0.44\textwidth}
    \centering
    \includegraphics[width=\linewidth]{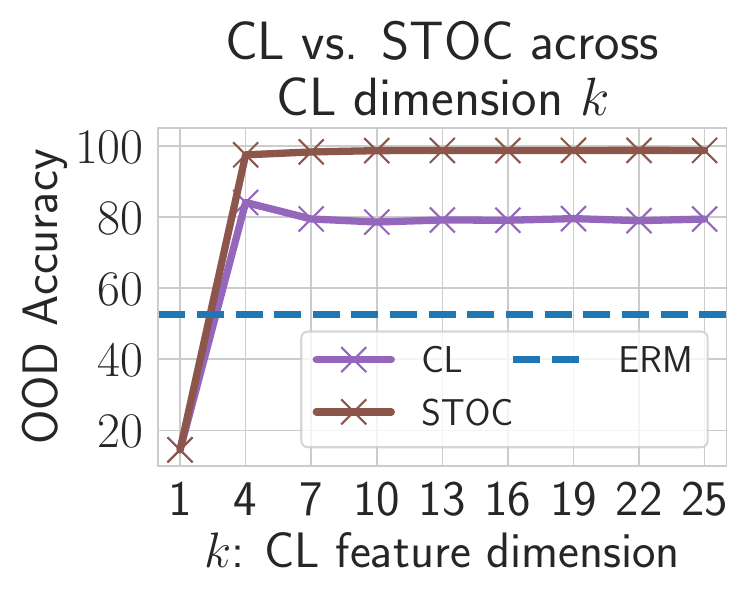}
    \caption*{}
  \end{subfigure}
  \hfill
  \begin{subfigure}{0.44\textwidth}
    \centering
    \includegraphics[width=\linewidth]{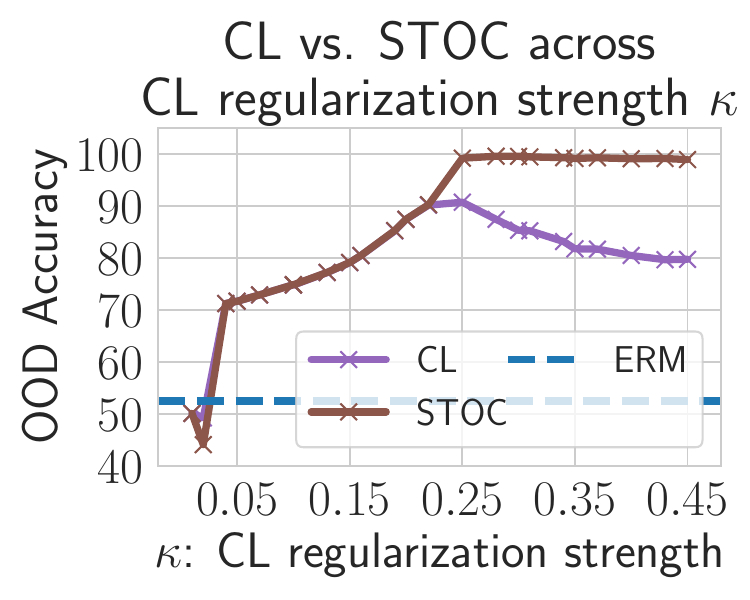}
    \caption*{}
  \end{subfigure}
  \caption{\textbf{Ablations on pretraining hyperparameters:} In the UDA setup we plot the performance of CL and STOC as we vary two pretraining hyper-parameters: \emph{(left)} the output dimension $(k)$ of the feature extractor $\Phi$; and \emph{(right)} the strength $(\kappa)$ of the regularizer in the Barlow Twins objective in \eqref{eq:cont-loss-redefined}. While ablating on $k$ we fix $\kappa=0.5$, and while ablating on $\kappa$ we fix $k=10$. Other problem parameters are taken from Example 1.}
  \label{fig:toy-ablations}
\end{figure}

We conduct two ablations on the hyperparameters for contrastive pretraining. First, we vary the dimensionality $k$ of the linear feature extractor $\Phi \in \Real^{k \times d}$. Second, we vary the regularization strength $\kappa$ that enforces feature diversity in the Barlow Twins objective \eqref{eq:cont-loss-redefined}. In Figure~\ref{fig:toy-ablations} we plot these ablations in the UDA setup.

\textbf{Varying feature dimension.~~} We find that CL recovers the full set of predictive features (\ie both spurious and invariant) only when $k$ is large enough (Figure~\ref{fig:toy-ablations}\emph{(left)}). Since the dimensionality of the true feature is $5$ in our Example 1, reducing $k$ below the true feature dimension hurts CL. Once $k$ crosses a certain threshold, CL features completely capture the projection of the invariant feature $\winv$. After this point, it amplifies the component along $\winv$. It retains the amplification over the spurious feature $\wspu$ even as we increase $k$. This is confirmed by our finding that further increasing $k$ does not hurt CL performance. This is also inline with our theoretical observations, where we find that for suitable $\wstar$, the subspace spanned by $\winv$ and $\wspu$ are contained in a low rank space (as low as rank $2$) of the contrastive representations (Theorem~\ref{thm:bt-blockform}). Once CL has amplified the dependence along $\winv$ STOC improves over CL by unlearning any remaining dependence on the spurious $\wspu$. The above arguments for the CL trend also explain why the performance of STOC continues to remain $\approx 100\%$ as we vary $k$.

\textbf{Varying regularization strength.~~} In our main theoretical arguments we consider the constrained form of the Barlow Twins objective \eqref{eq:cont-loss} with a strict constraint of $\rho=0$ (we relax this theoretically as well, see \ref{appsubsec:bt-optimization}). For our experiments, we optimize the regularized version of this objective \eqref{eq:cont-loss-redefined}, where the constraint term now appears as a regularizer which enforces feature diversity, \ie the features learned through contrastive pretraining span orthogonal parts of the input space (as governed under the metric defined by augmentation covariance matrix $\Sigma_A$). If $\kappa$ is very low, then trivial solutions exist for the Barlow Twins objective. For \eg, $\phi \approx \mathbf{0}$ (zero vector) achieves very low invariance loss. When $\kappa < 0.05$, we find that CL recovers these trivial solutions (Figure~\ref{fig:toy-ablations}\emph{(right)}). Hence, both CL and STOC perform poorly. As we increase $\kappa$ the performance of both CL and STOC improve, mainly because the features returned by $\Phi_\cl$ now comprise of the predictive directions $\winv$ and $\wspu$, as predictive by our theoretical  arguments for $\rho=0$ (which corresponds to large $\kappa$). On the other hand, when $\kappa$ is too high optimization becomes hard since $\kappa$ directly effects the Lipschitz constant of the loss function. Hence, the performance of CL drops by some value. Note that this does not effect the performance of STOC since CL continues to amplify $\winv$ over $\wsp$ even if it is returning suboptimal solutions with respect to the optimization loss of the pretraining objective.

\subsection{Reconciling Practice: Experiments with deep networks in toy setup} \label{app:deep_toy}
\begin{figure}[!t]
  \centering
    \includegraphics[width=0.7\linewidth]{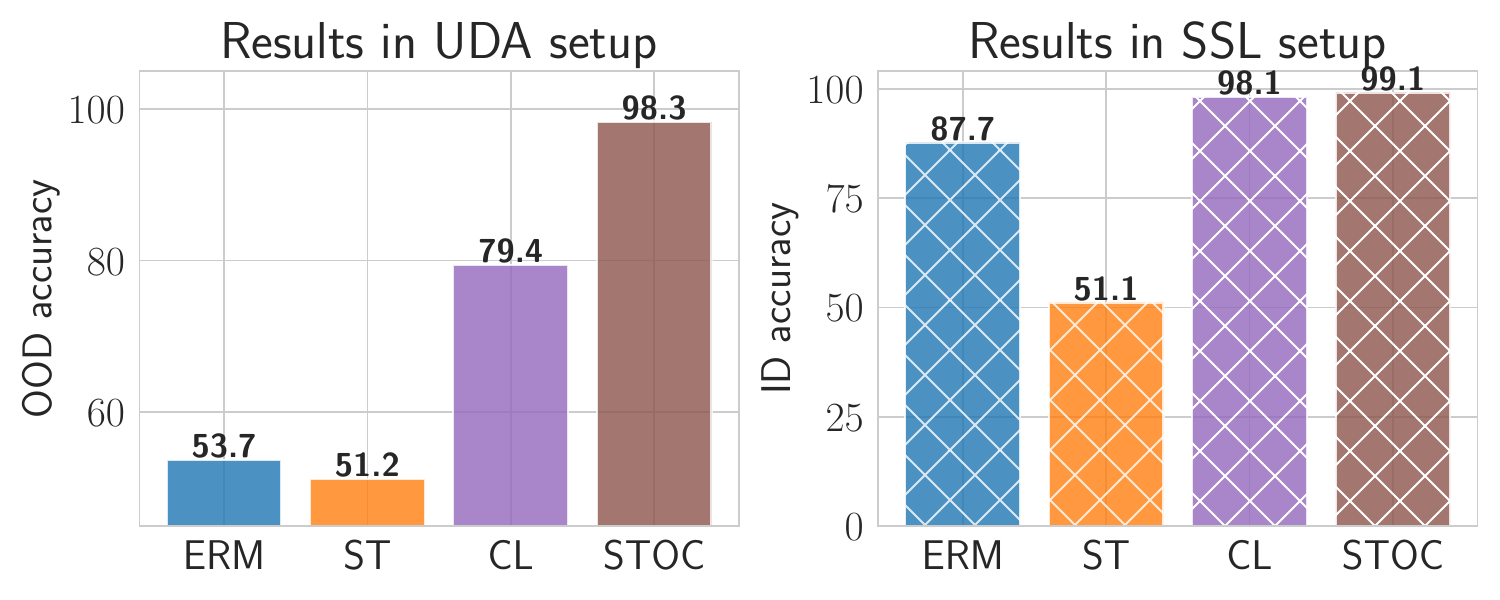}
    \caption{\textbf{Results with linear backbone:} We plot the OOD accuracy for ERM, CL, ST and STOC in the UDA setup and ID accuracy in the SSL setup when the feature extractor $\Phi$ is a linear network. Note, that the feature extractor is still fixed during CL and STOC.}
    \label{fig:app-toy-additional-1}
\end{figure}

\begin{figure}[!b]
  \centering
    \includegraphics[width=0.7\linewidth]{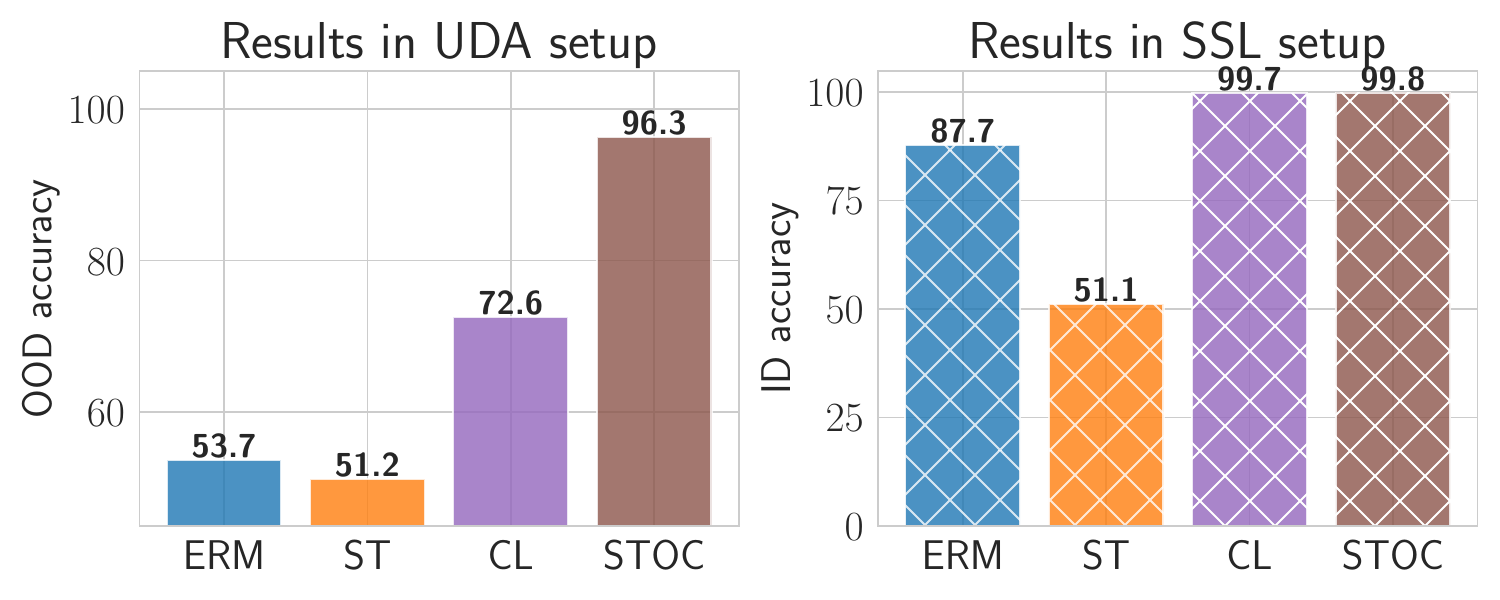}
    \caption{\textbf{Results with non-linear backbone:} We plot the OOD accuracy for ERM, CL, ST and STOC in the UDA setup and ID accuracy in the SSL setup when the feature extractor $\Phi$ is a non-linear one-hidden layer network with ReLU activations. Note, that the feature extractor is still fixed during CL and STOC.}
    \label{fig:app-toy-additional-2}
\end{figure}

In this section we delve into the details of \secref{subsec:deep-networks}, \ie, we analyze performance of different methods when we make some design choices that imitate practice. First, we look at experiments involving a deep non-linear backbone $\Phi$. Here, the non-linear $\Phi$ is learned during contrastive pretraining and fixed for CL and STOC. Then, we investigate trends when we continue to propagate gradients onto $\Phi$ during STOC (we call this full-finetuning). Unlike previous cases, this allows features to be updated.

\textbf{Results with non-linear feature extractor $\Phi$.~~} In ~\figref{fig:app-toy-additional-2} we plot the performance of the four methods when we use a non-linear feature extractor during contrastive pretraining. This feature extractor is a one-hidden layer neural network (hidden dimension is 500) with ReLU activations. We find that the trends observed with linear backbones in ~\figref{fig:app-toy-additional-1} are also replicated with the non-linear one. Specifically, we note that STOC improves over CL under distribution shifts, whereas CL is already close to optimal when there are no distribution shifts. We also see that CL and ST individually are subpar. In SSL, we see a huge drop in the performance of ST (over ERM) mainly because we only fit on pseudolabels during ST. This is different from practice where we continue to optimize loss on labeled data points while fitting the pseudolabels. Consequently, when we continue to optimize performance on source labeled data the performance of ST in SSL setup is improves from $51.1\% \to 72.6\%$.

\begin{figure}[!t]
  \centering
  \begin{subfigure}{0.44\textwidth}
    \centering
    \includegraphics[width=\linewidth]{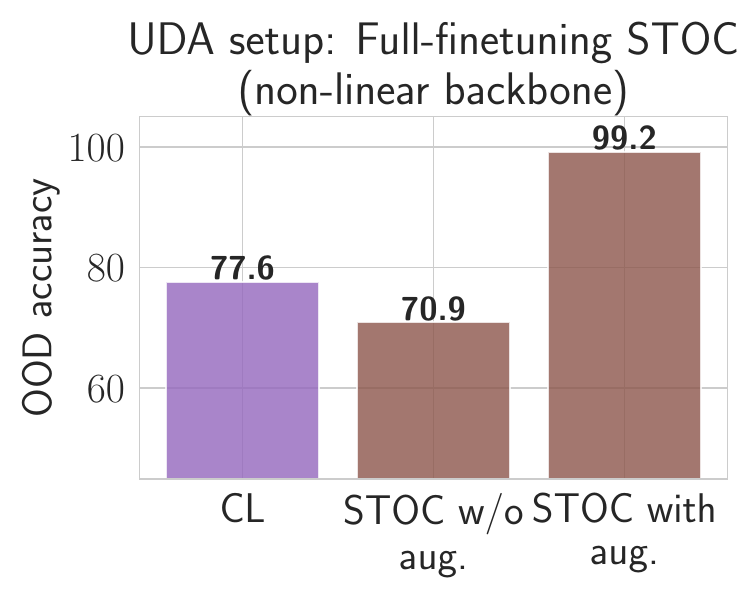}
    \caption*{}
  \end{subfigure}
  \hfill
  \begin{subfigure}{0.44\textwidth}
    \centering
    \includegraphics[width=\linewidth]{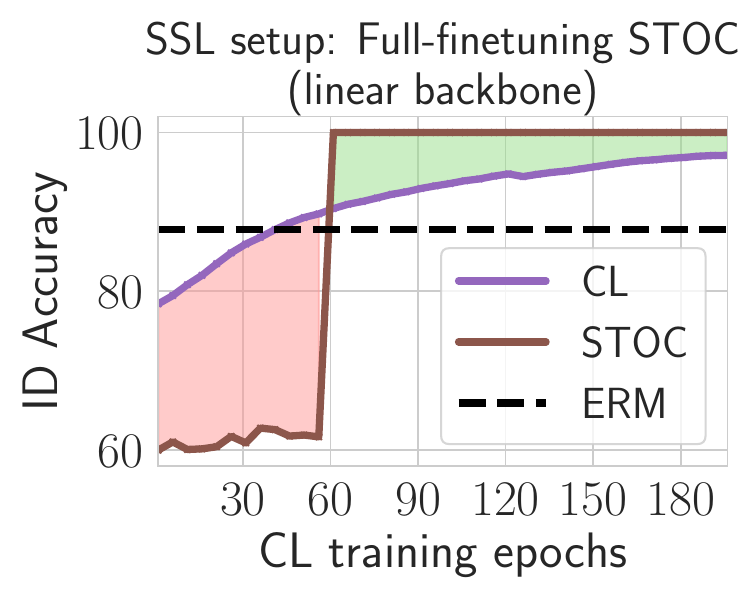}
    \caption*{}
  \end{subfigure}
  \caption{\textbf{Finetuning the contrastive representations during STOC:} We propagate gradients to the feature backbone $\Phi$ when running STOC algorithm. Note that CL still fixes the contrastive representations when learning a fixed linear head over it. On the \emph{(left)} we show results in UDA setup where we compare the performance of STOC with and without augmentations (along with other practical design choices like confidence thresholds and continuing to optimize source loss as done in FixMatch) when the feature backbone is non-linear. On the \emph{(right)} we show results for STOC and CL in the SSL setup when the feature backbone is linear.}
  \label{fig:toy-deep-nets}
\end{figure}

\textbf{Results with full fine-tuning.~~} Up till this point, we have only considered the case (for both SSL and UDA) where we fix the contrastive learned features when running CL and STOC, \ie, we only optimized the linear head $h$. Now, we shall consider the setting where gradients are propagated to $\Phi$ during STOC. Note that we still fix the representations for training the linear head during CL. Results for this setting are in Figure~\ref{fig:toy-deep-nets}. We show two interesting trends that imitate real world behaviors.

\emph{STOC benefits from augmentations during full-finetuning:} In the UDA setup we find that ST while updating $\phicl$ can hurt due to overfitting issues when training with the finite sample of labeled and unlabeled data (drop by $>7\%$ over CL).
This is due to overfitting on 
confident but incorrect pseudolabels on target data. 
This can exacerbate components along spurious feature $\wspu$ from source. 
One reasoning behind this is that deep neural networks can perfectly memorize them on finite unlabeled target data~\citep{zhang2016understanding}. 
Heuristics typically used in practice (\eg in FixMatch~\cite{sohn2020fixmatch}) help
avoid overfitting on incorrect pseudolabels: (i) confidence thresholding;
to pick confident pseudolabel examples; 
(ii) pseudolabel a different augmented input 
than the one
on which the self-training loss is optimized; and 
(iii) optimize source loss with labeled data simultaneously when fitting pseudolabels. 
Intuitively, thresholding introduces a curriculum where we only learn confident examples in the beginning whose pseudolabels are mainly determined by component along the invariant feature $\winv$. 
Augmentations prevent the neural network from memorizing incorrect pseudolabels and optimizing source loss prevents forgetting of features learned during CL. When we implement these during full-finetuning in STOC we see that STOC now improves over CL (by $>20\%$).

\emph{Can we improve contrastive pretraining features during STOC?~~} 
We find that self-training can also improve features learned during contrastive pretraining when we update the full backbone during STOC (see Figure~\ref{fig:toy-deep-nets}\emph{(right)}). Specifically, in the SSL setup we find that STOC can now improve substantially over CL. Recall, that when we fixed $\Phi_\cl$ this was not possible (see \ref{subsec:ssl-analysis} and \figref{fig:analysis-empirical}(b)). This is mainly because STOC can now improve performance beyond just recovering the generalization gap for the linear head (which is typically small). This feature improvement is observed even when we fully finetune a linear feature extractor. Similar trends are also observed with the non-linear backbone. But, it becomes harder to identify a good stopping criterion for CL training. 
Thus, it remains unclear if STOC and CL have complementary benefits for feature learning in UDA or SSL settings. Investigating this is an interesting avenue for future work.

\section{Formal Statements from \secref{sec:theory}}
\label{appsec:proofs}

Recall from Section~\ref{sec:problem-setup}
that we learn linear classifiers $h$ over features extractors $\Phi$.
We consider linear feature extractor i.e. $\Phi$ is a matrix in $\R^{d \times k}$ and
the linear layer $h: \R^k \to \R$ with a prediction as $\sgn(h^\top \Phi x)$. We use the exponential loss $\ell (f(x), y) = \exp\left(-y f(x)\right)$.

\subsection{Analysis of ERM and ST: Formal Statement of \thmref{thm:scratch_training}} \label{app:ST_analysis}
For ERM and ST, we train both $h$ and $\Phi$. This is equivalent to $\Phi = I_{d\times d}$ being identity and training a linear head $h$. Recall that the ERM classifier is obtained by minimizing the population loss on labeled 
source data: 
\begin{align}
    \herm = \argmin_{h} \Expt{(x,y)\sim\ProbS}{\ell(x, y)} \,. \label{eq:ERM_loss}
\end{align}

We split \thmref{thm:scratch_training} into \thmref{thm:ERM_scratch} and \thmref{thm:ST_scratch_formal}. 
Before we characterize the ERM solution, we recall some additional notation. 
Define $\winv$$=$$[\wstar, 0, ..., 0]^\top$, and $\wspu = [0, ..., 0, \nicefrac{\mathbf{1}_{d_\mathrm{sp}}}{\sqrt{d_\mathrm{sp}}}]^\top$. 
The following proposition characterizes $\herm$ and 0-1 error of the classifier 
on target: 
\begin{theorem}[ERM classifier and its error on target] \label{thm:ERM_scratch}
ERM classifier obtained as in \eqref{eq:ERM_loss} is given by 
$$\frac{\herm}{\norm{\herm}{2}} = \frac{\gamma \cdot \winv + \sqrt{\dsp} \cdot \wsp}{\sqrt{\gamma^2 + \dsp}}\,.$$
The target accuracy of $\herm$ is given by $0.5\cdot \erfc\left( -\nicefrac{\gamma^2}{\paren{\sqrt{2 \dsp}\cdot \sigmasp}} \right)$. 
\end{theorem}
\begin{proof}
    To prove this theorem, we first derive a closed-form expression for the ERM classifier and then use \lemref{lemma:error_target} to derive its  0-1 error on target. For Gaussian data with the same covariance matrices for class conditional $\ProbS(x|y=1)$ and $\ProbS(x|y=0)$, Bayes decision rule is given by the Fisher's linear discriminant direction (Chapter 4; \citet{bishop2006pattern}): 
    \begin{align*}
        h(x) = \begin{cases}
                1, & \text{if } h^\top x > 0 \\
                0, & \text{otherwise}
                \end{cases}
    \end{align*}
    where $h = 2\cdot \gamma(\winv) + 2\cdot \sqrt{\dsp}(\wsp)$. Plugging $h$ in \lemref{lemma:error_target} we get the desired result. 
\end{proof}
ST performs ERM
in the first stage using labeled data from the source,
and then subsequently updates the head $h$ 
by iteratively generating pseudolabels on the unlabeled target:
\begin{align}
\footnotesize
    \calL_\st(h) \; \coloneqq \Exp_{\ProbT(x)} \ell(h^\top x, \sgn(h^\top x))\,.
    \label{eq:ST_loss}
\end{align} 
Starting with $\hst^0 = \nicefrac{\herm}{\norm{\herm}{2}}$ (the classifier obtained with ERM) we perform the following iterative procedure for self-training: 
\begin{align}
 \hst^{t+1} = \frac{\hst^{t} - \eta \nabla_h \calL_\st (\hst^t)}{\norm{\hst^{t} - \eta \nabla_h \calL_\st (\hst^t)}{2}}
\end{align}
Next, we characterize ST solution: 

\begin{theorem}[ST classifier and its error on target] \label{thm:ST_scratch_formal}
Starting with ERM solution, ST will lead to: 
\begin{enumerate}
\renewcommand{\labelenumi}{(\roman{enumi})}
    \item (Necessary condition) $\hst^t = \wsp$ as $t \to \infty$, such that the target accuracy is 50\% for all
    $\sigmasp \ge 1$
    and $\gamma \le \frac{1}{2\sqrt{\sigmasp}}$. 
    \item (Sufficient condition) $\hst^t = \winv$  as $t \to \infty$, such that the target accuracy is 100\% when the problem parameters $\gamma, \sigmasp$ satisfy: $\gamma \ge \sigmasp$. 
\end{enumerate}
\end{theorem}
\begin{proof}
    The proof can be divided into two parts: (i) deriving closed-form expressions for updates on $\hst^t$ in terms of $\hst^{t-1}$ and (ii) obtaining conditions under which the component along $\winv$ monotonically increases or decreases with $t$ after re-normalizing the norm of updated $h$. For notation convenience, we denote $\hst$ with $h$ in the rest of the proof. 

    \paragraph{Part-1.} First, the loss of self-training with classifier $h \defeq [\hinv, \hsp]$ where $\hinv \in \Real^{\din}$  and $\hsp \in \Real^{\dsp}$ is given by: 
    \begin{align*}
        \calL_\st(h) &= \Expt{\ProbT(x)}{\ell(h^\top x, \sgn(h^\top x))} \numberthis \\ 
        &= \Expt{\ProbT(x)}{ \exp\left(-\sign(h^\top x) \cdot (h^\top x) \right) } \numberthis  \\
         &= \Expt{\ProbT(x)}{ \exp\left(- \abs{h^\top x} \right) } \numberthis \\
         &= \Expt{\ProbT(x)}{ \exp\left(- \abs{\hinv^\top \xin + \hsp^\top \xsp} \right) } \numberthis  \label{eq:int1_loss_st} \\
         &= \Exp_{ y \sim^U\{-1, 1\}  , z \sim \cN(0,1) } \left[\exp\left(- \left\vert \gamma \cdot y\cdot \hinv^\top \wstar \right.\right.\right. \\ & \qquad \qquad \qquad + \left. \left. \left. \left[ \sigmain(\norm{\hinv}{2}^2 - (\hinv^T \wstar)^2 ) + \sigmasp \cdot \norm{\hsp}{2} \right] \cdot z \right\vert \right) \right]\,. \numberthis \label{eq:int2_loss_st} \\
         &= \Expt{ z \sim \cN(0,1) }{\exp\left(- \abs{\gamma \cdot \hinv^\top \wstar + \left[ \sigmain(\norm{\hinv}{2}^2 - (\hinv^T \wstar)^2 ) + \sigmasp \cdot \norm{\hsp}{2} \right] \cdot z} \right)}\,, \label{eq:int3_loss_st} \numberthis 
    \end{align*}
where \eqref{eq:int1_loss_st} to \eqref{eq:int2_loss_st} is implied by simply replacing the definition of target distribution and \eqref{eq:int2_loss_st} to \eqref{eq:int3_loss_st} is implied by the symmetry of the function with respect to $y$ and $-y$ due to the symmetry of the absolute function and Gaussian distribution. 
For a classifier $h^t$, we denote $\mu_t = \gamma \cdot {\hinv^t}^\top \wstar$ and $\sigma_t = \left[ \sigmain(\norm{\hinv^t}{2}^2 - ({\hinv^t}^T \wstar)^2 ) + \sigmasp \cdot \norm{\hsp^t}{2} \right]$. With this notation, we can re-write the loss in \eqref{eq:int3_loss_st} as $\calL_\st(h^t) = \Expt{ z \sim \cN(0,\sigma_t^2) }{ \exp\left( -\abs{\mu_t + z}\right)}$. 

Now we derive a closed-form expression of $\calL_\st(h^t)$ in \lemref{lemma:g_def}: 
\begin{align}
    \calL_\st(h^t) &= \frac{1}{2} \left(  \expb{ \frac{\sigma_t^2}{2} - \mu_t} \cdot \erfcb{ -\frac{\mu_t}{\sqrt{2} \sigma_t} + \frac{\sigma_t}{\sqrt{2}} } + \expb{ \frac{\sigma_t^2}{2} + \mu_t} \cdot \erfcb{ \frac{\mu_t}{\sqrt{2} \sigma_t} + \frac{\sigma_t}{\sqrt{2}} } \right) \,.
\end{align}

Define the Mill's ratio as $\rb{x} = \expb{x^2/2} \cdot \erfcb{x / \sqrt{2}} \cdot \sqrt{\pi/2}$ as in \citet{baricz2008mills}. We will frequently use standard properties of the Mill's ratio. We list them in \lemref{lemma:mill_ratio} for completeness. 
Define: 
\begin{align*}
    \alpha_1(\mu_t, \sigma_t) &= - \expb{ \frac{\sigma_t^2}{2} - \mu_t} \cdot \erfcb{ -\frac{\mu_t}{\sqrt{2} \sigma_t} + \frac{\sigma_t}{\sqrt{2}} } + \expb{ \frac{\sigma_t^2}{2} + \mu_t} \cdot \erfcb{ \frac{\mu_t}{\sqrt{2} \sigma_t} + \frac{\sigma_t}{\sqrt{2}} } \,, \\
    &= \sqrt{\frac{2}{\pi}} \expb{-\frac{\mu_t^2}{2\sigma_t^2}} \left[\rb{\sigma_t + \frac{\mu_t}{\sigma_t}}  - \rb{ \sigma_t - \frac{\mu_t}{\sigma_t}} \right] \numberthis \label{eq:alpha_1} \\ 
    \alpha_2(\mu_t, \sigma_t) &= \expb{ \frac{\sigma_t^2}{2} - \mu_t} \cdot \erfcb{ -\frac{\mu_t}{\sqrt{2} \sigma_t} + \frac{\sigma_t}{\sqrt{2}} } + \expb{ \frac{\sigma_t^2}{2} + \mu_t} \cdot \erfcb{ \frac{\mu_t}{\sqrt{2} \sigma_t} + \frac{\sigma_t}{\sqrt{2}} } \\ 
    & \qquad \qquad \qquad -  \frac{2\sqrt{2}}{\sigma_t\sqrt{\pi}} \expb{-\frac{\mu_t^2}{2\sigma_t^2}} \\ 
    &= \sqrt{\frac{2}{\pi}} \expb{-\frac{\mu_t^2}{2\sigma_t^2}} \left[\rb{\sigma_t + \frac{\mu_t}{\sigma_t}}  + \rb{ \sigma_t - \frac{\mu_t}{\sigma_t}} - \frac{2}{\sigma_t} \right]   \,. \numberthis \label{eq:alpha_2} \\
\end{align*}

Let $\wt h^{t+1}$ denote the un-normalized gradient descent update at iterate $t+1$. We have: 
\begin{align}
    \wt h^{t+1} = h^t - \eta \cdot \frac{\partial \calL_\st(h^t)}{\partial h} \,.  
\end{align}

Now we will individually argue about the update of $\wt h^{t+1}$ along the first $\din$ dimensions and 
the last $\dsp$ dimensions. First, we have: 
\begin{align*}
    \wt h^{t+1}_{\mathrm{in}} &= \hinv^t - \eta \cdot \frac{\partial \calL_\st(h^t)}{\partial \hinv} \\
    &= \hinv^t  - \frac{\eta}{2} \left(  - \expb{ \frac{\sigma_t^2}{2} - \mu_t} \cdot \erfcb{ -\frac{\mu_t}{\sqrt{2} \sigma_t} + \frac{\sigma_t}{\sqrt{2}} } \right. \\ 
    & \qquad  \qquad \left. + \expb{ \frac{\sigma_t^2}{2} + \mu_t} \cdot \erfcb{ \frac{\mu_t}{\sqrt{2} \sigma_t} + \frac{\sigma_t}{\sqrt{2}} } \right) \cdot \gamma \cdot \wstar \\
    & \qquad  \qquad -\frac{\eta}{2} \left(  \expb{ \frac{\sigma_t^2}{2} - \mu_t} \cdot \erfcb{ -\frac{\mu_t}{\sqrt{2} \sigma_t} + \frac{\sigma_t}{\sqrt{2}} } \right. \\
    & \qquad \qquad  \left. + \expb{ \frac{\sigma_t^2}{2} + \mu_t} \cdot \erfcb{ \frac{\mu_t}{\sqrt{2} \sigma_t} + \frac{\sigma_t}{\sqrt{2}} } \right. \\ & \qquad \qquad \left. - \frac{2\sqrt{2}}{\sigma_t\sqrt{\pi}} \expb{-\frac{\mu_t^2}{2\sigma_t^2}} \right) \cdot  ( 2\hinv^t - 2 ({\hinv^t}^\top \wstar) \wstar) \cdot \sigmain^2 \\
    &= \hinv^t  - \frac{\eta}{2} \cdot \alpha_1(\mu_t, \sigma_t) \cdot \gamma \cdot \wstar -\frac{\eta}{2} \cdot \alpha_2(\mu_t, \sigma_t) \cdot  ( 2\hinv^t - 2 ({\hinv^t}^\top \wstar) \wstar) \cdot \sigmain^2 \,. \numberthis \label{eq:update_ST_inv_scratch} \\
\end{align*}
Notice that the update of $h^{t+1}_{\mathrm{in}}$ is split into two components, one along $\wstar$ and the other along the orthogonal component  $2\hinv^t - 2 ({\hinv^t}^\top \wstar) \wstar$. 
We will now argue that since at initialization, the component along 
$(I - \wstar {\wstar}^\top)$ 
is zero then it will remain zero. In particular, we have: 
\begin{align}
    {\hinv^0}^\top (I - \wstar {\wstar}^\top) \, \propto \, {\wstar}^\top (I - \wstar {\wstar}^\top) = 0 \,.
\end{align} 
With \eqref{eq:update_ST_inv_scratch}, we can argue that if $(I - \wstar {\wstar}^\top){\hinv^t} = 0$, 
then $(I - \wstar {\wstar}^\top) {\wt h^{t+1}_{\mathrm{inv}}}= 0$ implying that $(I - \wstar {\wstar}^\top) {\wt h^{t}_{\mathrm{in}}} = 0$ for all $t>0$. Hence, we have: 
\begin{align*}
    \wt h^{t+1}_{\mathrm{inv}} &= \hinv^t - \eta \cdot \frac{\partial \calL_\st(h^t)}{\partial \hinv} \\
    &= \hinv^t  - \frac{\eta}{2} \cdot \alpha_1(\mu_t, \sigma_t) \cdot \gamma \cdot \wstar  \,. \numberthis \label{eq:update_ST_inv_scratch_2} \\
\end{align*}
Second, we have the update $\wt h^{t+1}_{\mathrm{sp}}$ given by: 
\begin{align*}
    \wt h^{t+1}_{\mathrm{sp}} &= \hsp^t - \eta \cdot \frac{\partial \calL_\st(h^t)}{\partial \hsp} \\
    &= \hsp^t  - \frac{\eta}{2} \left(  \expb{ \frac{\sigma_t^2}{2} - \mu_t} \cdot \erfcb{ -\frac{\mu_t}{\sqrt{2} \sigma_t} + \frac{\sigma_t}{\sqrt{2}} } \right. \\ 
    & \qquad  \qquad \left. + \expb{ \frac{\sigma_t^2}{2} + \mu_t} \cdot \erfcb{ \frac{\mu_t}{\sqrt{2} \sigma_t} + \frac{\sigma_t}{\sqrt{2}} } - \frac{2\sqrt{2}}{\sigma_t\sqrt{\pi}} \expb{-\frac{\mu_t^2}{2\sigma_t^2}} \right) \cdot  \hsp^t \cdot \sigmasp^2 \\
    &= \hsp^t  -  \frac{\eta}{2} \cdot \alpha_2(\mu_t, \sigma_t) \cdot \hsp^t \cdot \sigmasp^2  \,. \numberthis \label{eq:update_ST_sp_scratch} \\
\end{align*}
Re-writing the expressions \eqref{eq:update_ST_inv_scratch_2} and \eqref{eq:update_ST_sp_scratch} for the update of $\wt h^{t+1}$, we have: 
\begin{align}
    \wt h^{t+1}_{\mathrm{in}} &= \hinv^t (1  - \frac{\eta}{2} \cdot \alpha_1(\mu_t, \sigma_t) \cdot \gamma^2 / \mu_t)\,. \label{eq:final_h_inv_exp} \\
    \wt h^{t+1}_{\mathrm{sp}} &= \hsp^t  ( 1 - \frac{\eta}{2} \cdot \alpha_2(\mu_t, \sigma_t) \cdot \sigmasp^2) \label{eq:final_h_sp_exp} \,.
\end{align}
Here, we replace $\hsp^t = \mu_t \cdot \wstar/\gamma$  in \eqref{eq:update_ST_inv_scratch_2} to get  \eqref{eq:final_h_inv_exp}. Updates in \eqref{eq:final_h_inv_exp} and \eqref{eq:final_h_sp_exp} show that $\wt h^{t+1}_{\mathrm{inv}}$ remains in the direction of $\hinv^t$ and $\wt h^{t+1}_{\mathrm{sp}}$ remains in the direction of $\hsp^t$.

\paragraph{Part-2.} 
Now we will derive conditions under which $\hinv^t$ and $\hsp^t$ will show monotonic behavior for necessary and sufficient conditions. We will first argue the condition under which ST will provably fail and converge to a classifier with a random target performance. For this, at every $t$, if we have: 
\begin{align}
    \frac{\norm{\wt h^{t+1}_{\mathrm{sp}}}{2}}{\norm{\wt h^{t+1}}{2}} > \norm{ \hsp^t }{2} \label{eq:necessary_initial}\,,
\end{align}
then we can argue that as $t \to \infty$, we have $\norm{ \hsp^t }{2} = 1$ and hence, the ST classifier will have random target performance. Thus, we will focus on conditions, under which the norm on $\norm{ \hsp^t }{2}$ increases with $t$.  Re-writing \eqref{eq:necessary_initial}, we have: 
\begin{align}
    {\norm{\wt h^{t+1}_{\mathrm{sp}}}{2}} &> \norm{\wt h^{t+1}}{2} \cdot {\norm{ \hsp^t }{2}} \\
    {\norm{\wt h^{t+1}_{\mathrm{sp}}}{2}} &> \left( \norm{\wt h^{t+1}_{\mathrm{sp}}}{2} + \norm{\wt h^{t+1}_{\mathrm{in}}}{2}\right) \cdot {\norm{ \hsp^t }{2}} \\
    {\norm{\wt h^{t+1}_{\mathrm{sp}}}{2}} \cdot \left( 1 -  {\norm{ \hsp^t }{2}}  \right)  &>  \norm{\wt h^{t+1}_{\mathrm{in}}}{2} \cdot {\norm{ \hsp^t }{2}} \\
    \frac{\norm{\wt h^{t+1}_{\mathrm{sp}}}{2}}{\norm{ \hsp^t }{2}} 
    &>  \frac{\norm{\wt h^{t+1}_{\mathrm{in}}}{2}}{{\norm{ \hinv^t }{2}}} 
 \label{eq:necessary}\,.
\end{align}

Plugging in \eqref{eq:final_h_inv_exp} and \eqref{eq:final_h_sp_exp} into \eqref{eq:necessary}, we get: 
\begin{align}
    \abs{1 - \frac{\eta}{2} \cdot \alpha_2(\mu_t, \sigma_t) \cdot \sigmasp^2} > \abs{1  - \frac{\eta}{2} \cdot \alpha_1(\mu_t, \sigma_t) \cdot \gamma^2 / \mu_t} \,.
\end{align}

For small enough $\eta$, we have the necessary condition for the failure of ST as: 
\begin{align}
     \alpha_2(\mu_t, \sigma_t) \cdot \sigmasp^2 < \alpha_1(\mu_t, \sigma_t) \cdot \gamma^2 / \mu_t  \label{eq:necessary_final}\,. 
\end{align}

Now we show in \lemref{lemma:cond_ST_sufficiency} and \lemref{lemma:cond_ST_necessary} that if the conditions assumed in the theorem continue to hold, then we can success and failure respectively.

\end{proof}

\begin{lemma}[Necessary conditions for ST]
Define $\alpha_1$ and $\alpha_2$ as in \eqref{eq:alpha_1} and \eqref{eq:alpha_2} respectively. 
If $\sigmasp \ge 1$ and $\gamma \le \frac{1}{2\sqrt{\sigmasp}}$, then we have for all $t$: 
\begin{align}
  \alpha_2(\mu_t, \sigma_t) \cdot \frac{\sigmasp^2 \cdot \mu_t}{\gamma^2} \le \alpha_1(\mu_t, \sigma_t) \,. \label{eq:ST_necessary_cond}
\end{align}
\label{lemma:cond_ST_necessary} 
\end{lemma}
\begin{proof}
We upper bound and lower bound $\alpha_1$ and $\alpha_2$ by using the properties of $\rb{\cdot}$. 
Recall: 
\begin{align}
    \alpha_1(\mu_t, \sigma_t) = \sqrt{\frac{2}{\pi}} \expb{-\frac{\mu_t^2}{2\sigma_t^2}} \left[\rb{\sigma_t + \frac{\mu_t}{\sigma_t}}  - \rb{ \sigma_t - \frac{\mu_t}{\sigma_t}} \right] \,.
\end{align}
and 
\begin{align}
    \alpha_2(\mu_t, \sigma_t) = \sqrt{\frac{2}{\pi}} \expb{-\frac{\mu_t^2}{2\sigma_t^2}} \left[\rb{\sigma_t + \frac{\mu_t}{\sigma_t}}  + \rb{ \sigma_t - \frac{\mu_t}{\sigma_t}} - \frac{2}{\sigma_t} \right] \,.
\end{align}

We now use Taylor's expansion on $\rb{\cdot}$ and we get: 
\begin{align}
  \rb{\sigma_t} +  \rbb{\prime}{\sigma_t} \cdot \left(\frac{\mu_t}{\sigma_t} \right) \le \rb{\sigma_t + \frac{\mu_t}{\sigma_t}} \le \rb{\sigma_t} +  \rbb{\prime}{\sigma_t} \cdot \left(\frac{\mu_t}{\sigma_t} \right) + R^{\prime\prime} \left(\frac{\mu_t}{\sigma_t} \right)^2
\end{align}
and similarly, we get: 
\begin{align}
  \rb{\sigma_t} -  \rbb{\prime}{\sigma_t} \cdot \left(\frac{\mu_t}{\sigma_t} \right) \le \rb{\sigma_t - \frac{\mu_t}{\sigma_t}} \le \rb{\sigma_t} - \rbb{\prime}{\sigma_t} \cdot \left(\frac{\mu_t}{\sigma_t} \right) + R^{\prime\prime} \left(\frac{\mu_t}{\sigma_t} \right)^2 \label{eq:taylor_expr}
\end{align}

where $R^{\prime\prime} = \rbb{{\prime\prime}}{\sigma_0}$. This is because  $\rbb{{\prime\prime}}{\cdot}$ takes positive values and is a decreasing function in $\sigma_t$ (refer to \lemref{lemma:mill_ratio}).  
We now lower bound $\alpha_1(\mu_t, \sigma_t)$ and upper bound $\alpha_2(\mu_t, \sigma_t)$: 
\begin{align}
    \frac{\alpha_1(\mu_t, \sigma_t)}{ \sqrt{\frac{2}{\pi}} \expb{-\frac{\mu_t^2}{2\sigma_t^2}} } \ge  2 \rbb{\prime}{\sigma_t} \cdot \left(\frac{\mu_t}{\sigma_t} \right) - R^{\prime\prime} \left(\frac{\mu_t}{\sigma_t} \right)^2
\end{align}

\begin{align}
    \frac{\alpha_2(\mu_t, \sigma_t)}{ \sqrt{\frac{2}{\pi}} \expb{-\frac{\mu_t^2}{2\sigma_t^2}} } \le  2 \rb{\sigma_t} +  2\cdot R^{\prime\prime} \left(\frac{\mu_t}{\sigma_t} \right)^2
\end{align}

Substituting the lower bound and upper bound in \eqref{eq:ST_necessary_cond} gives us the following as stricter a necessary condition (i.e., \eqref{eq:stricter_necessary} implies \eqref{eq:ST_necessary_cond}): 
\begin{align}
    & \left[ 2 \rb{\sigma_t} +  2\cdot R^{\prime\prime} \left(\frac{\mu_t}{\sigma_t} \right)^2 - \frac{2}{\sigma_t} \right] \cdot \frac{\sigmasp^2 \cdot \mu_t}{\gamma^2} \le 
    2 \rbb{\prime}{\sigma_t} \cdot \left(\frac{\mu_t}{\sigma_t} \right) - R^{\prime\prime} \left(\frac{\mu_t}{\sigma_t} \right)^2 
    \label{eq:stricter_necessary} \\
    \iff&   \left[ 2 \rb{\sigma_t} +  2\cdot R^{\prime\prime} \left(\frac{\mu_t}{\sigma_t} \right)^2 - \frac{2}{\sigma_t} \right] \cdot \frac{\sigmasp^2 }{\gamma^2} \le 
    2 \rbb{\prime}{\sigma_t} \cdot \left(\frac{1}{\sigma_t} \right) - R^{\prime\prime} \left(\frac{\mu_t}{\sigma_t^ 2} \right) \\
    \iff&   \left[ \rb{\sigma_t} +  R^{\prime\prime} \left(\frac{\mu_t}{\sigma_t} \right)^2 - \frac{1}{\sigma_t} \right] \cdot \frac{\sigmasp^2 }{\gamma^2} \le 
    \rb{\sigma_t} -  \frac{1}{\sigma_t} - \frac{R^{\prime\prime}}{2} \left(\frac{\mu_t}{\sigma_t^ 2} \right) \\ 
    \iff&   \left[R^{\prime\prime} \left(\frac{\mu_t}{\sigma_t} \right)^2  \right] \cdot \frac{\sigmasp^2 }{\gamma^2} + \frac{R^{\prime\prime}}{2} \left(\frac{\mu_t}{\sigma_t^ 2} \right)  \le 
    \left( \rb{\sigma_t} -  \frac{1}{\sigma_t} \right) \cdot \left( 1 - \frac{\sigmasp^2 }{\gamma^2}  \right) \\
    \iff&   \left[R^{\prime\prime} \left(\frac{\mu_t^2}{\sigma_t} \right)  \right] \cdot \frac{\sigmasp^2 }{\gamma^2} + \frac{R^{\prime\prime}}{2} \left(\frac{\mu_t}{\sigma_t} \right)  \le 
    \left(\sigma_t \rb{\sigma_t} -  1 \right) \cdot \left( 1 - \frac{\sigmasp^2 }{\gamma^2}  \right) \label{eq:final_st_necessary} 
\end{align}

Now, we will argue the monotonicity of LHS and RHS in \eqref{eq:final_st_necessary}. Observe that LHS is increasing in $\mu_t$ and decreasing in $\sigma_t$ and RHS is decreasing in $\sigma_t$ as $\left(\sigma_t \rb{\sigma_t} -  1 \right)$ is increasing (and the multiplier is negative). Moreover, if \eqref{eq:final_st_necessary} holds true for maximum value of RHS and minimum of LHS, then we would have \eqref{eq:ST_necessary_cond}.  Thus substituting $\mu_t = \gamma $ and $\sigma_t = \sigma_0$ in LHS and $\sigma_t = \sigmasp$ in RHS, we get: 
\begin{align}
    & \left[R^{\prime\prime} \left(\frac{\gamma^2}{\sigma_0} \right)  \right] \cdot \frac{\sigmasp^2 }{\gamma^2} + \frac{R^{\prime\prime}}{2} \left(\frac{\gamma}{\sigma_0} \right)  \le  \left(\sigmasp \rb{\sigmasp} -  1 \right) \cdot \left( 1 - \frac{\sigmasp^2 }{\gamma^2}  \right) \\
    \iff&  R^{\prime\prime} \cdot \frac{\sigmasp^2 }{\sigma_0} + \frac{R^{\prime\prime}}{2} \left(\frac{\gamma}{\sigma_0} \right)  \le  \left(\sigmasp \rb{\sigmasp} -  1 \right) \cdot \left( 1 - \frac{\sigmasp^2 }{\gamma^2}  \right) \\ 
\end{align}

Taking $\gamma \le \frac{1}{2\sqrt{\sigmasp}}$ and substituting $R^{\prime\prime} = \rbb{{\prime\prime}}{\sigma_0}$: 
\begin{align*}
    (5/4)\cdot \rbb{{\prime\prime}}{\sigma_0} \cdot \sigmasp  \le  \left(\sigmasp \rb{\sigmasp} -  1 \right) \cdot \left( 1 - 4\cdot\sigmasp^3  \right) \numberthis \label{eq:final_necessary_expr}
\end{align*}

Analytically solving the above expression, we get that \eqref{eq:final_necessary_expr} is satisfied for all values of $\sigmasp \ge 1$ when $\dsp \ge 1$. For example, the expression in \eqref{eq:final_necessary_expr} is also satisfied for the problem parameter used in the running example of the main paper. 

\end{proof}

As a remark, we note that in the proof of \lemref{lemma:cond_ST_necessary}, the conditions derived are loose because of the relaxations made to simply the proof. In principle, the proof (and hence the conditions) can be tightened by carefully propagating second-order terms (which depend on $\sigma_t$) in 
\eqref{eq:taylor_expr}.

\begin{lemma}[Sufficiency conditions  for ST]
Define $\alpha_1$ and $\alpha_2$ as in \eqref{eq:alpha_1} and \eqref{eq:alpha_2} respectively. If $\sigmasp \le \gamma$, then we have for all $t$: 
\begin{align}
  \alpha_2(\mu_t, \sigma_t) \cdot \frac{\sigmasp^2 \cdot \mu_t}{\gamma^2} \ge \alpha_1(\mu_t, \sigma_t) \,. \label{eq:ST_sufficiency_cond}
\end{align}
\label{lemma:cond_ST_sufficiency} 
\end{lemma}

\begin{proof}
    We upper bound and lower bound $\alpha_1$ and $\alpha_2$ by using the properties of $\rb{\cdot}$. 
Recall: 
\begin{align}
    \alpha_1(\mu_t, \sigma_t) = \sqrt{\frac{2}{\pi}} \expb{-\frac{\mu_t^2}{2\sigma_t^2}} \left[\rb{\sigma_t + \frac{\mu_t}{\sigma_t}}  - \rb{ \sigma_t - \frac{\mu_t}{\sigma_t}} \right] \,.
\end{align}
and 
\begin{align}
    \alpha_2(\mu_t, \sigma_t) = \sqrt{\frac{2}{\pi}} \expb{-\frac{\mu_t^2}{2\sigma_t^2}} \left[\rb{\sigma_t + \frac{\mu_t}{\sigma_t}}  + \rb{ \sigma_t - \frac{\mu_t}{\sigma_t}} - \frac{2}{\sigma_t} \right] \,.
\end{align}

We now use Taylor's expansion on $\rb{\cdot}$ and we get: 
\begin{align}
  \rb{\sigma_t} +  \rbb{\prime}{\sigma_t} \cdot \left(\frac{\mu_t}{\sigma_t} \right) \le \rb{\sigma_t + \frac{\mu_t}{\sigma_t}} \le \rb{\sigma_t} +  \rbb{\prime}{\sigma_t} \cdot \left(\frac{\mu_t}{\sigma_t} \right) + \rbb{{\prime\prime}}{\sigma_t} \cdot \left(\frac{\mu_t}{\sigma_t} \right)^2
\end{align}
and similarly, we get: 
\begin{align}
  \rb{\sigma_t} -  \rbb{\prime}{\sigma_t} \cdot \left(\frac{\mu_t}{\sigma_t} \right) + \rbb{{\prime\prime}}{\sigma_t} \cdot \left(\frac{\mu_t}{\sigma_t} \right)^2 \le \rb{\sigma_t - \frac{\mu_t}{\sigma_t}} \le \rb{\sigma_t} - \rbb{\prime}{\sigma_t} \cdot \left(\frac{\mu_t}{\sigma_t} \right) + R^{\prime\prime} \left(\frac{\mu_t}{\sigma_t} \right)^2 \label{eq:taylor_expr_repeat}
\end{align}

where $R^{\prime\prime} = \rbb{{\prime\prime}}{\sigma_0}$. This is because  $\rbb{{\prime\prime}}{\cdot}$ takes positive values and is a decreasing function in $\sigma_t$ (refer to \lemref{lemma:mill_ratio}).  
We now lower bound $\alpha_1(\mu_t, \sigma_t)$ and upper bound $\alpha_2(\mu_t, \sigma_t)$: 
\begin{align}
    \frac{\alpha_1(\mu_t, \sigma_t)}{ \sqrt{\frac{2}{\pi}} \expb{-\frac{\mu_t^2}{2\sigma_t^2}} } \le  2 \rbb{\prime}{\sigma_t} \cdot \left(\frac{\mu_t}{\sigma_t} \right)
\end{align}

\begin{align}
    \frac{\alpha_2(\mu_t, \sigma_t)}{ \sqrt{\frac{2}{\pi}} \expb{-\frac{\mu_t^2}{2\sigma_t^2}} } \ge  2 \rb{\sigma_t} +  \rbb{{\prime\prime}}{\sigma_t} \cdot \left(\frac{\mu_t}{\sigma_t} \right)^2 - \frac{2}{\sigma_t}
\end{align}

Substituting the lower bound and upper bound in \eqref{eq:ST_sufficiency_cond} gives us the following as stricter a sufficient condition (i.e., \eqref{eq:stricter_sufficiency} implies \eqref{eq:ST_sufficiency_cond}): 
\begin{align}
    & \left[ 2 \rb{\sigma_t} +  \rbb{{\prime\prime}}{\sigma_t} \cdot \left(\frac{\mu_t}{\sigma_t} \right)^2 - \frac{2}{\sigma_t} \right] \cdot \frac{\sigmasp^2 \cdot \mu_t}{\gamma^2} \ge 2 \rbb{\prime}{\sigma_t} \cdot \left(\frac{\mu_t}{\sigma_t} \right)
    \label{eq:stricter_sufficiency} \\
    \iff & \left[ 2 \rb{\sigma_t} +  \rbb{{\prime\prime}}{\sigma_t} \cdot \left(\frac{\mu_t}{\sigma_t} \right)^2 - \frac{2}{\sigma_t} \right] \ge 2 \rbb{\prime}{\sigma_t} \cdot \left(\frac{\mu_t}{\sigma_t} \right) \cdot \frac{\gamma^2}{\sigmasp^2 \cdot \mu_t} \\
    \iff & 2 \rb{\sigma_t} +  \rbb{{\prime\prime}}{\sigma_t} \cdot \left(\frac{\mu_t}{\sigma_t} \right)^2 - \frac{2}{\sigma_t} -  2 \rbb{\prime}{\sigma_t} \cdot \left(\frac{\mu_t}{\sigma_t} \right) \cdot \frac{\gamma^2}{\sigmasp^2 \cdot \mu_t}  \ge 0 \\
    \iff & 2 \rb{\sigma_t} \cdot \sigma_t +  \rbb{{\prime\prime}}{\sigma_t} \cdot \frac{\mu_t^2}{\sigma_t}  - 2 -  2 \rbb{\prime}{\sigma_t} \cdot \frac{\gamma^2}{\sigmasp^2}  \ge 0 \\
    \iff & 2 \rbb{\prime}{\sigma_t} +  \rbb{{\prime\prime}}{\sigma_t} \cdot \frac{\mu_t^2}{\sigma_t}  -  2 \rbb{\prime}{\sigma_t} \cdot \frac{\gamma^2}{\sigmasp^2}  \ge 0 \\
    \iff &  \rbb{{\prime\prime}}{\sigma_t} \cdot \frac{\mu_t^2}{\sigma_t}  + 2 \rbb{\prime}{\sigma_t}  \cdot \left[ 1 -  \frac{\gamma^2}{\sigmasp^2} \right] \ge 0
    \label{eq:stricter_sufficiency_2}    
\end{align}

Hence, when $ \left[ 1 -  \frac{\gamma^2}{\sigmasp^2} \right] \le 0$, we have condition in \eqref{eq:stricter_sufficiency_2} hold true as $\rbb{\prime}{\sigma_t}$ is always negative. Hence,  the condition $\gamma \ge \sigmasp$ gives us the necessary condition. 
\end{proof}

\subsubsection{Proof of Proposition \ref{prp:bt-closedform}}
For convenience, we first restate the Proposition~\ref{prp:bt-closedform} which gives us a closed form solution for \eqref{eq:cont-loss} when $\rho=0$. Then, we provide the proof, focusing first on the case of $k=1$, and then showing that extension to $k>1$ is straightforward and renders the final form in the proposition that follows.

\begin{proposition}[Barlow Twins solution]
\label{prp:bt-closedform-restated}
The solution for \eqref{eq:cont-loss} is $U_k^\top\Sigma_\mathsf{A}^{-1/2}$ where $U_k$ are the top $k$ eigenvectors of $\Sigma_\mathsf{A}^{-1/2}\, \tilde{\Sigma} \, \Sigma_\mathsf{A}^{-1/2}$. Here,  $\Sigma_\mathsf{A} \coloneqq \Exp_{a\sim\ProbA}[aa^\top]$ is the covariance over augmentations, and $\tilde{\Sigma} \coloneqq \Exp_{x\sim\ProbU} [\tilde{a}(x) \tilde{a}(x)^\top]$ is the covariance matrix of mean augmentations $\tilde{a}(x) \coloneqq \Exp_{\ProbA(a \mid x)}[a]$. 
\end{proposition}

\begin{proof}

We will use $\phi(x)$ to denote $\phi^\top x$ where $\phi \in \R^d$. Throughout the proof, we use $a$ to denote augmentation and $x$ to denote the input. We will use $\ProbA(a \mid x)$ as the probability measure over the space of augmentations $\calA$, given some input $x\in \calX$ (with corresponding density) $p_\mathsf{A}(\cdot \mid x)$. Next, we use $p_\mathsf{A}(\cdot)$ to denote the density associate with the marginal probability measure over augmentations: $\ProbA = \int_\calX \ProbA(a \mid x) \mathrm{d}\ProbU$. Finally, the joint distribution over positive pairs $A_+(a_1, a_2) = \int_{\calX} \ProbA(a_1 \mid x) \ProbA(a_2 \mid x)  \mathrm{d}\ProbU$, gives us the positive pair graph over augmentations.

Before we solve the optimization problem in \eqref{eq:cont-loss} for $\Phi \in \R^{k \times d}$ for any general $k$, let us first consider the case where $k=1$, \ie we only want to find a single linear projection $\phi$. The constraint $\rho=0$, transfers onto $\phi$ in the following way:
\begin{align}
    \E_{a \sim \ProbA} [\phi(a)^2] = 1 \quad \equiv \quad \phi^\top\Sigma_A\phi = 1
\end{align}

Under the above constraint we want to minimize the invariance loss, which according to Lemma~\ref{lem:inv-loss-operator} is given by $2\cdot \int_{\calA} \phi(a) L(\phi)(a) \; \mathrm{d}\ProbA $, where $L(\phi)(\cdot)$ is the following linear operator. 

\begin{align}
    L(\phi)(a) = \phi(a) - \int_{\calA} \frac{A_+(a, a')}{p_\mathsf{A}(a)}\cdot \phi(a') \; \mathrm{d}a'.
\end{align}
Based on the definition of the operator, we can reformulate the constrained optimization for contrastive pretraining as:

\begin{align}
    & \argmin_{\phi: \phi^\top\Sigma_A \phi = 1} \;\; \int_{\calA} \phi(a)\cdot L(\phi)(a) \; \mathrm{d}\ProbA \\
    & \implies \argmin_{\phi: \phi^\top\Sigma_A \phi = 1} \;\;  \E_{a \sim \ProbA} [\phi(a)^2] -  \int_\calA \int_\calA \phi(a)\cdot \phi(a') \cdot A_+(a, a') \; \mathrm{d}a \mathrm{d}a'  \\
    & \implies \argmin_{\phi: \phi^\top\Sigma_A \phi = 1} \;\;  \E_{a \sim \ProbA} [\phi(a)^2] -  \int_\calX\int_\calA\int_\calA p_\mathsf{A}(a \mid x) p_\mathsf{A}(a' \mid x) \cdot \phi(a) \phi(a')  \; \mathrm{d}\ProbU \\
    & \implies \argmin_{\phi: \phi^\top\Sigma_A \phi = 1} \;\;  \E_{a \sim \ProbA} [\phi(a)^2] -  \int_\calX  [\tilde{\phi}(x)]^2  \; \mathrm{d}\ProbU,
\end{align}

where $\tilde{\phi}(x) = \Exp_{a \sim \ProbA(\cdot \mid x)} \phi(x) = \Exp_{c \sim \unif[0, 1]^d} [\phi^\top(c \odot x)]$. Note that,

\begin{align}
    \tilde{\phi}(x)^2 &= \paren{\Exp_{c \sim \unif[0, 1]^d} [\phi^\top(c \odot x)]}^2 \\
    &= \phi^\top (\Exp_{c \sim \unif[0, 1]^d} [c \odot x]) (\Exp_{c \sim \unif[0, 1]^d} [c \odot x])^\top \phi \\
    \implies   \int_\calX  [\tilde{\phi}(x)]^2 \; \mathrm{d}\ProbU &= \phi^\top \tilde{\Sigma} \phi
\end{align}

Further, since $ \E_{a \sim \ProbA} [\phi(a)^2]  = \phi^\top \Sigma \phi$ we can now rewrite our main optimization problem for $k=1$ as:

\begin{align}
    &\argmin_{\phi: \phi^\top\Sigma_A \phi = 1} \;\; \phi^\top \Sigma_A \phi -  \phi^\top \tilde{\Sigma} \phi \\ 
     &= \argmax_{\phi: \phi^\top\Sigma_A \phi = 1}  \phi^\top \tilde{\Sigma} \phi
\end{align}

Recall that in our setup both $\tilde{\Sigma}$ and $\Sigma_A$ are positive definite and invertible matrices. To solve the above problem, let's consider a re-parameterization: $\phi' = \Sigma_A^{1/2} \phi$, thus $\phi^\top \Sigma_A \phi = 1$, is equivalent to the constraint $\|\phi'\|_2^2 = 1$. Based on this re-parameterization we are now solving:

\begin{align}
    \argmax_{\|\phi'\|_2^2=1} \;\;  \phi'^\top\Sigma_A^{-1/2} \cdot \tilde{\Sigma} \cdot \Sigma_A^{-1/2} \phi',
\end{align}

which is nothing but the top eigenvector for $\Sigma_A^{-1/2} \cdot \tilde{\Sigma} \cdot \Sigma_A^{-1/2}$. 

Now, to extend the above argument from $k=1$ to $k>1$, we need to care of one additional form of constraint in the form of feature diversity: $\phi_i^\top \Sigma_A \phi_j = 0$ when $i\neq j$. But, we can easily redo the reformulations above and arrive at the following optimization problem:

\begin{align}
    \argmax_{
    \small
    \begin{array}{cc}
         \|\phi_i'\|_2^2=1,\;\; \forall i  \\
         \phi_i'^\top \phi_j' = 0,\;\; \forall i\neq j  
    \end{array}
    } \;\;  \brck{\phi'_1, \phi'_2, \ldots, \phi'_k}^\top\Sigma_A^{-1/2} \cdot \tilde{\Sigma} \cdot \Sigma_A^{-1/2} \brck{\phi'_1, \phi'_2, \ldots, \phi'_k},
\end{align}

where $\phi_i' = \Sigma_{A}^{1/2} \phi_i$. The above is nothing but the top $k$ eigenvectors for the matrix $\Sigma_A^{-1/2} \cdot \tilde{\Sigma} \cdot \Sigma_A^{-1/2}$. This completes the proof of Proposition~\ref{prp:bt-closedform-restated}.   
\end{proof}
\vspace{3em}

\subsubsection{Analysis with \texorpdfstring{$\rho > 0$}{rho greater than zero}  in Contrastive Pretraining Objective \texorpdfstring{\eqref{eq:cont-loss}}{eq:cont-loss}}
\label{appsubsec:bt-optimization}

In \eqref{eq:cont-loss} we considered the strict version of the optimization problem where $\rho = 0$. Here, we will consider the following optimization problem that we optimize for our experiments in the simplified setup:

\begin{align}
    \footnotesize
    \calL_\cl(\Phi, \kappa) \; \coloneqq \; \Exp_{x \sim \ProbU}\Exp_{a_1, a_2 \sim \ProbA(\cdot \mid x)} \; \|\Phi(a_1) - \Phi(a_2)\|_2^2 + \kappa \cdot
    \norm{\Exp_{a \sim \ProbA}\brck{\Phi(a)\Phi(a)^\top} - \mathbf{I}_k}{F}^2,
     \label{eq:cont-loss-with-kappa}
\end{align}

where $\kappa > 0$ is some finite constant (note that every $\rho$ corresponds to some $\kappa$ and particularly $\rho = 0$, corresponds to $\kappa = \infty$). 
Let $\Phi^\star$ be the solution for \eqref{eq:cont-loss} with $\rho = 0$, \ie the solution described in Proposition~\ref{prp:bt-closedform}.
Now, we will show that in practice we can provably recover something close to $\Phi^\star$ when $\kappa$ is large enough. 

\begin{theorem}[Solution for \eqref{eq:cont-loss-with-kappa} is approximately equal to $\Phi^\star$] If $\hat{\Phi}$ is some solution that achieves low values of the objective $\calL_\cl(\Phi, \kappa)$ in \eqref{eq:cont-loss-with-kappa}, \ie,
$\calL_\cl(\hat{\Phi}, \kappa) \leq \epsilon$, then there exists matrix $W \in \R^{k \times k}$ such that:
\begin{align*}
    & \Exp_{a \sim \ProbA} \|W\cdot \Phi^\star(a) - \hat{\Phi}(a)\|_2^2 \leq \frac{k\epsilon}{2\gamma_{k+1}}, \\
    & \qquad \;\; \textrm{where,} \;\; \gamma_{k+1} \geq \frac{2\gamma_1^2}{k\epsilon} \cdot \paren{1-\sqrt{\frac{\epsilon}{\kappa}}} - \frac{\gamma_1}{k}, 
\end{align*}
where $\gamma_{k+1}$ is the the ${k+1}^{th}$ eigenvalue for $\mathbf{I}_d - \Sigma_A^{-1/2} \; \Tilde{\Sigma}\;\Sigma_A^{-1/2}$. Here, $\lambda_1 \leq \lambda_2 \leq \ldots \leq \lambda_d$. 
\label{thm:bt-optimization-result}
\end{theorem}

\begin{proof}

Since we know that $\calL_\cl(\hat{\Phi}, \kappa) \leq \epsilon$, we can individually bound the invariance loss and the regularization term:
\begin{align}
    \Exp_{x \sim \ProbU}\Exp_{a_1, a_2 \sim \ProbA(\cdot \mid x)} \; \|\hat{\Phi}(a_1) - \hat{\Phi}(a_2)\|_2^2 \leq \epsilon \\
    \norm{\Exp_{a \sim \ProbA}\brck{\hat{\Phi}(a)\hat{\Phi}(a)^\top} - \mathbf{I}_k}{F}^2 \leq \frac{\epsilon}{\kappa}
\end{align}
Thus,
\begin{align}
  &  \forall i\in[k]: \;\; 1- \sqrt{\frac{\epsilon}{\kappa}} \leq \hat{\phi}_i^\top \; \Sigma_A \; \hat{\phi}_i \leq 1 + \sqrt{\frac{\epsilon}{\kappa}} \\ 
  &  \forall i\in[k]: \;\; \Exp_{x\sim\ProbU}\Exp_{a_1, a_2 \sim \ProbA(\cdot \mid x)} (\hat{\phi}_i^\top a_1 - \hat{\phi}_i^\top a_2)^2 \leq \epsilon
\end{align}

Let $\phi^\star_1, \phi^\star_2, \phi^\star_3, \ldots, \phi^\star_d$ be the solution returned by the analytical solution for $\rho=0$, \ie the solution in Proposition~\ref{prp:bt-closedform}. Now, since $\Phi^\star$ would span $\R^{d}$ when $\Sigma_A$ is full rank, we can denote:
\begin{align}
    \hat{\phi}_i = \sum_{j=1}^{d} \eta^{(j)}_i \phi_j^\star
\end{align}

Now from Lemma~\ref{lem:inv-loss-operator}, the invariance loss for $\hat{\phi}_i$ can be written using the operator $L(\phi)(a) = \phi(a) - \int_\calA \frac{A_+(a, a')}{p_\mathsf{A}(a)} \phi(a')  \;\mathrm{d}a'$:

\begin{align}
    \textrm{Invariance Loss}(\hat{\phi}_i) &\coloneqq \Exp_{x\sim\ProbU}\Exp_{a_1, a_2 \sim \ProbA(\cdot \mid x)} (\hat{\phi}_i^\top a_1 - \hat{\phi}_i^\top a_2)^2 \\
    &= 2 \cdot \Exp_{a\sim \ProbA} [\hat{\phi}_i(a) L(\hat{\phi}_i)(a)] \\
    &= 2 \cdot \Exp_{a\sim \ProbA} \brck{\paren{ \sum_{j=1}^{d} \eta^{(j)}_i \phi_i^\star} L\paren{ \sum_{j=1}^{d} \eta^{(j)}_i \phi_j^\star}(a)} \\
        &= 2 \cdot \Exp_{a\sim \ProbA} \brck{\paren{ \sum_{j=1}^{d} \eta^{(j)}_i \phi_j^\star} \paren{ \sum_{j=1}^{d} \eta^{(j)}_i L(\phi_j^\star)(a)}} \\
        &= 2\cdot \sum_{j=1}^d \paren{\eta_i^{(j)}}^2  \Exp_{a\sim \ProbA} \brck{\phi_j^\star(a)L(\phi_j^\star)(a)} \\
        & \quad + 2\cdot \sum_{m=1,n=1,m\neq n}^{d} \eta_i^{(m)}\eta_i^{(n)} \Exp_{a\sim \ProbA} \brck{\phi_m^\star(a)L(\phi_n^\star)(a)}  
\end{align}

Since, $\phi_i^\star(\cdot)$ are eigenfunctions of the operator $L$~\citep{haochen2022theoretical}, we can conclude that:
\begin{align*}
    \sum_{m=1,n=1,m\neq n}^{d} \eta_i^{(m)}\eta_i^{(n)} \Exp_{a\sim \ProbA} \brck{\phi_m^\star(a)L(\phi_n^\star)(a)}  = 0,
\end{align*}
and if $\gamma_1 \leq \gamma_2 \leq \gamma_3 \ldots \leq  \gamma_d$ are the eigenvalues for $\phi_1^\star, \phi_2^\star, \phi_3^\star, \ldots, \phi_d^\star$ under the decomposition of $L(\phi)(\cdot)$ then:

\begin{align}
    \Exp_{x\sim\ProbU}\Exp_{a_1, a_2 \sim \ProbA(\cdot \mid x)} (\hat{\phi}_i^\top a_1 - \hat{\phi}_i^\top a_2)^2 = 2\cdot \sum_{j=1}^d \gamma_j \paren{\eta_i^{(j)}}^2  
\end{align}

Recall, we are also aware of a condition on the regularization term:
$1- \sqrt{\frac{\epsilon}{\kappa}} \leq \hat{\phi}_i^\top \; \Sigma_A \; \hat{\phi}_i \leq 1 + \sqrt{\frac{\epsilon}{\kappa}}$. 

\begin{align}
    \hat{\phi}_i^\top \; \Sigma_A \; \hat{\phi}_i &= \paren{\sum_{j=1}^{d} \eta^{(j)}_i \phi_j^\star}^\top \; \Sigma_A \; \paren{\sum_{j=1}^{d} \eta^{(j)}_i \phi_j^\star}  
    = \sum_{j=1}^{d} \paren{\eta^{(j)}_i}^2 \\
    & \quad \implies 1- \sqrt{\frac{\epsilon}{\kappa}} \leq \sum_{j=1}^{d} \paren{\eta^{(j)}_i}^2 \leq 1+ \sqrt{\frac{\epsilon}{\kappa}} \;\; \forall i. 
\end{align}

In order to show that the projection of $\hat{\phi}_i$ on $\Phi^*$ is significant, we need to argue that the term $\sum_{j={k+1}}^d \paren{\eta_i^{(j)}}^2$ is small. The argument for this begins with the condition on invariance loss, and the fact that $\gamma_1 \leq \gamma_2 \leq \ldots \leq \gamma_k \leq \gamma_{k+1} \leq \ldots \leq \gamma_d$:

\begin{align}
&    \frac{\epsilon}{2} \geq \sum_{j=k+1}^d \paren{\eta^{(j)}_i}^2 \gamma_j \geq \gamma_{k+1} \cdot \paren{\sum_{j=k+1}^d \paren{\eta^{(j)}_i}^2} \\
 &  \quad  \implies  \sum_{j=k+1}^d \paren{\eta^{(j)}_i}^2 \leq \frac{\epsilon}{2\gamma_{k+1}} 
\end{align}

Extending the above result $\forall i$ by simply adding the bounds completes the claim of our first result in Theorem~\ref{thm:bt-optimization-result}. Next, we will lower bound the eigenvalue $\gamma_{k+1}$. 
Recall that, $\sum_{j=1}^k \paren{\eta^{(j)}_i}^2 \geq 1 - \sqrt{\frac{\epsilon}{\kappa}} - \frac{\epsilon}{2\gamma_{k+1}}$. Thus,

\begin{align}
    \gamma_1 \cdot \paren{1 - \sqrt{\frac{\epsilon}{\kappa}} - \frac{\epsilon}{2\gamma_{k+1}}} \leq \sum_{j=1}^k \gamma_j \paren{\eta_i^{(j)}}^2 \leq k  \gamma_{k+1} \cdot \frac{\epsilon}{2\gamma_1}
\end{align}

We assume that all eigenvalues are strictly positive, which is true under our augmentation distribution. Given, $\gamma_{k+1} \geq \gamma_1$, we can rearrange the above to get:

\begin{align}
\gamma_{k+1} \geq \frac{2\gamma_1^2}{k\epsilon} \cdot \paren{1-\sqrt{\frac{\epsilon}{\kappa}}} - \frac{\gamma_1}{k}      
\end{align}
This completes the claim of our second result in Theorem~\ref{thm:bt-optimization-result}. 
\end{proof}
\vspace{3em}

\subsubsection{Proof of Theorem \ref{thm:bt-blockform}}

In this section, we prove our main theorem about the recovery of both spurious $\wsp$, invariant $\winv$ features by the contrastive learning feature backbone, and also the amplification of the invariant over the spurious feature (where amplification is defined relatively with respect to what is observed in the data distribution alone). We begin by defining some quantities needed for analysis, that are fully determined by the choice of problem parameters for the model in \eqref{eq:toy-model-main-paper}.

From Section~\ref{sec:theory}, we recall the definitions of $\winv \coloneqq \brck{w^\star, 0, \ldots, 0}$ and $\wspu \coloneqq \brck{0, \ldots 0, w'}$ where $w' = \mathbf{1}_{\dsp}/\sqrt{\dsp}.$
Let us now define $u_1, u_2$ as the top two eigenvectors of $\Sigma_A$ with eigenvalues $\lambda_1, \lambda_2 > 0$, (note that in our problem setup both $\Sigma_A$ and $\tilde{\Sigma}$ are full rank positive definite matrices), and $\tau \coloneqq \sqrt{\lambda_1/\lambda_2}$. Next we define $\alpha$ as the angle between $u_1$ and $\winv$, \ie, $\cos(\alpha) = u_1^\top \winv$. Based on the definitions of $\alpha$ and $\tau$, both of which are fully determined by the eigen decomposition of the post-augmentation feature covariance matrix $\Sigma_A$, we now restate  Theorem~\ref{thm:bt-blockform}:

\begin{theorem}[Formal; CL recovers both invariant $\winv$ and spurious $\wspu$ but amplifies $\winv$]
\label{thm:bt-blockform-formal}
Under Assumption~\ref{assm:augs} $(\wstar = \nicefrac{\mathbf{1}_{d_\mathrm{in}}}{\sqrt{\din}})$, the CL solution $\Phi_\cl$$=$$\brck{\phi_1, \phi_2, ..., \phi_k}$ satisfies $\phi_j^\top \winv = \phi_j^\top 
\wspu=0$ $\forall j\geq 3$. For $\tau, \alpha$ as defined above, the solution for $\phi_1, \phi_2$ is:
$$
\begin{bmatrix}
\wstar \cdot \cot(\alpha)/\tau ,\;\;\; \hfill \wstar \\ 
 w' \cdot 1 / \tau ,\;\;\; \hfill \;\;\;\;\;\;\;\; w' \cdot \cot(\alpha)
\end{bmatrix}\;\; \cdot \;\;
\begin{bmatrix}
\cos{\theta},\;\; \hfill \sin{\theta} \\ 
\sin{\theta},\;\; -\cos{\theta}
\end{bmatrix},$$
where $0 \leq \alpha, \theta \leq \pi/2$. Let us redefine $\phi_1 = c_1 \winv + c_3 \wspu$ and $\phi_2 = c_2 \winv + c_4 \wspu$.

For constants $K_1, K_2 >0$, $\gamma = \nicefrac{K_1K_2}{\sigmasp}$, $\dsp = \nicefrac{\sigmasp^2}{K_2^2}$, $\forall \epsilon > 0$, $\exists {\sigmasp}_{0}$, such that for $\sigmasp \geq {\sigmasp}_{0}$:
\begin{align*}
    \frac{K_1K_2^2\din}{2L\sigmain^2 ({\din}-1)} + \epsilon \; \geq \; &\;\,\frac{c_1}{c_3} \;  \geq \; \frac{K_1K_2^2\din}{2L\sigmain^2 ({\din}-1)} - \epsilon \\
 \frac{L\sqrt{\dsp}}{\gamma} + \epsilon  \; \geq \;  &\abs{ \frac{c_2}{c_4}} \; \geq \; \frac{L\sqrt{\dsp}}{\gamma} - \epsilon, 
\end{align*}
 where $L = 1 + {K_2^2}$.
\end{theorem}

\begin{proof}
We will first show that the only components of interest are $\phi_1, \phi_2$. Then, we will prove conditions on the amplification of $\winv$ over $\wspu$ in $\phi_1, \phi_2$. Following is the proof overview:

\begin{enumerate}[label=\Roman{*}.]
    \item  When $w^\star = \mathbf{1}_{\din}/\sqrt{\din}$, from the closed form expressions for $\Sigma_A$ and $\tilde{\Sigma}$, show that the solution returned by solving the Barlow Twins objective depends on $\winv$ and $\wspu$ only through the first two components $\phi_1, \phi_2$.
    \item For the components $\phi_1, \phi_2$, we will show that the dependence along $\winv$ is amplified compared to $\wspu$ when the target data sufficiently denoises the spurious feature (\ie, $\sigmasp$ is sufficiently large). 
\end{enumerate}

\textbf{Part-I:}

We can divide the space $\R^d$ into two subspaces that are perpendicular to each other. The first subspace is $\calW = \{b_1 \cdot \winv + b_2 \cdot \wspu : b_1, b_2 \in \Real\}$, \ie the rank $2$ subspace spanned by $\winv$ and $\wspu$. The second subspace is $\calW_\perp$ where $\calW_\perp = \{u \in \Real^d : u^\top \winv = 0, u^\top \wspu = 0 \}$. Then, from Lemma~\ref{lem:bt-covariance-zero} we can conclude that the matrix $\Sigma_A$ can be written as:

\begin{align}
    \label{eq:sigmaA}
    \Sigma_A  &=  {\Sigma}_{A_\calW}  +  {\Sigma}_{A_{\small {\calW}_\perp}} \nonumber \\ 
    \Sigma_{A_\calW} &= \frac{1}{4} \begin{bmatrix}
        \paren{\gamma^2 (1+\nicefrac{1}{3\din}) + \nicefrac{\sigmain^2}{3}(1-\nicefrac{1}{\din})}    \cdot \wstar \wstar^\top, \;\;\;\;\;\;\;\;\;\;\;\;\;\;\;\;\;\;\;\;\; \hfill \nicefrac{\gamma \sqrt{\dsp}}{2} \cdot \wstar w'^\top \\ 
        \nicefrac{\gamma \sqrt{\dsp}}{2} \cdot w'\wstar^\top, \;\;\;\;\;\;\;\;\;\; \hfill \paren{\nicefrac{\dsp}{2} + \nicefrac{4}{3} \cdot \sigmasp^2 + \nicefrac{1}{6}} \cdot w'w'^\top
    \end{bmatrix}, 
\end{align}
where ${\Sigma}_{A_{\small {\calW}_\perp}} \coloneqq \E_{a \sim \ProbA} \brck{\Pi_{\calW_\perp} (a) (\Pi_{\calW_\perp}(a))^\top}$ is the covariance matrix in the null space of $\calW$, and $\Pi_{\calW_\perp} (a)$ is the projection of augmentation $a$ into the null space of $\calW$,  \ie the covariance matrix in the space of non-predictive (noise) features. Similarly we can define:

\begin{align}
    \label{eq:tilde-sigma}
    \tilde{\Sigma} \;\;\; &= \;\;\; \tilde{\Sigma}_{\calW} \;\; + \;\; \tilde{\Sigma}_{{\small {\calW}_\perp}}  \nonumber \\
    \tilde{\Sigma}_{\calW}\;\; &= \;\; \frac{1}{4} \begin{bmatrix}
        \gamma^2  \cdot \wstar \wstar^\top, \hfill \nicefrac{\gamma \sqrt{\dsp}}{2} \cdot \wstar w'^\top \\ 
        \nicefrac{\gamma \sqrt{\dsp}}{2} \cdot w'\wstar^\top, \;\;\;\;\;\; \hfill \paren{\nicefrac{\dsp}{2} + \nicefrac{\sigmasp^2}{2}} \cdot w'w'^\top
    \end{bmatrix}
\end{align}

Here again $\tilde{\Sigma}_{{\small {\calW}_\perp}} \coloneqq \E_{x \sim \ProbU} \brck{\Pi_{\calW_\perp} (\Exp_{c\sim\unif[0,1]^d} (c \odot x)) (\Pi_{\calW_\perp}(\Exp_{c\sim\unif[0,1]^d} (c \odot x)))^\top}$ is the covariance matrix of mean augmentations after they are projected onto the null space of predictive features. The above decomposition also follows from result in Lemma~\ref{lem:bt-covariance-zero}. 

From Proposition~\ref{prp:bt-closedform}, the closed form expression for the solution returned by optimizing the Barlow Twins objective in \eqref{eq:cont-loss} is $U^\top \Sigma_A^{-1/2}$ where $U$ are the top-k eigenvectors of:
\begin{equation}
    \Sigma_A^{-1/2} \cdot \tilde{\Sigma} \cdot {\Sigma_A^{-1/2}} 
\end{equation}

When $\wstar = \mathbf{1}_{\din}/\sqrt{\din}$, then ${\Sigma}_{A_{\small {\calW}_\perp}} = \tilde{\Sigma}_{{\small {\calW}_\perp}} + B$ where $B$ is a diagonal matrix with diagonal given by $ \frac{1}{3}\cdot\mathrm{diag}(\tilde{\Sigma}_{{\small {\calW}_\perp}})$. Further, since $\mathrm{diag}(\tilde{\Sigma}_{{\small {\calW}_\perp}}) = p \cdot \mathbbm{1}_d$ for some constant $p > 0$, the eigenvectors of $\tilde{\Sigma}_{{\small {\calW}_\perp}}$ and ${\Sigma}_{A_{\small {\calW}_\perp}}$ are exactly the same. Hence, when we consider the SVD of the expression $\Sigma_A^{-1/2}\tilde{\Sigma}\Sigma_A^{-1/2}$, the matrices ${\Sigma}_{A_{\small {\calW}_\perp}}$ and $\tilde{\Sigma}_{{\small {\calW}_\perp}}$ have no effect on the SVD components that lie along the span of the predictive features. In fact, we only need to consider two  rank 2 matrices (first terms in \eqref{eq:tilde-sigma}, \eqref{eq:sigmaA}) and only do the SVD of $\Sigma_{A_\calW}^{-1/2} \cdot \tilde{\Sigma}_\calW \cdot \Sigma_{A_\calW}^{-1/2}$. 

There are only two eigenvectors of $\Sigma_{A_\calW}^{-1/2} \cdot \tilde{\Sigma}_\calW \cdot \Sigma_{A_\calW}^{-1/2}$. We use $\lambda_1, \lambda_2$ to denote the eigenvalues of $\Sigma_{A_\calW}$, and $\brck{\cos(\alpha)w^\star, \sin(\alpha)w'}^\top$, $\brck{\sin(\alpha)w^\star, -\cos(\alpha)w'}^\top$ for the corresponding eigenvectors. Similarly, we use $\tilde{\lambda}_1, \tilde{\lambda}_2$ to denote the eigenvalues of $\tilde{\Sigma}_{\calW}$, and  $\brck{\cos(\beta)w^\star, \sin(\beta)w'}^\top$, $\brck{\sin(\beta)w^\star, -\cos(\beta)w'}^\top$ for the corresponding eigenvectors. 
Let $\mathrm{SVD}_U(\cdot)$ denote the operation of obtaining the singular vectors of a matrix.
Then, to compute the components of the final expression: $\mathrm{SVD}_U(\Sigma_A^{-1/2}\tilde{\Sigma}\Sigma_A^{-1/2})^\top\Sigma_A^{-1/2}$ that lies along the span of predictive features (in $\calW$), we need only look at the decomposition of the following matrix:

\begin{align}
    \begin{bmatrix}
    \cos{\theta}\;\;,\hfill \sin(\theta) \\
    \sin{\theta}\;\;, \hfill -\cos(\theta)
    \end{bmatrix} = \mathrm{SVD}_U \paren{
    \begin{bmatrix}
        \nicefrac{1}{\sqrt{\lambda_1}}, \;\; 0 \\
        0, \;\; \nicefrac{1}{\sqrt{\lambda_2}}
    \end{bmatrix} \; \cdot \;
    \begin{bmatrix}
        \cos(\alpha-\beta), \hfill \;\; \sin(\alpha-\beta) \\
        \sin(\alpha-\beta), \hfill \;\; -\cos(\alpha-\beta)
    \end{bmatrix} \; \cdot \;
    \begin{bmatrix}
        \sqrt{\tilde{\lambda}_1}, \;\; 0 \\
        0, \;\; \sqrt{\tilde{\lambda}_2}
    \end{bmatrix}}
    \label{eq:final-svd-matrix}
\end{align}

Based on the above definitions of $\theta, \alpha, \lambda_1, \lambda_2$, we can then formulate $\phi_1$ and $\phi_2$ in the following way:

\begin{align}
    [\phi_1, \phi_2] = \begin{bmatrix}
        \wstar \cdot \frac{\cos(\alpha)}{\sqrt{\lambda_1}}, \hfill \;\; \wstar \cdot \frac{\sin(\alpha)}{\sqrt{\lambda_2}} \\
        w' \cdot \frac{\sin(\alpha)}{\sqrt{\lambda_1}}, \hfill \;\;  w' \frac{-\cos(\alpha)}{\sqrt{\lambda_2}}
    \end{bmatrix} \cdot \begin{bmatrix}
    \cos{\theta}\;\;,\hfill \sin(\theta) \\
    \sin{\theta}\;\;, \hfill -\cos(\theta)
    \end{bmatrix}
\end{align}

To summarize, using arguments in Lemma~\ref{lem:bt-covariance-zero} and the fact that $\wstar = \mathbf{1}_{\din} /\sqrt{\din}$, we can afford to focus on just two rank two matrices $\Sigma_{A_\calW}, \tilde{\Sigma}_\calW$ in the operation: $\mathrm{SVD}_U(\Sigma_A^{-1/2}) \tilde{\Sigma} \Sigma_A^{-1/2}$. The other singular vectors from the SVD only impact directions that span $\calW_\perp$, and the singular vectors obtained by considering only the rank 2 matrices lie only in the space of $\calW$.

\textbf{Part-II:}

From the previous part we obtained forms of $\phi_1, \phi_2$ in terms of: $\lambda_1, \lambda_2, \alpha, \theta$, all of which are fully specified by the SVD of $\Sigma_{A_\calW}$ and $\tilde{\Sigma}_{\calW}$. If we define $\tau \coloneqq \frac{\sqrt{\lambda_1}}{\sqrt{\lambda_2}}$, we can evaluate $c_1, c_2, c_3, c_4$ as:

\begin{align}
    c_1 &= \frac{\cot(\alpha)}{\tau} + \tan(\theta) \label{eq:c1-expr}  \\
    c_2 &= -1 +\frac{\cot(\alpha)\tan(\theta)}{\tau} \label{eq:c2-expr} \\
    c_3 &= \frac{1}{\tau} - \cot(\alpha)\tan(\theta) \label{eq:c3-expr}  \\
    c_4 &= \frac{\tan(\theta)}{\tau} + \cot(\alpha) \label{eq:c4-expr} 
\end{align}

Now, we are ready to begin proofs for our claims on the amplification factors, \ie on the ratios $c_1/c_3$, $|c_2 / c_4|$. 

We will first prove some limiting conditions for $c_1/c_3$, followed by those on $|c_2/c_4|$. For each of these conditions we will rely on the forms for $c_1, c_2, c_3, c_4$ derived in the previous part, in terms of $\alpha, \theta, \tau$ (where $0 \leq \alpha, \theta \leq \pi/2$). We will also rely on some lemmas that characterize the asymptotic behavior of $\alpha, \theta$ and $\tau$ as we increase $\sigmasp$.
We defer the full proof of these helper lemmas to later sections. 

\textbf{Asymptotic behavior of $c_1/c_3$.}

From Lemma~\ref{lem:asym-tau-tantheta} and Lemma~\ref{lem:asym-cot-alpha-and-tan-theta}, when $\gamma = \nicefrac{K_1}{\sqrt{z}}$ and $\sigmasp = K_2 \sqrt{z}$, then:

\begin{align}
    \lim_{z \rightarrow \infty} \frac{c_1}{c_3} =  \frac{\cot{\alpha} + \tau \tan\theta}{ 1 - \tau \cot{\alpha}\tan\theta} =   \lim_{z \rightarrow \infty} \tau \tan \theta = \frac{K_1K_2^2}{(1+K_2^2) 2 \sigmain^2 (1-\nicefrac{1}{\din})},
    \label{eq:c1/c3}
\end{align}

where we apply Moore-Osgood when applying limits on  intermediate forms. We can do this since $\tau \tan \theta$ approaches a constant, and each of $\cot \alpha, \tau$ and $\tan \theta$ are continuous and smooth functions of $z$ (see Lemma~\ref{lem:svd-approximations}).

\textbf{Asymptotic behavior of $\abs{\nicefrac{c_2}{c_4}}$.}

When we consider the limiting behavior of $\nicefrac{c_2}{c_4 z}$, as we increase $z$ or equivalently $\sigmasp$ when $\gamma = \nicefrac{K_1}{\sqrt{z}}$ and $\sigmasp = K_2 \sqrt{z}$, then we get:

\begin{align}
    \lim_{z \rightarrow \infty} \abs{\frac{c_2}{c_4z}} = \abs{\frac{-1 + \cot(\alpha) \tan(\theta)}{\frac{\tan(\theta) z}{\tau} + \cot(\alpha)z}}.
    \label{eq:c2/c4z}
\end{align}

From Lemma~\ref{lem:asym-cot-alpha-and-tan-theta}, $\cot \alpha \tan \theta \rightarrow 0$.
Next, if we consider $\lim_{z \rightarrow \infty} \nicefrac{z \tan \theta}{\tau} = \lim_{z \rightarrow \infty} \tau \tan \theta \cdot \nicefrac{z}{\tau^2}$. 
For $\nicefrac{z}{\tau^2}$, we invoke Lemma~\ref{lem:asym-z-by-tausq}, which states that when $\gamma = \nicefrac{K_1}{\sqrt{z}}$ and $\sigmasp = K_2 \sqrt{z}$, then:

\begin{align}
    \lim_{z \rightarrow \infty} \frac{z}{\tau^2} = \frac{2\nicefrac{\sigmain^2}{3}(1-\nicefrac{1}{\din})}{1+\nicefrac{4}{3} K_2^2}.
    \label{eq:z-by-tausq}
\end{align}

Further, in our bound on $c_1/c_3$, we derived that $\tau \tan \theta \rightarrow \nicefrac{K_1K_2^2}{(1+K_2^2) 2 \sigmain^2 (1-\nicefrac{1}{\din})}$. Once again using Moore-Osgood we can plug this along with \eqref{eq:z-by-tausq} to get:

\begin{align}
    \lim_{z \rightarrow \infty} \frac{\tan(\theta) z}{\tau} = \frac{K_1 K_2^2}{(1+K_2^2)(3+4K_2^2)}. \label{eq:tan-theta-z-by-tau}
\end{align}

Finally, from Lemma~\ref{lem:asym-cot-alpha-z}, when $\gamma = \nicefrac{K_1}{\sqrt{z}}$ and $\sigmasp = K_2 \sqrt{z}$, then:
\begin{align}
    \lim_{z \rightarrow \infty} \frac{z}{\tan\alpha} = \frac{K_1}{(1+\nicefrac{4}{3}K_2^2)}.
    \label{eq:z-cot-alpha}
\end{align}

Plugging, \ref{eq:tan-theta-z-by-tau} and \ref{eq:z-cot-alpha} into \ref{eq:c2/c4z} we get the following limit:
\begin{align}
    \lim_{z \rightarrow \infty} \abs{\frac{c_2}{c_4z}} = \frac{1+K_2^2}{K_1}.
    \label{eq:c2/c4z-2}
\end{align}

Since $z = \nicefrac{K_1 \sqrt{\dsp}}{\gamma} $,
\begin{align}
    \lim_{z \rightarrow \infty} \abs{\frac{c_2\gamma}{c_4 K_1 \sqrt{\dsp}}} = \frac{1+K_2^2}{K_1} 
    \;\; \implies \;\; \lim_{z \rightarrow \infty} \abs{\frac{c_2\gamma}{c_4 \sqrt{\dsp}}} = 1+K_2^2
    \label{eq:c2/c4z-3}
\end{align}

Since both $\nicefrac{c_1}{c_3}$ and $\abs{\nicefrac{c_2}{c_4}}$ are continuous functions of $z$, with $\liminf_{z\rightarrow \infty}$ and $\limsup_{z\rightarrow \infty}$ converging to the limits in \ref{eq:c1/c3} and \ref{eq:c2/c4z} for both quantities respectively, we conclude that $\forall \epsilon > 0$ there exists ${\sigmasp}_0$ such that for all $\sigmasp \geq {\sigmasp}_0$, the following is true:

\begin{align}
    \frac{K_1K_2^2\din}{2L\sigmain^2 ({\din}-1)} + \epsilon \; \geq \; &\;\,\frac{c_1}{c_3} \;  \geq \; \frac{K_1K_2^2\din}{2L\sigmain^2 ({\din}-1)} - \epsilon \\
 \frac{(1+K_2^2)\sqrt{\dsp}}{\gamma} + \epsilon  \; \geq \;  &\abs{ \frac{c_2}{c_4}} \; \geq \; \frac{(1+K_2^2)\sqrt{\dsp}}{\gamma} - \epsilon, 
\end{align}

This completes both Part-I and Part-II of the proof for Theorem~\ref{thm:bt-blockform}.

\end{proof}

\subsubsection{Proof of Corollary~\ref{corollary:BT_ERM}}
\begin{corollary}[CL improves OOD error over ERM but is still imperfect] \label{corollary:BT_ERM-formal}
For $\gamma, \sigmasp, \dsp$ defined as in Theorem~\ref{thm:bt-blockform}, $\exists {\sigmasp}_{1}$ such that $\forall \sigmasp \geq {\sigmasp}_{1},$ the target accuracy of CL (linear predictor on $\Phi_\cl$) is $\geq 0.5 \erfc\paren{-{L'} \cdot \nicefrac{\gamma}{\sqrt{2}\sigmasp}}$ and $\leq 0.5 \erfc\paren{-4L' \cdot \nicefrac{\gamma}{\sqrt{2}\sigmasp}}$, where $L' = \nicefrac{K_2^2 K_1}{\sigmain^2 (1-\nicefrac{1}{\din})}$. 
When ${\sigmasp}_1$ $>$ $\sigmain \sqrt{1-\nicefrac{1}{\din}}$, the lower bound on accuracy is strictly better than ERM from scratch.
\end{corollary}

\begin{proof}
    Recall from Theorem~\ref{thm:bt-blockform-formal}, all $\phi_j$, for $j \geq 3$, lie in the null space of $\winv$ and $\wspu$. Since, the predictive features are strictly contained in the rank two space spanned by $\winv$ and $\wspu$, without loss of generality we can restrict ourselves to the case where $k=2$, and when doing training a head $h=[h_1, h_2]^\top \in \R^2$ over contrastive pretrained representations using source labeled data, we get the following max margin solution:

\begin{align}
    h_1 = c_1 \cdot \gamma + c_3 \cdot \sqrt{\dsp} \nonumber\\
    h_2 = c_2 \cdot \gamma + c_4 \cdot \sqrt{\dsp} 
\end{align}

Without loss of generality we can divide both $h_1$ and $h_2$ by $h_1$ and get the final classifier to be $\phi_1 + \frac{h_2}{h_1} \cdot \phi_2$:
\begin{align}
    &(c_1 \winv + c_3 \wspu) + \frac{h_2}{h_1} \cdot (c_2 \winv + c_4 \wspu) \nonumber \\
    &= (c_1 \winv + c_3 \wspu) + \frac{(c_2\gamma + c_4 \sqrt{\dsp})}{(c_1 \gamma + c_3  \sqrt{\dsp})} \cdot (c_2 \winv + c_4 \wspu)
\end{align}

From Lemma~\ref{lemma:error_target}, we can derive the target accuracy of the classifier $h$ on top of CL representations to be the following:

\begin{align}
 0.5  \erfc\paren{-\frac{c_1 + \beta c_2}{c_3 + \beta c_4}\cdot \frac{\gamma}{\sqrt{2}\sigmasp}}
\end{align}
where $\beta = \nicefrac{(c_2\gamma + c_4 \sqrt{\dsp})}{(c_1 \gamma + c_3  \sqrt{\dsp})} $.

Substituting $\beta$ into the expression $\frac{c_1 + \beta c_2}{c_3 + \beta c_4}$ we get:

\begin{align}
    \frac{c_1^2 \gamma  + c_1 c_3 \sqrt{\dsp} + c_2^2 \gamma + c_2 c_4 \sqrt{\dsp}}{c_1c_3\gamma + c_3^2 \sqrt{\dsp} + c_2c_4\gamma + c_4^2 \sqrt{\dsp}} 
\end{align}

We first substitute expressions for $c_1, c_2, c_3, c_4$  from \eqref{eq:c1-expr}, \eqref{eq:c2-expr}, \eqref{eq:c3-expr} and \eqref{eq:c4-expr} in the above expression.
Then for $\gamma  = K_1/\sqrt{z}, \sigmasp = K_2 \sqrt{z}$, we substitute the expressions for $\cot \alpha$, $\tan \theta$, and $\tau = \nicefrac{\lambda_1}{\lambda_2}$ with their corresponding closed form expressions (as functions of $z$) from Lemma~\ref{lem:svd-approximations}. On the resulting expression we apply 
 do repeated applications of L'Hôpital's rule to get the following result:  

\begin{align}
    \lim_{z \rightarrow \infty}  \frac{c_1^2 \gamma  + c_1 c_3 \sqrt{\dsp} + c_2^2 \gamma + c_2 c_4 \sqrt{\dsp}}{c_1c_3\gamma + c_3^2 \sqrt{\dsp} + c_2c_4\gamma + c_4^2 \sqrt{\dsp}} = \frac{2 K_2^2 K_1}{\sigmain^2 (1-\nicefrac{1}{\din})} 
    \label{eq:asy}
\end{align}

Based on $\gamma, \dsp, \sigmasp$ defined in Theorem~\ref{thm:bt-blockform}, and \eqref{eq:asy} we can conclude that $\exists {\sigmasp}_{1}$ such that for all $\sigmasp \geq {\sigmasp}_{1}$:
\begin{align}
 \frac{4 K_2^2 K_1}{\sigmain^2 (1-\nicefrac{1}{\din})}  \;  \geq \; \frac{c_1^2 \gamma  + c_1 c_3 \sqrt{\dsp} + c_2^2 \gamma + c_2 c_4 \sqrt{\dsp}}{c_1c_3\gamma + c_3^2 \sqrt{\dsp} + c_2c_4\gamma + c_4^2 \sqrt{\dsp}} \; \geq \; \frac{K_2^2 K_1}{\sigmain^2 (1-\nicefrac{1}{\din})}
    \label{eq:asy-2} 
\end{align}

Finally, applying \eqref{eq:asy-2} to Lemma~\ref{lemma:error_target}, we conclude the following: When $\gamma = \nicefrac{K_1K_2}{\sigmasp}, \dsp = \nicefrac{\sigmasp^2}{K_2^2}$, there exists ${\sigmasp}_1$, such that for any $\sigmasp \geq {\sigmasp}_1$, target accuracy of CL is at least $0.5 \erfc\paren{-{L'} \cdot \frac{\gamma}{\sqrt{2}\sigmasp}}$ and at most $0.5 \erfc\paren{-4L' \cdot \frac{\gamma}{\sqrt{2}\sigmasp}}$, where $L' = \frac{K_2^2 K_1}{\sigmain^2 (1-\nicefrac{1}{\din})}$.

\paragraph{Comparison with ERM.} Recall from Theorem~\ref{thm:ERM_scratch} the performance of ERM classifier (trained from scratch) is $0.5 \erfc\paren{-\nicefrac{\gamma^2}{\sqrt{2\dsp}\sigmasp}}$. The lower bound on the performance of classifier over CL representations is strictly better than ERM when:
\begin{align*}
& \frac{\gamma}{\sqrt{\dsp}} < L' \\ 
& \impliedby \frac{K_2^2 K_1}{\sigmain^2(1-\nicefrac{1}{\din})} > \frac{\gamma}{\sqrt{\dsp}} \impliedby \frac{K_2^2 K_1}{\sigmain^2(1-\nicefrac{1}{\din})} > \frac{K_1K_2^2}{\sigmasp^2} \\
& \impliedby \sigmasp > \sigmain \sqrt{1-\nicefrac{1}{\din}} \impliedby {\sigmasp}_1 > \sigmain \sqrt{1-\nicefrac{1}{\din}}.
    \end{align*}

This completes our proof of Corollary~\ref{corollary:BT_ERM}.

\end{proof}

\subsection{Analysis of STOC: Formal Statement of \thmref{thm:SToverCL}} 
Recall ERM solution over contrastive pretraining. We showed that without loss of generality when $k$ (the output dimensionality of $\Phi$)  is greater than 2, we can restrict $k$ to 2 and the $\Phi$ can be denoted as $[\phi_1,  \phi_2]^\top$ where
$\phi_1 = c_1 \wstar + c_3 \wsp$ and $\phi_2 = c_2 \wstar + c_4 \wsp$. 
The ERM solution of the linear head is then given by $h_1, h_2 \in \Real$:
\begin{align}
    h_1 = c_1 \cdot \gamma + c_3 \cdot \sqrt{\dsp}\,, \; \; \mathrm{ and } \; \;
    h_2 = c_2 \cdot \gamma + c_4 \cdot \sqrt{\dsp} \,.
\end{align}

STOC performs self-training of the linear head over the CL solution. 
Before introducing the result, we need some additional notation. Let $h^t$ denote the solution of the linear head at iterate $t$. Without loss of generality, assume that the coefficients in $\phi_1 = c_1 \winv + c_3 \wspu$ and $\phi_2 = c_2 \winv + c_4 \wspu$ are such that $c_2$ is positive and $c_1, c_3,$ and $c_4$ are negative. Moreover, for simplicity of exposition, assume that $\abs{c_4} > \abs{c_3}$.

\begin{theorem} \label{thm:STOC_formal}
Under the conditions of \corollaryref{corollary:BT_ERM-formal} and when $\frac{\gamma^2}{\sigmasp} \ge \left[ \frac{- c_3 - c_4}{(c_2 + c_1)\cdot \abs{c_1}}\right] \vee \left[ \frac{c_4}{c_1 \cdot c_2} \right]$, the target accuracy of ST over CL is lower bounded by $0.5\cdot \erfc\left({-\abs{\nicefrac{c2}{c4}}\cdot \nicefrac{\gamma}{(\sqrt{2} \sigma_2)}}\right) \ge 0.5\cdot \erfc\left(-L\cdot\nicefrac{\sqrt{\dsp}}{(\sqrt{2} \sigmasp)}\right)$ with $L \ge 1$.
\end{theorem}

Before proving \thmref{thm:STOC_formal}, we first connect the condition $\frac{\gamma^2}{\sigmasp} \ge \left[ \frac{- c_3 - c_4}{(c_2 + c_1)\cdot \abs{c_1}}\right] \vee \left[ \frac{c_4}{c_1 \cdot c_2} \right]$ with the result obtained with contrastive learning. 

\paragraph{Remark 1. {} {}} We first argue that $\left[ \frac{- c_3 - c_4}{(c_2 + c_1)\cdot \abs{c_1}}\right]$ term dominates and hence, if we have $\frac{\gamma^2}{\sigmasp} \ge \left[ \frac{- c_3 - c_4}{(c_2 + c_1)\cdot \abs{c_1}}\right]$, then we get the result in \thmref{thm:STOC_formal}. First, recall that as $\sigmasp$ increases, we have $\abs{\frac{c_3}{c_1}}$ converge to $\frac{2L\sigmain^2 ({\din}-1)}{K_1K_2^2\din}$, $c_2 \rightarrow 1$ and $\frac{c_1}{c_2} \rightarrow 0$. Using these limits, we get: 
\begin{equation}
   \frac{\gamma^2}{\sigmasp} = \frac{K_1^2}{K_2 \cdot z^{3/2}} \ge  \frac{2L\sigmain^2 ({\din}-1)}{K_1K_2^2\din} \,.\label{eq:condtion_STOC}
\end{equation}
which reduces the following condition: $\dsp \le K_1^2 K_2^{2/3} \cdot \left(\frac{\din}{2L\sigmain^2 ({\din}-1)}\right)^{2/3}$.

\begin{proof}
First, we create an outline of the proof. We argue about the updates of $h^t$ showing that both $h^t_1$ and $h^t_2$ increase with $\abs{h^t_2}$ becoming greater than $\abs{h^t_1}$ for some large $t$. Then we show that $\abs{h^t_2} \ge \abs{h^t_1}$ is sufficient to obtain near-perfect target generalization. 

\paragraph{Part 1.}
    Recall the loss of used for self-training of $h$: 
     \begin{align*}
        \calL_\st(h) &= \Expt{\ProbT(x)}{\ell(h^\top \Phi x, \sgn(h^\top \Phi x))} \numberthis \\ 
         &= \Expt{\ProbT(x)}{ \exp\left(- \abs{h^\top \Phi x} \right) } \numberthis \\
         &= \Expt{ z \sim \cN(0,1) }{ \exp\left(- \abs{c_1 \gamma h_1 + c_2 \gamma h_2 + (c_3 \sigmasp h_1 + c_4 \sigmasp h_2)\cdot z } \right) } \,.\numberthis  \label{eq:loss_stoc_1} \\
    \end{align*}

Define $\mu_t = c_1 \gamma h_1^t + c_2 \gamma h_2^t$ and $\sigma_t = c_3  \sigmasp h_1^t + c_4 \sigmasp h_2^t$. With this notation, we can re-write the loss in \eqref{eq:loss_stoc_1} as $\calL_\st(h^t) = \Expt{ z \sim \cN(0,\sigma_t^2) }{ \exp\left( -\abs{\mu_t + z}\right)}$. 

Similar to the the treatment in \thmref{thm:ST_scratch_formal}, we now derive a closed-form expression of $\calL_\st(h^t)$ in \lemref{lemma:g_def}: 
\begin{align}
    \calL_\st(h^t) &= \frac{1}{2} \left(  \expb{ \frac{\sigma_t^2}{2} - \mu_t} \cdot \erfcb{ -\frac{\mu_t}{\sqrt{2} \sigma_t} + \frac{\sigma_t}{\sqrt{2}} } + \expb{ \frac{\sigma_t^2}{2} + \mu_t} \cdot \erfcb{ \frac{\mu_t}{\sqrt{2} \sigma_t} + \frac{\sigma_t}{\sqrt{2}} } \right) \,.
\end{align}

Define: 
\begin{align*}
    A_1(\mu_t, \sigma_t) &= \expb{ \frac{\sigma_t^2}{2} - \mu_t} \cdot \erfcb{ -\frac{\mu_t}{\sqrt{2} \sigma_t} + \frac{\sigma_t}{\sqrt{2}} } \\
    &=  \sqrt{\frac{2}{\pi}} \expb{-\frac{\mu_t^2}{2\sigma_t^2}} \rb{\sigma_t - \frac{\mu_t}{\sigma_t}} \,, \numberthis \label{eq:A_1} \\ 
    A_2(\mu_t, \sigma_t) &= \expb{ \frac{\sigma_t^2}{2} + \mu_t} \cdot \erfcb{ \frac{\mu_t}{\sqrt{2} \sigma_t} + \frac{\sigma_t}{\sqrt{2}} } \\
    &=  \sqrt{\frac{2}{\pi}} \expb{-\frac{\mu_t^2}{2\sigma_t^2}} \rb{\sigma_t + \frac{\mu_t}{\sigma_t}} \,, \numberthis \label{eq:A_2} \\
    A_3(\mu_t, \sigma_t) &= \frac{2\sqrt{2}}{\sqrt{\pi}} \expb{-\frac{\mu_t^2}{2\sigma_t^2}} \,. \numberthis \label{eq:A_3} \\
\end{align*}

Let $\wt h^{t+1}$ denote the un-normalized gradient descent update at iterate $t+1$. We have: 
\begin{align}
    \wt h^{t+1} = h^t - \eta \cdot \frac{\partial \calL_\st(h^t)}{\partial h} \,.  
\end{align}

Now we will individually argue about the update of $\wt h^{t+1}$. First, we have: 
\begin{align*}
    \wt h^{t+1}_{1} &= h^t_1 - \eta \cdot \frac{\partial \calL_\st(h^t)}{\partial h_1} \\
    \wt h^{t+1}_{1} &= h^t_1 - \eta \cdot \underbrace{\left[ A_1 \cdot (\sigma_t c_3 \sigmasp - c_1 \gamma) + A_2 \cdot (\sigma_t c_3 \sigmasp + c_1 \gamma) - A_3 c_3 \sigmasp \right]}_{\delta_1} \numberthis \label{eq:update_ST_1} \,. \\
\end{align*}
and second, we have: 
\begin{align*}
    \wt h^{t+1}_{2} &= h^t_2 - \eta \cdot \frac{\partial \calL_\st(h^t)}{\partial h_2} \\
    \wt h^{t+1}_{2} &= h^t_2 - \eta \cdot \underbrace{\left[ A_1 \cdot (\sigma_t c_4 \sigmasp - c_2 \gamma) + A_2 \cdot (\sigma_t c_4 \sigmasp + c_2 \gamma) - A_3 c_4 \sigmasp \right]}_{\delta_2} \numberthis \label{eq:update_ST_2} \,. \\
\end{align*}

We will now argue the conditions under which $h^{t+1}_{2}$ increases till its value reaches $1/\sqrt{2}$. In particular, we will argue that when $h^{t}_{2}$ is negative, the norm $\abs{h^{t}_{2}}$ decreases and when $h^{t}_{2}$ becomes positive, then its norm increases.  We show that the following three conditions are sufficient to argue the increasing value of $h^{t}_{2}$: for all $t$, we have (i) $\mu_t \ge \mu_c$ and $\abs{\sigma_t} < \sigma_c$ for constant $\mu_c = \abs{c_1 \cdot \gamma}/2$ and $\sigma_c = \abs{c_4 \sigmasp}$; (ii) $\delta_2 < 0$; (iii) $\abs{\delta_2} \ge {\delta_1}$. 
In \lemref{lem:STOC_delta_comp_lemma}, we argue that our assumption on the initialization of the backbone learned with BT implies the previous three conditions.

\textbf{Case-1.} When $h^{t}_{2}$ is negative (and after the update, it remains negative). Then we want to argue the following: 
\begin{alignat}{3}
    &&\frac{(h^{t}_{2} - \eta \delta_2)^2}{(h^{t}_{2} - \eta \delta_2)^2 + (h^{t}_{1} - \eta \delta_1)^2} &\le (h_2^t)^2 \\
    &\Rightarrow\quad &\frac{(h^{t}_{2} - \eta \delta_2)^2}{(h_2^t)^2} &\le (h^{t}_{2} - \eta \delta_2)^2 + (h^{t}_{1} - \eta \delta_1)^2 \\
    &\Rightarrow\quad &\frac{{h^{t}_{2}}^2  + \eta^2 \delta_2^2 - 2\eta \delta_2 h^{t}_{2}}{(h_2^t)^2} &\le {h^{t}_{2}}^2 + \eta^2 \delta_2^2 -  2\eta h^{t}_{2} \delta_2  + {h^{t}_{1}}^2 + \eta^2 \delta_1^2 -  2\eta h^{t}_{1} \delta_1 \\
    &\Rightarrow\quad & 1 +\frac{\eta^2 \delta_2^2 - 2\eta \delta_2 h^{t}_{2}}{(h_2^t)^2} &\le  1 + \eta^2 \delta_2^2 -  2\eta h^{t}_{2} \delta_2  + \eta^2 \delta_1^2 -  2\eta h^{t}_{1} \delta_1  \\ 
    &\Rightarrow\quad & \eta^2 \delta_2^2 - 2\eta \delta_2 h^{t}_{2} &\le \left[ \eta^2 \delta_2^2 -  2\eta h^{t}_{2} \delta_2  + \eta^2 \delta_1^2 -  2\eta h^{t}_{1} \delta_1\right] (h_2^t)^2  \\ 
    &\Rightarrow\quad & \eta^2 \delta_2^2 (h_1^t)^2 - 2\eta \delta_2 h^{t}_{2} (h_1^t)^2 &\le  \eta^2 \delta_1^2 (h_2^t)^2 -  2\eta h^{t}_{1} \delta_1 (h_2^t)^2   \\ 
    &\Rightarrow\quad & \eta^2 \delta_2^2 (h_1^t)^2  - \eta^2 \delta_1^2 (h_2^t)^2  &\le   2\eta \delta_2 h^{t}_{2} (h_1^t)^2 - 2\eta h^{t}_{1} \delta_1 (h_2^t)^2   \\ 
    &\Rightarrow\quad & \left[\eta \delta_2 (h_1^t)  - \eta \delta_1 (h_2^t)\right]\left[ \eta \delta_2 (h_1^t)  + \eta \delta_1 (h_2^t) \right]  &\le   2  h^{t}_{2} h_1^t  \left[\eta \delta_2 (h_1^t)  - \eta \delta_1 (h_2^t)\right] \label{eq:neg_ineq1} \\
    &\Rightarrow\quad & \left[ \eta \delta_2 (h_1^t)  + \eta \delta_1 (h_2^t) \right]  &\le   2  h^{t}_{2} h_1^t \label{eq:neg_ineq2}  
\end{alignat}

Since $\delta_2 < 0$, $\abs{\delta_2} \ge \abs{\delta_1}$ and $h_2^t < h_1^t < 0$, we have $\left[\eta \delta_2 (h_1^t)  - \eta \delta_1 (h_2^t)\right]$ as positive. This implies 
inequality \eqref{eq:neg_ineq1} to \eqref{eq:neg_ineq2} and for small enough $\eta$, \eqref{eq:neg_ineq2} will continue to hold true. 

\textbf{Case-2.} When $h^{t}_{2}$ is positive but less than $1/\sqrt{2}$. Then we want to argue the following: 
\begin{alignat}{3}
    &&\frac{(h^{t}_{2} - \eta \delta_2)^2}{(h^{t}_{2} - \eta \delta_2)^2 + (h^{t}_{1} - \eta \delta_1)^2} &\ge (h_2^t)^2 \\
    &\Rightarrow\quad &\frac{(h^{t}_{2} - \eta \delta_2)^2}{(h_2^t)^2} &\ge (h^{t}_{2} - \eta \delta_2)^2 + (h^{t}_{1} - \eta \delta_1)^2 \\
    &\Rightarrow\quad &\frac{{h^{t}_{2}}^2  + \eta^2 \delta_2^2 - 2\eta \delta_2 h^{t}_{2}}{(h_2^t)^2} &\ge {h^{t}_{2}}^2 + \eta^2 \delta_2^2 -  2\eta h^{t}_{2} \delta_2  + {h^{t}_{1}}^2 + \eta^2 \delta_1^2 -  2\eta h^{t}_{1} \delta_1 \\
    &\Rightarrow\quad & 1 +\frac{\eta^2 \delta_2^2 - 2\eta \delta_2 h^{t}_{2}}{(h_2^t)^2} &\ge  1 + \eta^2 \delta_2^2 -  2\eta h^{t}_{2} \delta_2  + \eta^2 \delta_1^2 -  2\eta h^{t}_{1} \delta_1  \\ 
    &\Rightarrow\quad & \eta^2 \delta_2^2 - 2\eta \delta_2 h^{t}_{2} &\ge \left[ \eta^2 \delta_2^2 -  2\eta h^{t}_{2} \delta_2  + \eta^2 \delta_1^2 -  2\eta h^{t}_{1} \delta_1\right] (h_2^t)^2  \\ 
    &\Rightarrow\quad & \eta^2 \delta_2^2 (h_1^t)^2 - 2\eta \delta_2 h^{t}_{2} (h_1^t)^2 &\ge  \eta^2 \delta_1^2 (h_2^t)^2 -  2\eta h^{t}_{1} \delta_1 (h_2^t)^2   \\ 
    &\Rightarrow\quad & \eta^2 \delta_2^2 (h_1^t)^2  - \eta^2 \delta_1^2 (h_2^t)^2  &\ge   2\eta \delta_2 h^{t}_{2} (h_1^t)^2 - 2\eta h^{t}_{1} \delta_1 (h_2^t)^2   \\ 
    &\Rightarrow\quad & \left[\eta \delta_2 (h_1^t)  - \eta \delta_1 (h_2^t)\right]\left[ \eta \delta_2 (h_1^t)  + \eta \delta_1 (h_2^t) \right]  &\ge   2  h^{t}_{2} h_1^t  \left[\eta \delta_2 (h_1^t)  - \eta \delta_1 (h_2^t)\right] \label{eq:pos_ineq1} \\
    &\Rightarrow\quad & \left[ \eta \delta_2 (h_1^t)  + \eta \delta_1 (h_2^t) \right]  &\ge   2  h^{t}_{2} h_1^t \label{eq:pos_ineq2}  
\end{alignat}

Since $\delta_2 < 0$, $\abs{\delta_2} \ge \abs{\delta_1}$, $h_1^t \le -1/\sqrt{2}$ and  $ 0 < h_2^t < 1/\sqrt{2}$, we have $\left[\eta \delta_2 (h_1^t)  - \eta \delta_1 (h_2^t)\right]$ as positive. This implies 
inequality \eqref{eq:pos_ineq1} to \eqref{eq:pos_ineq2}. 
Focusing on \eqref{eq:pos_ineq2}, we note that $h_1^t \cdot \delta_2$ is positive and greater in magnitude than $h_2^t \cdot \delta_1$. Moreover, since $h^{t}_{2} h_1^t$ is negative, \eqref{eq:pos_ineq2} will continue to hold true.

Now, when $h^{t}_{2}$ is positive and greater than $1/\sqrt{2}$, then $h^{t}_{2}$  will stay in that region. Convergence of STOC together with conditions of convergence as in  \lemref{lemma:convergence_STOC} will imply that the at convergence $h^{t}_{2}$ will remain greater than $1/\sqrt{2}$, such that $\frac{h_1^{t_c}}{h_2^{t_c}} = \frac{\delta_1}{\delta_2}$. Now we bound the target error of STOC.

\textbf{Part 2.} To bound the accuracy at any iterate $t$ when $h^{t}_{2} \ge 1/\sqrt{2}$, we have from \lemref{lemma:error_target}: 

\begin{align}
        \Expt{\ProbT}{y\cdot \left( {h^t}^\top \phi_\cl x \right) > 0} 
        &= \Expt{z \sim \cN(0,1)}{ z > - \frac{c_1 \gamma h^t_1 + c_2 \gamma h^t_2}{\abs{c_3 \sigmasp h^t_1 + c_4 \sigmasp h^t_2}}  } \label{eq:STOC_target_err}\,. 
    \end{align}

We now upper bound and lower bound the fraction $\frac{c_1 \gamma h^t_1 + c_2 \gamma h^t_2}{\abs{c_3 \sigmasp h^t_1 + c_4 \sigmasp h^t_2}} $ in RHS in \eqref{eq:STOC_target_err}: (i) $c_1 \gamma h^t_1 + c_2 \gamma h^t_2 \ge c_2 \gamma h^t_2$ since both $c_1 \gamma h^t_1$ and $c_2 \gamma h^t_2$ have same sign; (ii) $\abs{c_3 \sigmasp h^t_1 + c_4 \sigmasp h^t_2} \le \abs{c_4 \sigmasp h^t_2}$ because $\abs{c_4 \sigmasp h^t_2} \ge \abs{c_3 \sigmasp h^t_1}$ and they have opposite signs. Hence, from \eqref{eq:STOC_target_err}, we have: 

\begin{align}
        \Expt{\ProbT}{y\cdot \left( {h^t}^\top \phi_\cl x \right) > 0} 
        &= \Expt{z \sim \cN(0,1)}{ z > - \frac{ c_2 \gamma h^t_2}{\abs{c_4 \sigmasp h^t_2}} } = \Expt{z \sim \cN(0,1)}{ z > - \frac{ c_2 \gamma}{\abs{c_4 \sigmasp}}}\label{eq:STOC_target_err_final}\,. 
\end{align}
Substituting the definition of $\erfc$, 
the expression \eqref{eq:STOC_target_err_final} gives us the  required lower bound on the target accuracy. 

\end{proof}

\begin{lemma}[Convergence of STOC] \label{lemma:convergence_STOC}
    Assume the gradient updates as in \eqref{eq:update_ST_1} and \eqref{eq:update_ST_2}. Then STOC converges at $t = t_c$ when $\frac{h_1^{t_c}}{h_2^{t_c}} = \frac{\delta_1}{\delta_2}$. For $t>t_c$, \eqref{eq:update_ST_1} and \eqref{eq:update_ST_2} make no updates to the linear $h$. 
\end{lemma}
\begin{proof}
When the gradient updates $\delta_1$ and $\delta_2$ are such that $h_{1}^{t+1}$ matches $h_{1}^{t}$, we have convergence of STOC. 

\begin{alignat}{3}
    &&\frac{(h^{t}_{2} - \eta \delta_2)^2}{(h^{t}_{2} - \eta \delta_2)^2 + (h^{t}_{1} - \eta \delta_1)^2} &=(h_2^t)^2 \\
    &\Rightarrow\quad &\frac{(h^{t}_{2} - \eta \delta_2)^2}{(h_2^t)^2} &= (h^{t}_{2} - \eta \delta_2)^2 + (h^{t}_{1} - \eta \delta_1)^2 \\
    &\Rightarrow\quad &\frac{{h^{t}_{2}}^2  + \eta^2 \delta_2^2 - 2\eta \delta_2 h^{t}_{2}}{(h_2^t)^2} &= {h^{t}_{2}}^2 + \eta^2 \delta_2^2 -  2\eta h^{t}_{2} \delta_2  + {h^{t}_{1}}^2 + \eta^2 \delta_1^2 -  2\eta h^{t}_{1} \delta_1 \\
    &\Rightarrow\quad & 1 +\frac{\eta^2 \delta_2^2 - 2\eta \delta_2 h^{t}_{2}}{(h_2^t)^2} &=  1 + \eta^2 \delta_2^2 -  2\eta h^{t}_{2} \delta_2  + \eta^2 \delta_1^2 -  2\eta h^{t}_{1} \delta_1  \\ 
    &\Rightarrow\quad & \eta^2 \delta_2^2 - 2\eta \delta_2 h^{t}_{2} &= \left[ \eta^2 \delta_2^2 -  2\eta h^{t}_{2} \delta_2  + \eta^2 \delta_1^2 -  2\eta h^{t}_{1} \delta_1\right] (h_2^t)^2  \\ 
    &\Rightarrow\quad & \eta^2 \delta_2^2 (h_1^t)^2 - 2\eta \delta_2 h^{t}_{2} (h_1^t)^2 &=  \eta^2 \delta_1^2 (h_2^t)^2 -  2\eta h^{t}_{1} \delta_1 (h_2^t)^2   \\ 
    &\Rightarrow\quad & \eta^2 \delta_2^2 (h_1^t)^2  - \eta^2 \delta_1^2 (h_2^t)^2  &=   2\eta \delta_2 h^{t}_{2} (h_1^t)^2 - 2\eta h^{t}_{1} \delta_1 (h_2^t)^2   \\ 
    &\Rightarrow\quad & \left[\eta \delta_2 (h_1^t)  - \eta \delta_1 (h_2^t)\right]\left[ \eta \delta_2 (h_1^t)  + \eta \delta_1 (h_2^t) \right]  &=   2  h^{t}_{2} h_1^t  \left[\eta \delta_2 (h_1^t)  - \eta \delta_1 (h_2^t)\right] \label{eq:eq_ineq}
\end{alignat}

Thus either $\left[\eta \delta_2 (h_1^t)  - \eta \delta_1 (h_2^t)\right] = 0$ or  $\left[ \eta \delta_2 (h_1^t)  + \eta \delta_1 (h_2^t) \right]  =   2  h^{t}_{2} h_1^t$. Since $\eta$ is such that $h_1 - \eta \delta_1 < 0$, $\left[ \eta \delta_2 (h_1^t)  + \eta \delta_1 (h_2^t) \right]  \ne  2  h^{t}_{2} h_1^t$ implying that $\left[\eta \delta_2 (h_1^t)  - \eta \delta_1 (h_2^t)\right] = 0$ giving us the required condition. 
\end{proof}

\begin{lemma}\label{lem:STOC_delta_comp_lemma}
Under the initialization conditions assumed in \thmref{thm:STOC_formal}, for all $t$, we have: (i) $\mu_t \ge \mu_c$ and $\abs{\sigma_t} \le \sigma_c$ for constant $\mu_c = \abs{c_1 \cdot \gamma}/2$ and $\sigma_c = \abs{c_4 \sigmasp}$; (ii) $\delta_2 < 0$; (iii) $\abs{\delta_2} \ge {\delta_1}$, where $\delta_1 = A_1 \cdot (\sigma_t c_3 \sigmasp - c_1 \gamma) + A_2 \cdot (\sigma_t c_3 \sigmasp + c_1 \gamma) - A_3 c_3 \sigmasp$ and $\delta_2 = A_1 \cdot (\sigma_t c_4 \sigmasp - c_2 \gamma) + A_2 \cdot (\sigma_t c_4 \sigmasp + c_2 \gamma) - A_3 c_4 \sigmasp$ for $A_1, A_2$ and $A_3$ defined in \eqref{eq:A_1},  \eqref{eq:A_2}, and  \eqref{eq:A_3}. 
\end{lemma}
\begin{proof}
    Recall, $\mu_t = c_1 \gamma h_1^t + c_2 \gamma h_2^t$ and $\sigma_t = c_3  \sigmasp h_1^t + c_4 \sigmasp h_2^t$. 
    First, we argue that $\mu_t$ increases from the initialization value. Notice that $\mu_0 = c_1 \gamma h_1^0 + c_2 \gamma h_2^0$. Due to \corollaryref{corollary:BT_ERM-formal}, we have $h_2^0 \ge 0$. And since $\abs{c_2} > \abs{c_1}$, we get $\mu_0 \ge {\abs{c_1 \gamma}}$ as both $c_1$ and $h_1^0$ are of same sign.
    Moreover, as training progresses with $h_1^t$ remaining negative and $h_2^t$ remaining positive, we have $\mu_t$ stays greater than $\mu_0$.

        Recall the definition of $A_1, A_2$, and $A_3$ in \eqref{eq:A_1},  \eqref{eq:A_2}, and  \eqref{eq:A_3}. Moreover, recall the definition of $\alpha_1(\mu_t, \sigma_t)$ and $\alpha_2(\mu_t, \sigma_t)$:  
        \begin{align}
            \alpha_1(\mu_t, \sigma_t) = \sqrt{\frac{2}{\pi}} \expb{-\frac{\mu_t^2}{2\sigma_t^2}} \left[\rb{\sigma_t + \frac{\mu_t}{\sigma_t}}  - \rb{ \sigma_t - \frac{\mu_t}{\sigma_t}} \right] \,.
        \end{align}
        and 
        \begin{align}
            \alpha_2(\mu_t, \sigma_t) = \sqrt{\frac{2}{\pi}} \expb{-\frac{\mu_t^2}{2\sigma_t^2}} \left[\rb{\sigma_t + \frac{\mu_t}{\sigma_t}}  + \rb{ \sigma_t - \frac{\mu_t}{\sigma_t}} - \frac{2}{\sigma_t} \right] \,.
        \end{align}

        Thus, we have $\alpha_1(\mu_t, \sigma_t) \cdot A_3 = A_1 \cdot \sigma_t $ and $\alpha_2(\mu_t, \sigma_t) \cdot A_3 =  \sigma_t \cdot \left( A_2 \cdot - \frac{2}{\sigma_t} A_3 \right) $. Replacing the definition of $A_1$, $A_2$, and $A_3$  in $\delta_1$ and $\delta_2$, we get: 

        \begin{align*}
            \delta_1 = \sigma_t c_3 \sigmasp \cdot \alpha_2(\mu_t, \sigma_t) + c_1 \gamma \alpha_1(\mu_t, \sigma_t) \quad \text{and} \quad \delta_2 = \sigma_t c_4 \sigmasp \cdot \alpha_2(\mu_t, \sigma_t) + c_2 \gamma \alpha_1(\mu_t, \sigma_t) \numberthis \label{eq:delta_exp}
        \end{align*}

        We now upper bound and lower bound $\alpha_1$ and $\alpha_2$ by using the properties of $\rb{\cdot}$. 
        We use Taylor's expansion on $\rb{\cdot}$ and we get: 
        \begin{align}
          \rb{\sigma_t} +  \rbb{\prime}{\sigma_t} \cdot \left(\frac{\mu_t}{\sigma_t} \right) \le \rb{\sigma_t + \frac{\mu_t}{\sigma_t}} \le \rb{\sigma_t} +  \rbb{\prime}{\sigma_t} \cdot \left(\frac{\mu_t}{\sigma_t} \right) + \rbb{{\prime\prime}}{\sigma_t} \cdot \left(\frac{\mu_t}{\sigma_t} \right)^2
        \end{align}
        and similarly, we get: 
        \begin{align}
          \rb{\sigma_t} -  \rbb{\prime}{\sigma_t} \cdot \left(\frac{\mu_t}{\sigma_t} \right) + \rbb{{\prime\prime}}{\sigma_t} \cdot \left(\frac{\mu_t}{\sigma_t} \right)^2 \le \rb{\sigma_t - \frac{\mu_t}{\sigma_t}} \le \rb{\sigma_t} - \rbb{\prime}{\sigma_t} \cdot \left(\frac{\mu_t}{\sigma_t} \right) + R^{\prime\prime} \left(\frac{\mu_t}{\sigma_t} \right)^2 \label{eq:taylor_expr_repeat_stoc}
        \end{align}
        
        where $R^{\prime\prime} = \rbb{{\prime\prime}}{\sigma_0}$. This is because  $\rbb{{\prime\prime}}{\cdot}$ takes positive values and is a decreasing function in $\sigma_t$ (refer to \lemref{lemma:mill_ratio}).  
        We now lower bound $\alpha_1(\mu_t, \sigma_t)$ and upper bound $\alpha_2(\mu_t, \sigma_t)$: 
        \begin{align}
            \frac{\alpha_1(\mu_t, \sigma_t)}{ \sqrt{\frac{2}{\pi}} \expb{-\frac{\mu_t^2}{2\sigma_t^2}} } \le  2 \rbb{\prime}{\sigma_t} \cdot \left(\frac{\mu_t}{\sigma_t} \right)
        \end{align}

        \begin{align}
            \frac{\alpha_2(\mu_t, \sigma_t)}{ \sqrt{\frac{2}{\pi}} \expb{-\frac{\mu_t^2}{2\sigma_t^2}} } \ge  2 \rb{\sigma_t} +  \rbb{{\prime\prime}}{\sigma_t} \cdot \left(\frac{\mu_t}{\sigma_t} \right)^2 - \frac{2}{\sigma_t}
        \end{align}

        \textbf{Part-1. {} {}} We first prove that $\delta_2 \le 0$. 
         Substituting the lower bound and upper bound in \eqref{eq:delta_exp} gives us the following as stricter a sufficient condition (i.e., \eqref{eq:stricter_sufficiency_delta_2} implies $\delta_2 \le 0$): 
        \begin{align}
            & \left[ 2 \rb{\sigma_t} +  \rbb{{\prime\prime}}{\sigma_t} \cdot \left(\frac{\mu_t}{\sigma_t} \right)^2 - \frac{2}{\sigma_t} \right] \cdot \frac{\sigmasp \cdot (-c_4)}{\gamma\cdot c_2} \ge 2 \rbb{\prime}{\sigma_t} \cdot \left(\frac{\mu_t}{\sigma_t} \right)
            \label{eq:stricter_sufficiency_delta_2} \\
            \iff & \left[ 2 \rb{\sigma_t} +  \rbb{{\prime\prime}}{\sigma_t} \cdot \left(\frac{\mu_t}{\sigma_t} \right)^2 - \frac{2}{\sigma_t} \right] \ge 2 \rbb{\prime}{\sigma_t} \cdot \left(\frac{\mu_t}{\sigma_t} \right) \cdot \frac{\gamma\cdot c_2}{\sigmasp \cdot (-c_4)} \\
            \iff & 2 \rb{\sigma_t} +  \rbb{{\prime\prime}}{\sigma_t} \cdot \left(\frac{\mu_t}{\sigma_t} \right)^2 - \frac{2}{\sigma_t} -  2 \rbb{\prime}{\sigma_t} \cdot \left(\frac{\mu_t}{\sigma_t} \right) \cdot  \frac{\gamma\cdot c_2}{\sigmasp \cdot (-c_4)}  \ge 0 \\
            \iff & 2 \rb{\sigma_t} \cdot \sigma_t +  \rbb{{\prime\prime}}{\sigma_t} \cdot \frac{\mu_t^2}{\sigma_t}  - 2 -  2 \rbb{\prime}{\sigma_t} \cdot \mu_t \cdot \frac{\gamma\cdot c_2}{\sigmasp \cdot (-c_4)}  \ge 0 \\
            \iff & 2 \rbb{\prime}{\sigma_t} +  \rbb{{\prime\prime}}{\sigma_t} \cdot \frac{\mu_t^2}{\sigma_t}  -  2 \rbb{\prime}{\sigma_t} \cdot \mu_t \cdot \frac{\gamma\cdot c_2}{\sigmasp \cdot (-c_4)}  \ge 0 \\
            \iff &  \rbb{{\prime\prime}}{\sigma_t} \cdot \frac{\mu_t^2}{\sigma_t}  + 2 \rbb{\prime}{\sigma_t}  \cdot \left[ 1 -  \mu_t \cdot \frac{\gamma\cdot c_2}{\sigmasp \cdot (-c_4)} \right] \ge 0
        \end{align}

        Thus, if we have $\mu_t \ge \frac{\sigmasp \cdot (-c_4)}{\gamma\cdot c_2}$, then \eqref{eq:stricter_sufficiency_delta_2} holds true.

        \textbf{Part-2. {} {}} Next, we prove that $\abs{\delta_2} \ge \delta_1$. 
        Substituting the lower bound and upper bound in \eqref{eq:delta_exp} gives us the following as stricter a sufficient condition (i.e., \eqref{eq:stricter_sufficiency_delta_comp} implies $\abs{\delta_2} \ge \delta_1$): 
       \begin{align}
            & \left[ 2 \rb{\sigma_t} +  \rbb{{\prime\prime}}{\sigma_t} \cdot \left(\frac{\mu_t}{\sigma_t} \right)^2 - \frac{2}{\sigma_t} \right] \cdot \frac{\sigmasp \cdot (-c_4 -c_3)}{\gamma\cdot (c_2 + c_1)} \ge 2 \rbb{\prime}{\sigma_t} \cdot \left(\frac{\mu_t}{\sigma_t} \right)
            \label{eq:stricter_sufficiency_delta_comp} \\
            \iff & \left[ 2 \rb{\sigma_t} +  \rbb{{\prime\prime}}{\sigma_t} \cdot \left(\frac{\mu_t}{\sigma_t} \right)^2 - \frac{2}{\sigma_t} \right] \ge 2 \rbb{\prime}{\sigma_t} \cdot \left(\frac{\mu_t}{\sigma_t} \right) \cdot \frac{\gamma\cdot (c_2 + c_1)}{\sigmasp \cdot (-c_4 -c_3)}  \\
            \iff & 2 \rb{\sigma_t} +  \rbb{{\prime\prime}}{\sigma_t} \cdot \left(\frac{\mu_t}{\sigma_t} \right)^2 - \frac{2}{\sigma_t} -  2 \rbb{\prime}{\sigma_t} \cdot \left(\frac{\mu_t}{\sigma_t} \right) \cdot   \frac{\gamma\cdot (c_2 + c_1)}{\sigmasp \cdot (-c_4 -c_3)}  \ge 0 \\
            \iff & 2 \rb{\sigma_t} \cdot \sigma_t +  \rbb{{\prime\prime}}{\sigma_t} \cdot \frac{\mu_t^2}{\sigma_t}  - 2 -  2 \rbb{\prime}{\sigma_t} \cdot \mu_t \cdot \frac{\gamma\cdot (c_2 + c_1)}{\sigmasp \cdot (-c_4 -c_3)}  \ge 0 \\
            \iff & 2 \rbb{\prime}{\sigma_t} +  \rbb{{\prime\prime}}{\sigma_t} \cdot \frac{\mu_t^2}{\sigma_t}  -  2 \rbb{\prime}{\sigma_t} \cdot \mu_t \cdot \frac{\gamma\cdot (c_2 + c_1)}{\sigmasp \cdot (-c_4 -c_3)}  \ge 0 \\
            \iff &  \rbb{{\prime\prime}}{\sigma_t} \cdot \frac{\mu_t^2}{\sigma_t}  + 2 \rbb{\prime}{\sigma_t}  \cdot \left[ 1 -  \mu_t \cdot \frac{\gamma\cdot (c_2 + c_1)}{\sigmasp \cdot (-c_4 -c_3)} \right] \ge 0
        \end{align}
        
    Thus, if we have $\mu_t \ge \frac{\sigmasp \cdot (-c_4 -c_3)} {\gamma\cdot (c_2 + c_1)}$, then \eqref{eq:stricter_sufficiency_delta_comp} holds true which in-turn implies $\abs{\delta_2} \ge \delta_1$. 
    Plugging in $\mu_t \ge \mu_0$, we get the required condition.       
    
\end{proof}

\subsection{Analysis for SSL} \label{app:SSL_analysis}
\label{subsec:ssl-analysis}
For SSL analysis, we argue that the projection learned by contrastive pretraining can significantly improve the generalization of the linear head learned on top, leaving little to no room for improvement for self-training. Our analysis leverages the margin-based bound for linear models from \citet{kakade2008complexity}. Before introducing the result, we present some additional notation. Let $\err_D(w)$ denote 0-1 error of a classifier on a distribution $D$. Define 0-1 error with margin $\xi$ as $\wh \err_\xi(w) = \sum_{i=1}^n \frac{\indict{y_i w^\top x_i \le \xi}}{n}$.

\begin{theorem}[generalization bound for margin loss] 
For all classifiers $w$ and margin $\gamma$, we have with probability at least $1-\delta$: 
\begin{align}
    \err_T(w) \le \wh \err_\xi(w) + 4 \frac{B}{\xi}\sqrt{\frac{1}{n}} + \sqrt{\frac{\log(2/\delta)}{n}} + \sqrt{\frac{\log(\log_2(4B/\xi))}{n}} \,, \label{eq:margin_bound}
\end{align}
where $B = 4 \max(\max(\sigmain, \sigmasp), 1) \cdot \paren{\sqrt{\din + \dsp} + \sqrt{\log\paren{\nicefrac{2n}{\delta}}}} + \gamma$ is a high probability upper bound on the $\ell_2$ norm of the input points $x$. 
\end{theorem}
\begin{proof}
    The result is a trivial application of union bound over: (1) Corollary 6 in \citet{kakade2008complexity}; and (2) high probability bound over norms of sub-gaussian random variables (Sec. 5.2 in ~\cite{wainwright2019high}). 
\end{proof}

When $\wh \err_\xi(w)$ is close to zero, 
the denominating term in RHS of \eqref{eq:margin_bound} is $ \nicefrac{4B}{\xi}\sqrt{\nicefrac{1}{n}}$. From \propref{prp:bt-closedform},  CL solution $\phi_\cl$ obtained on the target domain alone (for SSL setup) is $\winv$ when $k=1$. Intuitively, since the target data has only one predictive feature (along $\winv$), CL directly recovers this predictive feature as it is the predominant direction that minimizes invariance loss.
Consequently, projecting the inputs on the CL solution mainly reduces the value of $B$ on the projected data. This happens because the effective dimension is reduced from $\sqrt{d}=\sqrt{\din + \dsp}$ to $\sqrt{k}$  (which is $=1$ for $k=1$), which is the output dimension of the feature extractor $\phi_\cl$. 
Additionally, since $\winv$ is recovered by $\phi_\cl$, the maximum margin between the two classes remains $\gamma$, thus for any $\xi \leq \gamma $, $\exists w$ such that $\wh \err_\xi(w) = 0$. 

Assuming we can recover the linear predictor that minimizes the empirical loss, the only dominating term left in the upper bound in \eqref{eq:margin_bound} is $\nicefrac{4B}{\xi}\sqrt{\nicefrac{1}{n}}$. 
When we reduce this term, we get a tighter upper bound for linear probing. As a result, in the SSL setup, linear probing performed on top of CL features results in a predictor with a much smaller value of the upper bound,  when compared with linear probing done on inputs directly.
Even for larger $k$, as long as $k=o(d)$ the generalization error bound for the CL predictor under the SSL setup reduces drastically compared to ERM. 
This explains why doing further self-training over the CL predictor in the SSL setup does not result in big gains on the target accuracy as compared to the UDA setting.

\section{Limitations of Prior Work}
\label{app:limitations-prior-work}

\subsection{Contrastive learning analysis} \label{app:CL_prior_work_assumption}

Prior works that analyze contrastive learning show that minimizers of the CL objective recover clusters in the augmentation graph, which weights pairs of augmentations with their probability of being sampled as a positive pair~\citep{haochen2021provable,cabannes2023ssl,saunshi2022understanding,johnson2022contrastive}. 
When there is no distribution shift in the downstream task, assumptions made on the graph in the form of consistency of augmentations with downstream labels, is sufficient to ensure that a linear probed head has good ID 
generalization. Under distribution shift, these assumptions are not sufficient and stronger ones are needed. \textit{E.g.}, 
some works assume that same-domain/class examples are weighted higher that cross-class cross-domain pairs~\cite{haochen2022beyond,shen2022connect}. 

Using notation defined in ~\cite{shen2022connect}, the assumption on the augmentation graph requires  cross-class and same-domain  weights ($\beta$) to 
 be higher than  cross-class and cross-domain weights ($\gamma$).  
It is unclear if examples from different classes in the same domain will be ``connected'' if strong spurious features exist in the source domain and augmentations fail to mask them completely (\eg, image background may not be completely masked by augmentations but it maybe perfectly predictive of the label on source domain). In such cases, the linear predictor learnt over CL would fail to generalize OOD. 
In our toy setup as well, the connectivity assumption fails since on source  $\xsp$ is perfectly predictive of the label and the augmentations are imperfect, \ie, augmentations do not mask $\xsp$ and examples of different classes do not overlap in source (\ie, $\beta=0$). On the other hand, since $\xsp$ is now random on target, augmentations of different classes may overlap, \ie, $\gamma > 0$, thus breaking the connectivity assumption. This is also highlighted in our empirical findings of CL furnishing representations that do not fully enable linear transferability from source to target (see~\secref{sec:discussion}). These empirical findings also call into question 
existing assumptions on data augmentations, highlighting that
perfect linear transferability may not typically hold in practice.  
It is in this setting that we believe self-training can improve over contrastive learning by unlearning source-only features and improving linear transferability.

\subsection{Self-training analysis} \label{app:iterative-propagation}

Some prior works on self-training view it as consistency regularization   
that constrain pseudolabels of original samples to be consistent with all their augmentations~\citep{cai2021theory, wei2020theoretical, sohn2020fixmatch}. 
This framework abstracts the role played by the optimization algorithm and instead evaluates the global minimizer of a population objective that enforces consistency of pseudolabels. In addition, certain expansion assumptions on class-conditional distributions are needed to ensure that pseudolabels have good accuracy on source and target domains.  
This framework does not account for challenges involved in propagating labels iteratively. 
For \eg, when augmentation distribution has long tails, the consistency of pseudolabels depends on the sampling frequency of ``favorable'' augmentations. 
As an illustration, consider our augmentation distribution in the toy setup in \secref{sec:theory}. If it were not uniform over dimensions, but instead something that was highly skewed, then a large number of augmentations need to be sampled for every data point to propagate pseudolabels successfully from source labeled samples to target unlabeled samples during self-training. This might hurt the performance of ST when we are optimizing for only finitely many iterations and over finitely many datapoints. 
This is why in our analysis we instead adopt the iterative analysis of self-training~\cite{chen2020self}.

\section{Additional Lemmas} \label{app:extra_lemma_st}
In this section we define some additional lemmas that we use in our theoretical analysis in \ref{appsec:proofs}.

\begin{lemma}[Upper bound and lower bounds on $\erfc$; \citet{kschischang2017complementary}] \label{lemma:erfc_bounds}
Define $\erfc(x) = \frac{2}{\sqrt{\pi}} \cdot \int_{x}^{\infty} \exp(-z^2) \cdot dz$. Then we have:  
\begin{align*}
  \frac{2}{\sqrt{\pi}} \cdot \frac{\exp(-x^2)}{x + \sqrt{x^2+2}} < \erfc(x) \le \frac{2}{\sqrt{\pi}} \cdot  \frac{\exp(-x^2)}{x + \sqrt{x^2+4/\pi}}
\end{align*}
\end{lemma}

\begin{lemma}[Properties of Mill's ratio~\citep{baricz2008mills}] \label{lemma:mill_ratio}
Define the Mill's ratio as $\rb{x} = \expb{x^2/2} \cdot \erfcb{x / \sqrt{2}} \cdot \sqrt{\pi/2}$. Then following assertions are true: 
(i) $\rb{x}$ is a strictly decreasing log-convex function; 
(ii) $\mathrm{r}^{\prime}(x) = x\cdot \rb{x} - 1$ is an increasing function with $\mathrm{r}^{\prime}(x) < 0$ for all x; 
(iii) $\mathrm{r}^{\prime\prime}(x) = \rb{x} + x^2  \cdot \rb{x}  - x$ is a decreasing function with $\mathrm{r}^{\prime\prime}(x) >0$ for all $x$; 
(iv) $x^2 \cdot \mathrm{r}^{\prime}(x)$ is a decreasing function of $x$. 
\end{lemma}

\begin{lemma}[invariance loss as product with operator $L$]
\label{lem:inv-loss-operator}
    The invariance loss for some $\phi \in \R^d$ is given as: $2\cdot\int_\calA \phi(a) \cdot L(\phi)(a) \; \mathrm{d}\ProbA$ where the operator $L$ is defined as:
    \begin{align*}
        L(\phi)(a) = \phi(a) - \int_{\calA} \frac{A_+(a, a')}{p_\mathsf{A}(a)} \cdot \phi(a') \; \mathrm{d}a'
    \end{align*}
\end{lemma}
\begin{proof}
 The invariance loss for $\phi$ is given by:
 \begin{align}
     &\Exp_{x\sim\ProbU}\Exp_{a_1, a_2 \sim \ProbA(\cdot \mid x)} (a_1^\top \phi - a_2^\top \phi)^2 = 2\Exp_{x\sim\ProbU}\Exp_{a\sim \ProbA(\cdot \mid x)} \brck{\phi(a)^2} \nonumber \\
     &\fourquad\fourquad\fourquad\fourquad - 2\Exp_{a_1, a_2 \sim A_+(\cdot, \cdot)} \brck{\phi(a_1)\phi(a_2)} \\
     &= 2\cdot\int_\calA \phi(a)^2 \; \mathrm{d}\ProbA - 2\cdot \int_\calA \phi(a) \paren{\int_\calA \frac{A_+(a, a_2)}{p_\mathsf{A}(a)} \cdot \phi(a_2) \; \mathrm{d}a_2 } \; \mathrm{d}\ProbA  \\
    &= 2\cdot\int_\calA \phi(a) \cdot L(\phi)(a) \; \mathrm{d}\ProbA
 \end{align}
\end{proof}

\begin{lemma}
    \label{lem:bt-covariance-zero}
    If $\calW$ is the space spanned by $\winv$ and $\wspu$, and $\calW_\perp$ is the null space for $\calW$, then for any $u \in \calW$ and any $v \in \calW_\perp$, the covariance along these directions $\Exp_{a \sim \ProbA} [a^\top u v^\top a] = 0$. 
\end{lemma}
\textit{Proof:}
We can write the covariance over augmentations after we break down the augmentation $a$ into two projections: $a = \Pi_\calW (a) + \Pi_{\calW_\perp} (a)$
\begin{align}
  \Exp_{a \sim \ProbA} [a^\top u v^\top a] &= \Exp_{a \sim \ProbA} \brck{\paren{u^\top(\Pi_\calW (a) + \Pi_{\calW_\perp} (a))} \paren{v^\top(\Pi_\calW (a) + \Pi_{\calW_\perp} (a))}} \\
  &= \Exp_{a \sim \ProbA} \brck{\paren{u^\top\Pi_\calW (a)} \paren{v^\top \Pi_{\calW_\perp} (a)}} \\
  &= u^\top\paren{ \Exp_{a \sim \ProbA} \brck{\Pi_\calW (a)\Pi_{\calW_\perp} (a)^\top}} v = 0
\end{align}
where the last inequality follows from the fact that $\Exp_{a \sim \ProbA} \brck{\Pi_\calW (a)\Pi_{\calW_\perp} (a)^\top} = \Exp_{a \sim \ProbA} \brck{\Pi_\calW (a)} \Exp_{a \sim \ProbA} \brck{\Pi_{\calW_\perp} (a)}^\top$, since the noise in the null space of $\calW$ is drawn independent of the component along $\calW$, and furthermore the individual expectations evaluate to zero. 

\begin{lemma}[closed-form expressions for eigenvalues and eigenvectors of $\Sigma_A, \tilde{\Sigma}$]
    \label{lem:svd-approximations}
    For a $2 \times 2$ real symmetric matrix $\begin{bmatrix}a, \;\; b \\c, \;\; d\end{bmatrix}$ the eigenvalues $\lambda_1, \lambda_2$ are given by the following expressions:
    \[
        \lambda_1 = \frac{(a + b + \delta)}{2}, \;\;
        \lambda_2 = \frac{(a + b - \delta)}{2},  
    \]
    where $\delta= \sqrt{4c^2 + (a-b)^2}$.
    Further, the eigenvectors are given by $U = \begin{bmatrix}
        \cos(\theta), \hfill \sin(\theta) \\
        \sin(\theta), \hfill -cos(\theta)
    \end{bmatrix}$, where:
    \[    
    \tan(\theta) = \frac{b - a + \delta}{2c}.
    \]
    For full proof of these statements see~\cite{deledalle2017closed}. Here, we will use these statements to arrive at closed form expressions for the eigenvalues and eigenvectors of $\Sigma_A$, $\tilde{\Sigma}$.
\end{lemma}
\begin{proof}

We can now substitute the above formulae with $a,b,c,d$ taken from the expressions of $\Sigma_A$ and $\tilde{\Sigma}$, to get the following values:  $\lambda_1,\lambda_2$ are the eigenvalues of $\Sigma_A$, with $\alpha$ determining the corresponding eigenvectors $[\cos(\alpha), \sin(\alpha)], [\sin(\alpha), -\cos(\alpha)]$; and  $\tilde{\lambda}_1,\tilde{\lambda}_2$ are the eigenvalues of $\tilde{\Sigma}$, with $\beta$ determining the corresponding eigenvectors: $[\cos(\beta), \sin(\beta)], [\sin(\beta), -\cos(\beta)]$.

\begin{align}
    \lambda_1 &= \frac{1}{8}\Bigg( \gamma^2\paren{1+\frac{1}{3\din}} + \frac{\sigmain^2}{3} \paren{1-\frac{1}{\din}} + \frac{\dsp}{2} + \frac{2\sigmasp^2}{3} + \frac{1}{6} \nonumber \\ 
    \quad & +\sqrt{\gamma^2\dsp + \paren{\paren{\gamma^2\paren{1+\frac{1}{3\din}} + \frac{\sigmain^2}{3} \paren{1-\frac{1}{\din}}} - \paren{\frac{\dsp}{2} + \frac{2\sigmasp^2}{3} + \frac{1}{6}} }^2}  \Bigg) 
\end{align}
\begin{align}
    \lambda_2 &= \frac{1}{8}\Bigg( \gamma^2\paren{1+\frac{1}{3\din}} + \frac{\sigmain^2}{3} \paren{1-\frac{1}{\din}} + \frac{\dsp}{2} + \frac{2\sigmasp^2}{3} + \frac{1}{6} \nonumber  \\ 
    \quad & -\sqrt{\gamma^2\dsp + \paren{\paren{\gamma^2\paren{1+\frac{1}{3\din}} + \frac{\sigmain^2}{3} \paren{1-\frac{1}{\din}}} - \paren{\frac{\dsp}{2} + \frac{2\sigmasp^2}{3} + \frac{1}{6}} }^2}  \Bigg) \\
    \tilde{\lambda}_1 &= \frac{1}{8}\Bigg( \gamma^2 + \frac{\dsp}{2} + \frac{\sigmasp^2}{2}  
    +\sqrt{\gamma^2\dsp + \paren{\gamma^2 - \paren{\frac{\dsp}{2} + \frac{\sigmasp^2}{2}}}^2} \Bigg) \\
    \tilde{\lambda}_2 &= \frac{1}{8}\Bigg( \gamma^2 + \frac{\dsp}{2} + \frac{\sigmasp^2}{2}  
     -\sqrt{\gamma^2\dsp + \paren{\gamma^2 - \paren{\frac{\dsp}{2} + \frac{\sigmasp^2}{2}}}^2} \Bigg)  \\
    \tan(\alpha) &=  \frac{1}{\gamma\sqrt{\dsp}}\Bigg( \frac{\dsp}{2} + \frac{2\sigmasp^2}{3} + \frac{1}{6} -\paren{\gamma^2\paren{1+\frac{1}{3\din}} + \frac{\sigmain^2}{3} \paren{1-\frac{1}{\din}}} \nonumber \\ 
    \quad & +\sqrt{\gamma^2\dsp + \paren{\paren{\gamma^2\paren{1+\frac{1}{3\din}} + \frac{\sigmain^2}{3} \paren{1-\frac{1}{\din}}} - \paren{\frac{\dsp}{2} + \frac{2\sigmasp^2}{3} + \frac{1}{6}} }^2}  \Bigg) \\
    \tan(\beta) &= \frac{1}{\gamma\sqrt{\dsp}}\Bigg(\frac{\dsp}{2} + \frac{\sigmasp^2}{2} - \gamma^2  
    +\sqrt{\gamma^2\dsp + \paren{\gamma^2 - \paren{\frac{\dsp}{2} + \frac{\sigmasp^2}{2}}}^2} \Bigg) 
\end{align}

Consider the subclass of problem parameters, $\dsp = z, \gamma = \nicefrac{K_1}{\sqrt{z}}$ and $ \sigmasp = K_2 \sqrt{z}$ for fixed   constants $K_1, K_2 > 0$ and some variable $z > 0$, which we can vary to give us different problem instances for our toy model in \eqref{eq:toy-model-main-paper}.

\begin{align}
    \lambda_1 &= \frac{1}{8}\Bigg( \frac{K_1^2}{z}\left(1+\frac{1}{3\din}\right) + \frac{\sigmain^2}{3}\left(1-\frac{1}{\din}\right) + \frac{z}{2} + \frac{2K_2^2z}{3} + \frac{1}{6}  \nonumber \\
    & \quad\quad  +\sqrt{K_1^2 + \left(\left(\frac{K_1^2}{z}\left(1+\frac{1}{3\din}\right) + \frac{\sigmain^2}{3} \left(1-\frac{1}{\din}\right)\right) - \left(\frac{z}{2} + \frac{2K_2^2z}{3} + \frac{1}{6}\right) \right)^2}  \Bigg) \\
    \lambda_2 &= \frac{1}{8}\Bigg( \frac{K_1^2}{z}\left(1+\frac{1}{3\din}\right) + \frac{\sigmain^2}{3}\left(1-\frac{1}{\din}\right) + \frac{z}{2} + \frac{2K_2^2z}{3} + \frac{1}{6} \nonumber \\
    &\qquad -\sqrt{K_1^2 + \left(\left(\frac{K_1^2}{z}\left(1+\frac{1}{3\din}\right) + \frac{\sigmain^2}{3} \left(1-\frac{1}{\din}\right)\right) - \left(\frac{z}{2} + \frac{{2K_2^2z}}{3} + \frac{1}{6}\right) \right)^2}  \Bigg) 
\end{align}
\begin{align}
    \tilde{\lambda}_1 &= \frac{1}{8}\Bigg( \frac{K_1^2}{z} + \frac{z}{2} + \frac{K_2^2z}{2} +\sqrt{K_1^2 + \left(\frac{K_1^2}{z} - \left(\frac{z}{2} + \frac{K_2^2z}{2}\right)\right)^2} \Bigg) \\
    \tilde{\lambda}_2 &= \frac{1}{8}\left(\frac{K_1^2}{z} + \frac{z}{2} + \frac{K_2^2z}{2} -\sqrt{K_1^2 + \left(\frac{K_1^2}{z} - \left(\frac{z}{2} + \frac{K_2^2z}{2}\right)\right)^2} \right)
\end{align}
\begin{align}
    \tan(\alpha) &= \frac{1}{K_1} \Bigg( \frac{z}{2} + \frac{2K_2^2z}{3} + \frac{1}{6} - \Bigg( \frac{K_1^2}{z}\paren{1+\frac{1}{3\din}} +\frac{\sigmain^2}{3} \paren{1-\frac{1}{\din}}\Bigg) \nonumber   \\ 
    & \qquad + \sqrt{K_1^2 + \paren{\frac{K_1^2}{z}\paren{1+\frac{1}{3\din}} + \frac{\sigmain^2}{3} \paren{1-\frac{1}{\din}} - \paren{\frac{z}{2} + \frac{2K_2^2z}{3} + \frac{1}{6}}}^2} \Bigg) \\
    \tan(\beta) &=  \frac{1}{K_1} \paren{\frac{z}{2} + \frac{K_2^2z}{2} - \frac{K_1^2}{z} + \sqrt{K_1^2 + \paren{\frac{K_1^2}{z} - \paren{\frac{z}{2} + \frac{K_2^2z}{2}}}^2}} 
\end{align}

From \citet{stewart1993early}, we can use the closed form expression for the singular vectors of a $2\times 2$ full rank asymmetric matrix $\begin{bmatrix}
    a,\;\; b\\
    c,\;\; d
\end{bmatrix}$. The singular vectors are given by  
\[\begin{bmatrix}
    \cos{\theta},\;\;\hfill \sin{\theta}\\
    \sin{\theta},\;\;\hfill -\cos{\theta}
\end{bmatrix},
\] 
where, $\tan(2\theta)$ is given by:
\begin{align*}
    \tan(2\theta) = \frac{2ac+2bd}{a^2+b^2-c^2-d^2}.
\end{align*}

Now, substituting the values in the expression from~\eqref{eq:final-svd-matrix}, we get singular vectors of the above form where $\theta \in [0, \nicefrac{\pi}{2}]$ satisfies:

\begin{align}
    \theta &= \frac{1}{2} \tan^{-1}\paren{\frac{2\tan(\beta-\alpha);  \cdot(\tilde{\lambda}_1-\tilde{\lambda}_2) \cdot \sqrt{\lambda_1 \lambda_2}}{(\lambda_2\tilde{\lambda}_1-\lambda_1\tilde{\lambda}_2) - (\lambda_1\tilde{\lambda}_1- \lambda_2\tilde{\lambda}_2)\cdot \tan^2(\alpha-\beta)}} 
\end{align}

\end{proof}

\begin{lemma}[asymptotic behavior of $\tau \tan \theta$]
    \label{lem:asym-tau-tantheta} For $\gamma = \nicefrac{K_1}{\sqrt{z}}$, $\sigmasp = K_2 \sqrt{z}$,
    \[
    \lim_{z \rightarrow \infty} \tau \tan \theta = \frac{K_1K_2^2}{(1+K_2^2) 2 \sigmain^2 (1-\nicefrac{1}{\din})}
    \]
\end{lemma}
\begin{proof}
    In order to determine the asymptotic nature of $\tan(\theta)$ as $z \rightarrow \infty$, we take the limit of a slightly different term first, since we have the closed form expression of $\tan (2\theta)$.

    \begin{align*}
        \lim_{z \rightarrow \infty} \tau \tan(2 \theta) &= \sqrt{\frac{\lambda_1}{\lambda_2}} \cdot \frac{2\tan(\alpha-\beta)\cdot(\nicefrac{\tilde{\lambda}_1}{\tilde{\lambda_2}}-1)}{(\nicefrac{\tilde{\lambda}_1}{\tilde{\lambda}_2}-\nicefrac{\lambda_1}{\lambda_2}) - (\nicefrac{\lambda_1\tilde{\lambda}_1}{\lambda_2\tilde{\lambda}_2}- 1)\cdot \tan^2(\alpha-\beta)} \\
        &= 2 \tan (\alpha - \beta) \cdot \frac{\nicefrac{\tilde{\lambda_1}}{\tilde{\lambda_2}} - 1}{\nicefrac{\tilde{\lambda_1}}{\tilde{\lambda_2}} \cdot \nicefrac{\lambda_2}{\lambda_1} - 1},
    \end{align*}

since it is easy to see that $\lim_{z \rightarrow \infty} \tan^2(\alpha - \beta) \cdot \paren{\frac{\lambda_1\tilde{\lambda_1}}{\lambda_2 \tilde{\lambda_2}}-1}=0$.

If we use $\tan(\alpha - \beta) = \frac{\tan \alpha - \tan \beta}{1+\tan \alpha \tan \beta}$, and substitute the functions of $z$, for all the quantities in the above expression using Lemma~\ref{lem:svd-approximations}, we derive:
$    
\lim_{z \rightarrow \infty} \tau \tan 2\theta  = \nicefrac{2 K_1K_2^2}{(1+K_2^2)2 \sigmain^2 (1- \nicefrac{1}{\din})}.
$

Since $\tau \rightarrow \infty$, $\tan(2\theta) \rightarrow 0$, and further from Taylor approximation of $\tan(2\theta)$, $\tan (2\theta) \rightarrow 2\theta$. We can use this to derive the limit for $\tau \tan \theta$, which would just be $\nicefrac{1}{2} \cdot \nicefrac{2 K_1K_2^2}{(1+K_2^2)2 \sigmain^2 (1- \nicefrac{1}{\din})} = \nicefrac{K_1K_2^2}{(1+K_2^2)2 \sigmain^2 (1- \nicefrac{1}{\din})}$. 

\end{proof}

\begin{lemma}[asymptotic behaviors of $\cot{\alpha}, \tan \theta$]
    \label{lem:asym-cot-alpha-and-tan-theta}
For $\gamma = \nicefrac{K_1}{\sqrt{z}}$, $\sigmasp = K_2 \sqrt{z}$ following the expressions in Lemma~\ref{lem:svd-approximations},
\[
\lim_{z \rightarrow \infty} \cot \alpha = 0, \quad\quad \lim_{z \rightarrow \infty} \tan \theta = 0. 
\]
\end{lemma} 
\begin{proof}
    For $\tan \theta$, since $\tau \rightarrow \infty$, and $\tau \tan \theta$ approaches a constant (from Lemma~\ref{lem:asym-tau-tantheta}), we conclude $\lim_{z \rightarrow \infty} \tan \theta = 0$. For $\cot \alpha$, 
    $$
    \lim_{z \rightarrow \infty}  
    \frac{z}{2} + \frac{2K_2^2z}{3} + \frac{1}{6} - \Bigg( \frac{K_1^2}{z}\paren{1+\frac{1}{3\din}} +\frac{\sigmain^2}{3} \paren{1-\frac{1}{\din}}\Bigg)   = \infty,
    $$ 
    and,
    $$
    \lim_{z \rightarrow \infty}  \sqrt{K_1^2 + \paren{\frac{K_1^2}{z}\paren{1+\frac{1}{3\din}} + \frac{\sigmain^2}{3} \paren{1-\frac{1}{\din}} - \paren{\frac{z}{2} + \frac{2K_2^2z}{3} + \frac{1}{6}}}^2}  = \infty.
    $$
    Thus, $\cot \alpha \rightarrow 0$.
\end{proof}

\begin{lemma}[asymptotic behavior of $z \cot{\alpha}$]
    \label{lem:asym-cot-alpha-z}
    For $\gamma = \nicefrac{K_1}{\sqrt{z}}$, $\sigmasp = K_2 \sqrt{z}$ following the expressions in Lemma~\ref{lem:svd-approximations},
    \[
    \lim_{z \rightarrow \infty} z \cot \alpha = \frac{K_1}{1+\nicefrac{4}{3} K_2^2}.
    \]
\end{lemma}
\begin{proof}
    The expression for $z \cot \alpha$ or $\nicefrac{z}{\tan \alpha}$ follows from Lemma~\ref{lem:svd-approximations}:
    \begin{align*}
    & \lim_{z \rightarrow \infty} z \cot \alpha  = \frac{zK_1}{p+\sqrt{p^2 + K_1^2}},
    \end{align*}

where $p = \frac{K_1^2}{z}\paren{1+\frac{1}{3\din}} + \frac{\sigmain^2}{3} \paren{1-\frac{1}{\din}} - \paren{\frac{z}{2} + \frac{2K_2^2z}{3} + \frac{1}{6}}$.
Applying L'Hôpital's (relevant expressions are continuous in $z$) rule we get: $\lim_{z\rightarrow \infty} z \cot \alpha = \frac{K_1}{1+\nicefrac{4}{3}K_2^2}$.

\end{proof}

\begin{lemma}[asymptotic behavior of $\nicefrac{z}{\tau^2}$]
    \label{lem:asym-z-by-tausq}
    For $\gamma = \nicefrac{K_1}{\sqrt{z}}$, $\sigmasp = K_2 \sqrt{z}$ following the expressions in Lemma~\ref{lem:svd-approximations},
    \[
    \lim_{z \rightarrow \infty} \nicefrac{z}{\tau^2} = \frac{2\nicefrac{\sigmain^2}{3}(1-\nicefrac{1}{\din})}{1+\nicefrac{4}{3} K_2^2}. 
    \]
\end{lemma}
\begin{proof}
    For $\tau = \nicefrac{\lambda_1}{\lambda_2}$, substituting the relevant expressions from Lemma~\ref{lem:svd-approximations}, we get:
    \begin{align*}
        \nicefrac{z}{\tau^2} &= \frac{z\lambda_2}{\lambda_1} \\
            &= z \cdot \frac{\nicefrac{2K_1^2}{z}\paren{1+\nicefrac{1}{3\din}}+ 2\sigmain^2\paren{1-\nicefrac{1}{\din}} + p - \sqrt{K_1^2 + p^2}}{\nicefrac{2K_1^2}{z}\paren{1+\nicefrac{1}{3\din}}+ 2\sigmain^2\paren{1-\nicefrac{1}{\din}} + p + \sqrt{K_1^2 + p^2}},
    \end{align*}
    where $p = \nicefrac{z}{2} + \nicefrac{2K_2^2z}{3} + \nicefrac{1}{6}$. Applying L'Hôpital's (relevant expressions are continuous in $z$) rule we get: $\lim_{z\rightarrow \infty} \nicefrac{z}{\tau^2} = \frac{2\nicefrac{\sigmain^2}{3}(1-\nicefrac{1}{\din})}{1+\nicefrac{4}{3} K_2^2}$.
\end{proof}

\begin{lemma}[0-1 error of a classifier on target] \label{lemma:error_target}
Assume a classifier of the form $w = l_1 \cdot \winv + l_2 \cdot \wsp$ where $l_1, l_2 \in \Real$ and $\winv$$=$$[\wstar, 0, ..., 0]^\top$, and $\wspu = [0, ..., 0, \nicefrac{\mathbf{1}_{d_\mathrm{sp}}}{\sqrt{d_\mathrm{sp}}}]^\top$. Then the target accuracy of this classifier is given by $0.5\cdot\erfcb{-\frac{l_1\cdot \gamma}{\sqrt{2} \cdot l_2 \cdot \sigmasp}}$.
\end{lemma}
\begin{proof}
    Assume $(x,y) \sim \ProbT$. Accuracy of $w$ is given by $\Expt{\ProbT}{(\sign \left( w^\top x \right) =y)}$. 
    \begin{align*}
        \Expt{\ProbT}{\sign \left( w^\top x \right) =y} &= \Expt{\ProbT}{y\cdot \sign \left( w^\top x \right) = 1} \\
        &= \Expt{\ProbT}{y\cdot  (w^\top x ) > 0} \\
        &= \Expt{\ProbT}{y\cdot  (x^\top  (l_1 \cdot \winv + l_2 \cdot \wsp) ) > 0} \\
        &= \Expt{\ProbT}{y\cdot  (\gamma \cdot l_1 \cdot y  + l_2 \cdot \sigmasp ) > 0} \\
        &= \Expt{z \sim \cN(0,1)}{(\gamma \cdot l_1  + y\cdot l_2 \cdot \sigmasp \cdot z ) > 0} \\
        &= \Expt{z \sim \cN(0,1)}{ y\cdot l_2 \cdot \sigmasp   \cdot z > - \gamma \cdot l_1   } \\
        &= \Expt{z \sim \cN(0,1)}{ l_2 \cdot \sigmasp   \cdot z > - \gamma \cdot l_1   } \\
        &= \Expt{z \sim \cN(0,1)}{ z > - \frac{\gamma \cdot l_1}{l_2 \cdot \sigmasp}  } \\
    \end{align*}
    Using the definition of $\erfc$ function, we get the aforementioned accuracy expression. 
\end{proof}

\begin{lemma} \label{lemma:g_def}
    For $\sigma >0$ and $\mu \in \Real$, we have 
    \begin{align}
       g(\mu, \sigma) &\defeq \Expt{ z \sim \cN(0,\sigma) }{ \exp\left( -\abs{\mu + z}\right)} \\
       &= \frac{1}{2} \left(  \expb{ \nicefrac{\sigma^2}{2} - \mu} \cdot \erfcb{ -\nicefrac{\mu}{\sqrt{2} \sigma} + \nicefrac{\sigma}{\sqrt{2}} } + \expb{ \nicefrac{\sigma^2}{2} + \mu} \cdot \erfcb{ \nicefrac{\mu}{\sqrt{2} \sigma} + \nicefrac{\sigma}{\sqrt{2}} } \right)
     \end{align}
\end{lemma}

\begin{proof} The proof uses simple algebra and the definition of $\erfc$ function. 
    \begin{align*}
       g(\mu, \sigma) &\defeq \Expt{ z \sim \cN(0,\sigma) }{ \exp\left( -\abs{\mu + z}\right)} \\
        &= \frac{1}{\sqrt{2\pi}} \int_z \expb{-\abs{\mu + z}} \cdot \expb{-\frac{z^2}{2\sigma^2}} dz \\
        &= \frac{1}{\sqrt{2\pi}} \int_{-\infty}^{\infty} \expb{-\abs{\mu + z}} \cdot \expb{-\frac{z^2}{2\sigma^2}} dz \\
        &= \frac{1}{\sqrt{2\pi}} \int_{-\mu}^{\infty} \expb{-{\mu + z}} \cdot \expb{-\frac{z^2}{2\sigma^2}} dz + \frac{1}{\sqrt{2\pi}} \int_{-\infty}^{-\mu} \expb{{\mu + z}} \cdot \expb{-\frac{z^2}{2\sigma^2}} dz \\
        &= \expb{\sigma^2/2 - \mu} \int_{\frac{-\mu}{\sqrt{2} \sigma} + \frac{\sqrt{2}\sigma}{2}}^{\infty} \exp(-z^2) dz + \expb{\sigma^2/2 + \mu} \int_{-\infty}^{\frac{-\mu}{\sqrt{2} \sigma} - \frac{\sqrt{2}\sigma}{2}} \exp(-z^2) dz  \\
       &= \frac{1}{2} \left(  \expb{ \nicefrac{\sigma^2}{2} - \mu} \cdot \erfcb{ -\nicefrac{\mu}{\sqrt{2} \sigma} + \nicefrac{\sigma}{\sqrt{2}} } + \expb{ \nicefrac{\sigma^2}{2} + \mu} \cdot \erfcb{ \nicefrac{\mu}{\sqrt{2} \sigma} + \nicefrac{\sigma}{\sqrt{2}} } \right)
     \end{align*}
\end{proof}

\end{document}